\documentclass[lettersize,journal]{IEEEtran}
\usepackage{diagbox}
\usepackage{amsmath,amsfonts}
\usepackage{algorithmic}
\usepackage{algorithm}
\usepackage{array}
\usepackage[caption=false,font=normalsize,labelfont=sf,textfont=sf]{subfig}
\usepackage{float}
\usepackage{array}
\usepackage{textcomp}
\usepackage{stfloats}
\usepackage{url}
\usepackage{verbatim}
\usepackage{graphicx}
\usepackage{cite}
\usepackage{xcolor}
\usepackage{multirow}
\usepackage[T1]{fontenc}

\begin{document}
	
	\title{PreCM: The Padding-based Rotation Equivariant Convolution Mode for Semantic Segmentation}
	
	\author{Xinyu Xu, Huazhen Liu, Tao Zhang,~\IEEEmembership{Member,~IEEE}, Huilin Xiong,~\IEEEmembership{Member,~IEEE},\\and Wenxian Yu,~\IEEEmembership{Senior Member,~IEEE}
		\thanks{This paper was supported in part by the National Natural Science Foundation of China under Grant 62201343 and 62231010. \it(Corresponding author: Tao Zhang.)}
		\thanks{X. Xu, T. Zhang, H. Xiong, and W. Yu are with the Shanghai Key Laboratory of Intelligent Sensing and Recognition, School of Sensing Science and Engineering, Shanghai Jiao Tong University, Shanghai 200240, China.}
		\thanks{H. Liu is with the Intelligent Photoelectric Sensing Institute, School of Sensing Science and Engineering, Shanghai Jiao Tong University, Shanghai 200240, China.}
		\thanks{This paper have supplementary downloadable materials and codes available at https://github.com/XinyuXu414, provided by the author. The materials include the impact of interpolation on rotation equivariance, the derivation of rotation equivariance for the proposed framework, the derivation of Eq.(21), and a specific application example. Contact  sjtu-\,-zt@sjtu.edu.cn for further questions about this work.}
	}
	
	\markboth{Journal of \LaTeX\ Class Files,~Vol.~14, No.~8, August~2021}%
	{Shell \MakeLowercase{\textit{et al.}}: A Sample Article Using IEEEtran.cls for IEEE Journals}

	
	\maketitle
	
	\begin{abstract}
		Semantic segmentation is an important branch of image processing and computer vision. With the popularity of deep learning, various convolutional neural networks have been proposed for pixel-level classification and segmentation tasks. In practical scenarios, however, imaging angles are often arbitrary, encompassing instances such as water body images from remote sensing and capillary and polyp images in the medical domain, where prior orientation information is typically unavailable to guide these networks to extract more effective features. In this case, learning features from objects with diverse orientation information poses a significant challenge, as the majority of CNN-based semantic segmentation networks lack rotation equivariance  to resist the disturbance from orientation information. To address this challenge, this paper first constructs a universal convolution-group framework aimed at more fully utilizing orientation information and equipping the network with rotation equivariance. Subsequently, we mathematically design a padding-based rotation equivariant convolution mode (PreCM), which is not only applicable to multi-scale images and convolutional kernels but can also serve as a replacement component for various types of convolutions, such as dilated convolutions, transposed convolutions, and asymmetric convolution. To quantitatively assess the impact of image rotation in semantic segmentation tasks, we also propose a new evaluation metric, Rotation Difference (RD). The replacement experiments related to six existing semantic segmentation networks on three datasets (i.e., Satellite Images of Water Bodies, DRIVE, and Floodnet) show that, the average Intersection Over Union (IOU) of their PreCM-based versions respectively improve 6.91\%, 10.63\%, 4.53\%, 5.93\%, 7.48\%, 8.33\% compared to their original versions in terms of random angle rotation. And the average RD values are decreased by 3.58\%, 4.56\%, 3.47\%, 3.66\%, 3.47\%, 3.43\% respectively. The code can be download from https://github.com/XinyuXu414.
		
	\end{abstract}
	
	
	\section{Introduction}
	\IEEEPARstart{S}{emantic} segmentation has been a fundamental and challenging task in computer vision for many years \cite{1}. Since it can provide category information at the pixel level, it has been widely used in various fields, such as autonomous vehicles \cite{2}, water detection \cite{xu2024information}, road segmentation \cite{4}, and defect detection \cite{5}.
	
	Within the context of traditional machine learning, semantic segmentation methods are normally constructed as classifiers, like Support Vector Machine (SVM) \cite{7} and Random Forest (RF) \cite{8}. In addition, considering the use of texture information to ensure the consistency between pixel labels, context models such as Markov Random Fields (MRF) \cite{9} and Conditional Random Fields (CRF) \cite{10} are also utilized as classical frameworks for semantic segmentation. Although these algorithms are convenient and easy to implement, their poor feature extraction ability limits their generalization capability and segmentation accuracy.
	
	With the development of deep learning, CNNs have proven their abilities to extract more advanced features of images for semantic segmentation. Specifically, the translation equivariance of convolution kernels enables CNNs to better capture spatial features in images \cite{wei2023rotational}. In \cite{11}, Long \emph{et al}. used CNN-based fully convolutional networks to produce correspondingly-sized output with efficient inference and learning for semantic segmentation. Chen \emph{et al}. \cite{21} combined deep convolutional networks, atrous convolution, and fully connected conditional random fields to achieve semantic segmentation. Yang \emph{et al}. \cite{6} proposed Densely connected Atrous Spatial Pyramid Pooling (DenseASPP) for street scenes segmentation, which connected a set of atrous convolutional layers in a dense way to produce multi-scale features. Although these networks have achieved significant success in different semantic segmentation tasks, convolutional operations often fail to guarantee the rotation equivariance of networks, making it difficult for them to resist the interference caused by rotation \cite{15}.
	
	To address this issue, most research has focused on two aspects: data extension and network design \cite{della2019deep}. For the former, data augmentation is a prevalent approach since it can enhance the diversity of the dataset, enabling the model to generalize better to unknown data. In \cite{17}, Mikołajczyk \emph{et al}. investigated the influence of multiple data augmentation methods on the image classification task, and proposed a new data augmentation method. Choi \emph{et al}. \cite{choi} presented a data augmentation method based on Generative Adversarial Networks (GANs), which could enhance the performance of segmentation networks on the target domain. Olsson \emph{et al}. \cite{olsson} developed a semi-supervised semantic segmentation algorithm that used ClassMix, a new data augmentation technique to generate augmentation by mixing unlabeled samples. Although data augmentation technology can effectively help networks reduce their attentions to irrelevant features by introducing various rotated data, it also has some obvious limitations. For example, the more orientation changes are considered, the more time costs and calculation budgets need to be consumed \cite{19,fei2024rotation}.
	
	In the field of network design, the majority of current efforts fucus on achieving the strict rotation equivariance at the four specific angles 0$^\circ$, 90$^\circ$, 180$^\circ$, and 270$^\circ$ \cite{14,16,romero2020attentive,he2021efficient}. The reason is that interpolation had been theoretically identified as a crucial factor affecting rotation equivariance on the other angles except for [0$^\circ$, 90$^\circ$, 180$^\circ$, 270$^\circ$] \cite{marcos,mo2024ric,fu2024rotation} (seeing more explanations Appendix A). Among them, equivariant networks based on group theory have gained widespread application due to their solid mathematical theoretical foundation and excellent experimental performance \cite{14,16,romero2020attentive}. However, in practice, we have observed that these methods have strict requirements on the sizes of images and convolution kernels, as well as network hyperparameters, which limits their versatility to some extent \cite{14}. In light of this, this paper first proposes a padding-based rotation equivariant convolution mode (PreCM) that can be flexibly applied to multi-scale images and convolution kernels. Then, we further design it as a replacement component for traditional convolution layers, enabling easy substitution of different types of convolution patterns (such as dilated convolution, transposed convolution, and asymmetric convolution). The experiments on three segmentation datasets prove that, once the replacement is adopted, we can  not only realize the strict rotation equivariance at 0$^\circ$, 90$^\circ$, 180$^\circ$, 270$^\circ$, but also achieve obvious improvements in performance and robustness at the other rotation angles.
	
	In a nutshell, our contributions are as follows:
	\begin{itemize}
		\item{Based on the group theory, we first build a rotation equivariant convolution-group framework to extract the features related to orientational information, and also mathematically prove its rotation equivariance.}
		\item{For the proposed rotation equivariant convolution-group framework, we further give a concrete implementation, i.e., designing the padding-based rotation equivariant convolution mode (PreCM) that can be not only flexibly applied to multi-scale images and convolution kernels, but also used as a replacement component to replace convolution operations for making networks rotation equivariant.}
		\item{To quantitatively assess the impact of rotation on segmentation results, a new metric named RD (rotation difference) is correspondingly proposed.}
		\item{We select six semantic segmentation networks and replace their convolution layers with PreCM. Extensive experiments on binary segmentation, small-sample segmentation, and multi-class segmentation tasks fully validate the effectiveness of PreCM in enhancing performance and robustness. Furthermore, compared to the networks with data augmentation, the networks employing PreCM also demonstrate comparable segmentation performance without increasing training samples.}
		
	\end{itemize}
	
	The rest of this paper is organized as follows. Related work is discussed in Section II. Section III describes the methodology. The experimental results and analyses are given in Section IV. Section V concludes this paper.


	\section{Related Work}
	In this section, the works related to semantic segmentation networks and rotation equivariance are introduced.
	\subsection{Semantic Segmentation Networks}
	Convolutional Neural Networks (CNNs) have become one of the indispensable tools for achieving semantic segmentation. With the aid of CNN-based methods, we are able to perform pixel-level precise segmentation on images, thereby achieving advanced semantic understanding of image content \cite{20}. In this vein, Ronneberger \emph{et al}. \cite{12} proposed U-net for biomedical image segmentation. Badrinarayanan \emph{et al} \cite{23}. presented a novel and practical deep fully convolutional neural network SegNet for semantic pixel-wise segmentation. Zhao \emph{et al}. \cite{24} designed the pyramid scene parsing network PSPNet with the global context information. Chen \emph{et al}. \cite{chen} built the network DeepLab that used dilated convolution and fully connected CRFs to achieve semantic segmentation.
	
	In addition, given the real-time requirements and computational resource constraints, many lightweight semantic segmentation networks with few parameters are also considered. For example, Paszke \emph{et al}. \cite{13} built an efficient deep neural network architecture named ENet, which was specifically created for the tasks requiring low latency operation. In \cite{25}, Romera  \emph{et al}. proposed a deep architecture ERFNet that can run in real time while providing accurate semantic segmentation. Yu \emph{et al}. \cite{26} designed an efficient and effective network BiSeNet V2 with a good trade-off between speed and accuracy, wherein the spatial details and categorical semantics were separately treated so as to achieve real-time semantic segmentation. Wang \emph{et al}. \cite{wang2019lednet} presented a lightweight network LEDNet to deal wiht the problem of extensive computational burden for the task of real-time semantic segmentation.
	
	In a summary, semantic segmentation is an important branch of scene interpretation. How to establish more efficient semantic segmentation networks is still worthy studying.
	
	\subsection{Rotation Equivariant Networks}
	Despite the high segmentation accuracy achieved by the existing segmentation networks, they exhibit certain potential sensitivity to orientation when extracting semantic information, leading to a decline in performance when processing images with varying imaging angles. Therefore, some researchers study the rotation equivariance of networks for enhancing the feature extraction ability. In \cite{worrall2017harmonic}, Worrall \emph{et al}.  proposed Harmonic Networks (H-Nets), which achieved translation and rotation equivariance by replacing conventional CNN filters with circular harmonics. Mo \emph{et al}. \cite{mo2024ric} proposed a rotation-invariant coordinate network (RIC-CNN), which achieved natural invariance to
	arbitrary rotations around the input center without additional trainable parameters or data augmentation.
	
	Besides, with the help of group theory, Cohen \emph{et al}. \cite{14} proposed the group equivariant convolutional neural network (G-CNNs), whose core innovation lies in the first adoption of group equivariant convolutions based on \{0$^\circ$, 90$^\circ$, 180$^\circ$, 270$^\circ$\}. Since then, G-CNNs have been widely used and improved. For instance, in \cite{16}, Li \emph{et al}. further developed a deep rotation equivariant network consisting of cycle layers, isotonic layers, and decycle layers, which applied rotation transformation to filters rather than feature maps. Han \emph{et al}. \cite{29} designed a rotation-equivariant detector ReDet, wherein G-CNNs were introduced as well.  Weiler \emph{et al}. \cite{weiler2019general} built a general solution of the kernel space constraint for arbitrary representations of the Euclidean group E(2) and its subgroups based on group representations theory. Although this group equivariant convolution is extensively used, the networks derived from it require strict restrictions on the size of images and convolution kernels, which will be proved in \ref{3.1}. From the perspective of networks' input and output, these restrictions indicate that we have to resample the images, leading to a loss of context information. From the perspective of network utilization, they make many classical combinations of feature map sizes and convolution kernel parameters as well as some classical convolutions (like asymmetric convolution) hard to work. Therefore, it is necessary for us to redesign a new group equivariant convolution that can be applied in multi-scale images and   convolution kernels.
	
	\section{Methodology}
	\subsection{Preliminary Knowledge}
	\label{3.1}
\begin{figure}[!t]
		\centering
		\includegraphics[width=2.5in]{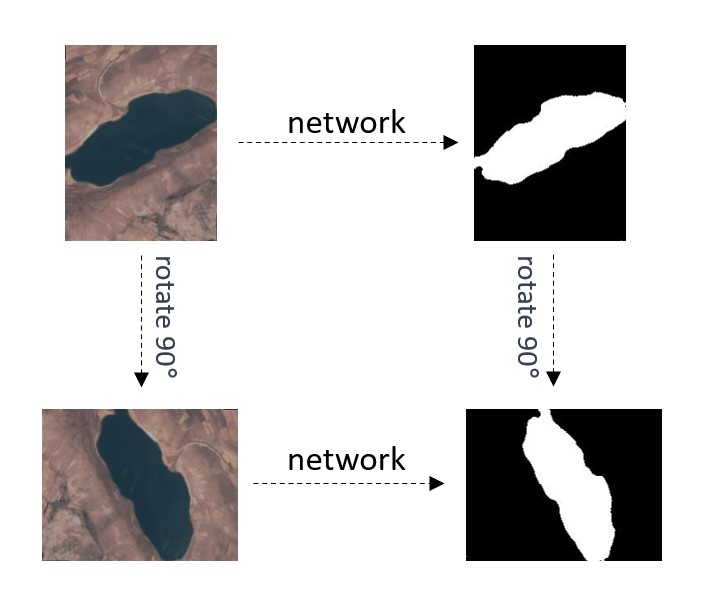}
		\caption{An example of a network that has rotation equivariance.}
		\label{fig_1}
	\end{figure}

	\textbf{Rotation Equivariance } In neural networks, rotation equivariance refers to the property that when an image is rotated, the features extracted from it undergo the same rotation transformation \cite{14}. Specifically, if the input of the network is rotated, the output will do the same ordered rotation transformation. Fig. \ref{fig_1} gives an example of a network that has rotation equivariance. Mathematically, the rotation equivariance can be defined as
	\begin{equation}
		\label{equi}
		g(\sigma x) = \sigma g(x),
	\end{equation}
	where $\sigma$ is a transformation operation, $g$($\cdot$) represents a function that maps one feature space to the other feature space, and $x$ represents the input image.
	
	From Eq. (\ref{equi}), we can observe that image rotation and mapping are commutative, meaning that the feature map obtained by first rotating the image and then mapping it is identical to the feature map obtained by first mapping the image and then rotating it. If Eq. (\ref{equi}) is satisfied, then the network has rotation equivariance.
	
	\textbf{Restriction } In previous work, rotation equivariant networks based on group theory have gained widespread application due to their solid mathematical theoretical foundation and excellent experimental performance. In fact, the related group-based method relies on a fundamental convolution distributive law, which is the core for achieving rotation equivariance \cite{14}. In math, it can be expressed as
	\begin{equation}
		({r_\sigma }\varphi ) * ({r_\sigma }f) = {r_\sigma }(\varphi  * f),
	\end{equation}
	where $f$ is feature map, $\varphi$ represents convolution kernel, and ${r_\sigma }$ means rotation operation.
	
\begin{figure}[t!]
		\centering
		\includegraphics[width=3in]{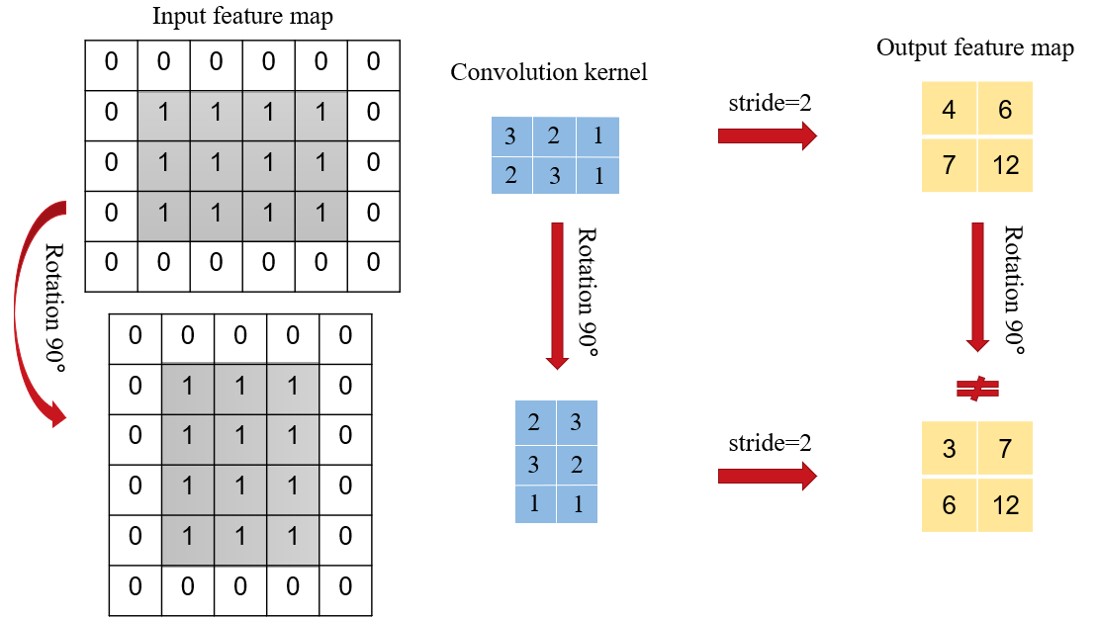}
		\caption{An example of the limitation of convolution distributive law.}
		\label{Fig_2}
	\end{figure}

	However, in practical experimental deployments, we have found that this law does not hold unconditionally, and previous work has also identified this issue \cite{16}. Fig. \ref{Fig_2} provides a concrete example, clearly demonstrating that the result of convolution between a rotated kernel and a rotated image is not equivalent to the result of rotating the output of the convolution between the kernel and the image. Apparently, this fact restricts the applications of the existing group theory-related methods. Given this, we propose a specific framework in this paper that can be implemented at multiple-scale figures and convolution kernels, and innovatively integrates it as a replacement for various traditional convolutions.
	
	\subsection{Rotation Equivariant Convolution-group Framework}
	\textbf{Symbol Clarification } The group theory  proposed in  \cite{14} has already shown great potential to keep the networks rotation equivariant. Inspired by it, we here construct a rotation equivariant convolution-group framework, which mainly contains three layers. Noted that, different from the previous applications of group theory on networks, we use group theory to represent the processing modes of features and information, rather than feature maps.
	
	First of all, a four-element cyclic group $\sigma  = \{ {\sigma _0},{\sigma _1},{\sigma _2},{\sigma _3}\} $  is introduced, satisfying
	\begin{equation}
		\label{eq3}
		\sigma _i^{ - 1} = {\sigma _{(4 - i)\bmod (4)}},
	\end{equation}
	\begin{equation}
		\label{eq4}
		{\sigma _i}{\sigma _j} = {\sigma _j}{\sigma _i}= {\sigma _{(i + j)\bmod (4)}},
	\end{equation}
	where ${\sigma _i},{\sigma _j} \in {\sigma}$, $mod$ indicates modulo operation. Each element in $\sigma$ represents the orientation information function at a certain layer of network. At the same time, $r = \{ {r_{{\sigma _i}}}\} $ is defined to represent the orientation information of feature map or convolution kernel, which isomorphisms with $\sigma$, viz.,
	\begin{equation}
		\label{5}
		{r_{{\sigma _i}{\sigma _j}}} = {r_{{\sigma _i}}}{r_{{\sigma _j}}},
	\end{equation}
	where  the subscript $\sigma$ is the index of $r$. Meanwhile, for any convolution kernel $\varphi$ and feature map $f$, we define
	\begin{equation}
		{\sigma _i}(\varphi ,f) = {r_{{\sigma _i}}}\varphi { \otimes _{{\sigma _i}}}f,
	\end{equation}
	here, ${ \otimes _{{\sigma _i}}}$ means the specific convolution mode under ${\sigma _i}$. The reason why ${ \otimes _{{\sigma _i}}}$ is designed is that, different hyperparameters in the convolution process, such as stride, padding, and convolution center, easily affect the equivariant acquisition of rotation information. So, to solve this problem as well as ensure the consistency of rotation information acquisition, convolution distributive law must be satisfied.
	For this goal,  it is necessary for us to simultaneously utilize different convolution modes, which details will be given and discussed in the next subsection.	
	
	Then, the related operations of group theory are introduced. Since the networks are composed of multiple layers, we naturally represent them using the binary operation of groups. More specifically, we use direct product to represent the extraction of rotation information, and quotient group to represent the compression and fusion of rotation information. Hereinafter, the symbol $\overline{\times}$ and $\overline{/}$ respectively mean the direct product and the quotient group.
	
	\textbf{Network Design }Firstly,  the group ${G_1} = \left\{ {\left. {{\sigma _i}} \right|{\sigma _i} \in \sigma } \right\}$ is used to extract the rotation information from  \{0$^\circ$, 90$^\circ$, 180$^\circ$, 270$^\circ$\} to form the first layer, i.e.,
	\begin{equation}
		[f_{{\sigma _i}}^1] = [{\sigma _i}({\varphi ^1},{f^1})] = [({r_{{\sigma _i}}}{\varphi ^1}){ \otimes _{{\sigma _i}}}{f^1}],
	\end{equation}
	where ${f^1}$ and ${\varphi ^1}$  respectively represent the input feature map and the convolution kernel of first layer. Subsequently, to make the network have rotation equivariance, two ideas can be easily got. One is to use the symmetric convolution kernel \cite{ma2020pcfnet}, degenerating $G_1$ to trivial group, i.e., $G_1  \cong \{ e\}$. However, this approach encounters the challenge of constrained parameters and limited feature space, making it difficult for the training process to converge. The other is to use the quotient group to compress the feature information of four orientations, expressed as $G_1 \overline{/}G_1  \cong \{ e\}$. But, this idea will cause 4 times the parameter consumption. Considering these, in this paper, we utilize the group direct product ${G_2} = \sigma  \overline{\times} \sigma  = \left\{ {\left. {({\sigma _j},{\sigma _i})} \right|{\sigma _j},{\sigma _i} \in \sigma } \right\}$ to extract the rotation information with the first layer, and the second layer of network is correspondingly designed as
	\begin{equation}
		\begin{aligned}
			\left[ {({\sigma _j},{\sigma _i})(\varphi _{\sigma _j^{ - 1}{\sigma _i}}^2,{\varphi ^1},{f^1})} \right] &= \left[{\sigma _j}(\varphi _{\sigma _j^{ - 1}{\sigma _i}}^2,{\sigma _i}({\varphi ^1},{f^1}))\right]\\
			&= \left[({r_{{\sigma _j}}}\varphi _{\sigma _j^{ - 1}{\sigma _i}}^2){ \otimes _{{\sigma _j}}}(f_{{\sigma _i}}^1)\right],
		\end{aligned}
	\end{equation}
	where $\varphi _{\sigma _j^{ - 1}{\sigma _i}}^2$ represents the convolution kernels corresponding to the relative orientation information between two layers.
	
	Then, to fuse the features of four directions, we use the quotient group to compress the results of Eq. (8). Let ${{\bar G}_2} = \left\{ {\left. {({\sigma _0},{\sigma _i})} \right|{\sigma _i} \in \sigma } \right\}$, the updated second layer will become
	\begin{equation}
		{G_2}' = {G_2}\overline{/}{{\bar G}_2} = \left\{ {\left. {({\sigma _j},{\sigma _0}){{\bar G}_2}} \right|{\sigma _j} \in \sigma } \right\},
	\end{equation}
	\begin{equation}
		\begin{aligned}
			\left[ f_{{\sigma _j}}^2 \right] &=\left[ {({\sigma _j},{\sigma _0}){{\bar G}_2}(\varphi _{\sigma _j^{ - 1}{\sigma _i}}^2,{\varphi ^1},{f^1})} \right] \\&= \left[ \sum\limits_{{\sigma _i} \in \sigma } {({\sigma _j},{\sigma _i})(\varphi _{\sigma _j^{ - 1}{\sigma _i}}^2,{\varphi ^1},{f^1})} \right].
		\end{aligned}
	\end{equation}
	Obviously, due to ${G_1} \cong {G_2}'$, the second layer $f_{{\sigma _j}}^2$ can be reused.
	
	At last, in order to obtain the equivariant result, we add the third layer to perform weighted fusion on the feature maps obtained from the second layer. Given the quotient group ${G_3} = {G_2}'\overline{/}{G_2}' \cong \left\{ e \right\}$, we design
	\begin{equation}
		{f^3} = {G_3}({\varphi ^3},\varphi _{\sigma _j^{ - 1}{\sigma _i}}^2,{\varphi ^1},{f^1}) = \sum\limits_{{\sigma _j} \in \sigma } {({r_{{\sigma _j}}}{\varphi ^3}){ \otimes _{{\sigma _j}}}(f_{{\sigma _j}}^2)},
	\end{equation}
	where ${\varphi ^3}$ denotes the convolution kernel of the third layer $f^3$.
	\begin{figure}[!t]
		\centering
		\includegraphics[width=3.5in]{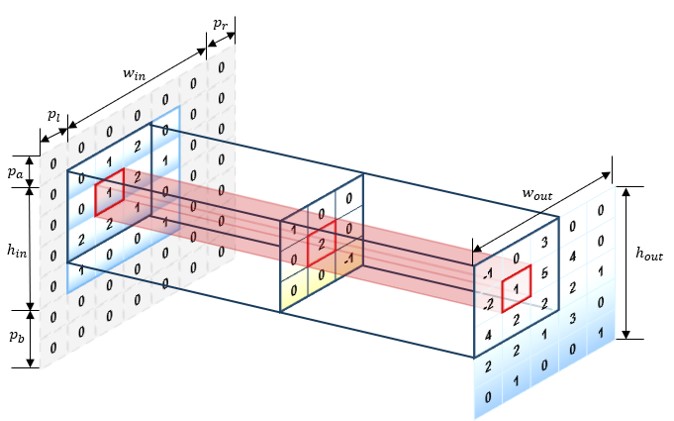}
		\captionsetup{justification=centering}
		\caption{ An example of padding-based convolution process.}
		\label{Fig_3}
	\end{figure}
	
	It should be stressed that, to completely achieve rotation equivariance of network, the three layers $f_{{\sigma _i}}^1$, $f_{{\sigma _j}}^2$, and $f^3$ must be used simultaneously. For the second layer, it can be used  in cycles. The detailed proof of the rotation equivariance of this convolution-group framework are given in Appendix B.
	
	\subsection{Padding-based Rotation Equivariant Convolution Mode (PreCM)}
	In the former subsection, we introduced ${ \otimes _{{\sigma _i}}}$ to represent different convolution modes. Hereinafter, a detailed realization scheme is proposed through padding. The reason why we choose padding as the tool is mainly due to that it can make different convolution modes have similar hyperparameters without discarding the original information as much as possible.
	
	In detail, we firstly define the convolution expression with padding as
	\begin{equation}\label{convsigma}
		\begin{aligned}
			{({f_{out}})_{{w_{out}} \times {h_{out}}}} &= {(\varphi )_{{w_\varphi } \times {h_\varphi }}}{ \otimes _{{\sigma _i}}}{({f_{in}})_{{w_{in}} \times {h_{in}}}}\\
			&= {(\varphi )_{{w_\varphi } \times {h_\varphi }}} \otimes [{({f_{in}})_{{w_{in}} \times {h_{in}}}} \oplus {P_{{\sigma _i}}}],
		\end{aligned}
	\end{equation}
	where $f_{out}$ represents the output feature map, $f_{in}$ means the input feature map, $({w_{out}},{h_{out}})$ and $({w_{in}},{h_{in}})$ are the width and height of the output and input feature maps, respectively. $({w_\varphi },{h_\varphi })$ is the width and height of convolution kernel. The symbol $\oplus$ means the operation of adding padding to the input feature map. ${P_{{\sigma _i}}}$ is the group of padding methods under ${\sigma _i}$ , viz., ${P_{{\sigma _i}}} = [{p_{{\sigma _i},a}},{p_{{\sigma _i},b}},{p_{{\sigma _i},l}},{p_{{\sigma _i},r}}]$, where ${p_{{\sigma _i},a}}$ is the padding above image, ${p_{{\sigma _i},b}}$ is the padding below image, ${p_{{\sigma _i},l}}$ is the padding on the left, and ${p_{{\sigma _i},r}}$ is the padding on the right. Note that, when ${\sigma _i}={\sigma _0}$, the subscript associated with the convolution mode is omitted. A simple example of padding-based convolution process is shown in Fig. \ref{Fig_3}.
	
\begin{figure*}[ht!]
		\centering
		\includegraphics[width=7in]{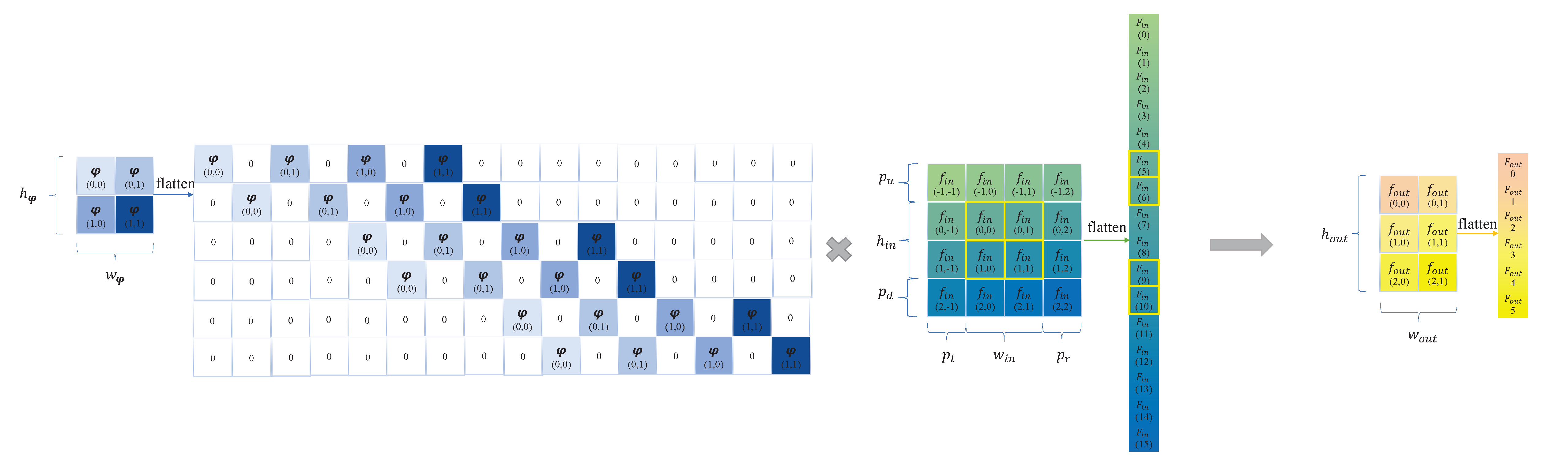}
		\captionsetup{justification=centering}
		\caption{ An example of convolution in flatten form.}
		\label{Fig_4}
	\end{figure*}

	Then, we further adopt the matrix expansion \cite{30} to make Eq. (12)  become the matrix multiplication form, that is,
	
	\begin{equation}
		\label{13}
		\begin{aligned}
			\begin{array}{l}
				{\left( {{F_{out}}} \right)_{{w_{out}}{h_{out}} \times 1}} = {\left( \Phi  \right)_{{w_{out}}{h_{out}} \times w{'_{in}}h{'_{in}}}}{\left( {{F_{in}}} \right)_{w{'_{in}}h{'_{in}} \times 1}},
			\end{array}
		\end{aligned}
	\end{equation}
	where $F_{out}$, $\Phi$, and $F_{in}$ respectively correspond to the matrix forms of $f_{out}, \varphi$, and $f_{in}$. $w{'_{in}} = {p_l} + {p_r} + {w_{in}},h{'_{in}} = {p_a} + {p_b} + {h_{in}}$.   The specific transform process is also given in Eq. (14),  where $m$ represents the $m$-th element of the column vector ${F_{out}}$, $n$ means the $n$-th element of the column vector ${F_{in}}$, $(u,v)$ denotes the $u$-th row and $v$-th column of matrix $\Phi$. $(w_{s},h_{s})$ is the stride in the directions of width and height, respectively, and $(w_{d},h_{d})$ are the dilation in the direction of width and height, respectively. $\left\langle A \right\rangle$ means the downward rounded result of $A$ and the symbol $\left\{ {A|B} \right\}$ is the remainder of $A$ divided by $B$. Fig. \ref{Fig_4} shows a concrete example.
	
	\begin{figure*}[ht]
		\begin{flalign}
			\label{flatten}
			\begin{aligned}
				\begin{array}{l}
					{F_{out}}(m) = flatten({f_{out}})(m): = {f_{out}}\left( {\left\langle {m/{w_{out}}} \right\rangle ,\{ m|{w_{out}}\} } \right),\\
					{F_{in}}(n) = flatten({f_{in}}\oplus{p_a},{p_b},{p_l},{p_r})(n): = {f_{in}}\left( {\left\langle {n/w{'_{in}}} \right\rangle  - {p_a},\left\{ {n|w{'_{in}}} \right\} - {p_l}} \right),\\
					\Phi (u,v) = sparse(\varphi )(u,v): = \varphi \left( {\frac{{\left\langle {v/w{'_{in}}} \right\rangle  - \left\langle {u/{w_{out}}} \right\rangle {h_s}}}{{{h_d}}},\frac{{\left\{ {v|w{'_{in}}} \right\} - \{ u|{w_{out}}\} {w_s}}}{{{w_d}}}} \right),\\
					m = 0,1,2,...,{w_{out}}{h_{out}} - 1,n = 0,1,2,...,h{'_{in}}w{'_{in}} - 1,u = 0,1,2,...,{w_{out}}{h_{out}} - 1,v = 0,1,2,...,h{'_{in}}w{'_{in}} - 1.
				\end{array}
			\end{aligned}
		\end{flalign}
		{\noindent}	 \rule[-10pt]{17.5cm}{0.05em}\\
	\end{figure*}

	After this, we use the form of matrix multiplication to derive the conditions required for rotation equivariance. Let
	\begin{equation}
		\label{deqn_ex1a}
		\begin{aligned}
			\begin{array}{l}
				{[{F_{in}}]_t} = flatten(({r_{{\sigma _t}}}{f_{in}}) \oplus {P_{{\sigma _t}}}),\\[1mm]
				{[{F_{out}}]_t} = flatten({r_{{\sigma _t}}}{f_{out}}),\\[1mm]
				{[\Phi ]_t} = sparse({r_{{\sigma _t}}}\varphi),
			\end{array}
		\end{aligned}
	\end{equation}
	where ${{\sigma_t} \in \sigma}$. Then, Eq. (2) can be rewritten as
	\begin{equation}
		\begin{aligned}
			&({r_{{\sigma _t}}}\varphi) { \otimes _{{\sigma _t}}}({r_{{\sigma _t}}}f) = {r_{{\sigma _t}}}(\varphi { \otimes _{{\sigma _0}}}f)\\[1mm]
			\Rightarrow &{\left[ \phi  \right]_t}{\left[ {{F_{in}}} \right]_t} = {\left[ {{F_{out}}} \right]_t},
		\end{aligned}
	\end{equation}
	Next, we will separately consider the left-hand side and the right-hand side of Eq. (16), and explore the conditions that need to be met for the equation to hold true.
	
	First, let us consider the right-hand side of Eq. (16).  ${[{F_{out}}]_t}$ is the result of flattening the output feature map after ${r_{{\sigma _t}}}$ (i.e., a column vector). Thus, there has a row transformation relationship between ${[{F_{out}}]_t}$ and ${[{F_{out}}]}$. Eq. (\ref{eq21}) shows the detailed process, wherein ${R_{t,out}}$ represents the row transformation function between ${[{F_{out}}]_t}$ and ${[{F_{out}}]}$ under ${\sigma _t}$. Further, we use the matrix form of convolution to describe Eq. (17), and obtain
	
	\begin{figure*}[ht]
		\begin{flalign}
			\label{eq21}
			\begin{array}{l}
				{[{F_{out}}]_t}(m) = {F_{out}}({R_{t,out}}(m)),\\[1mm]
				{R_{t,out}}(m) = \left[ {\begin{array}{*{20}{c}}
						{{w_{out}}}&1
				\end{array}} \right]\left( {\left[ {\begin{array}{*{20}{c}}
							{\cos \frac{{\pi t}}{2}}&{\sin \frac{{\pi t}}{2}}\\
							{ - \sin \frac{{\pi t}}{2}}&{\cos \frac{{\pi t}}{2}}
					\end{array}} \right]} \right.\left. {\left[ {\begin{array}{*{20}{c}}
							{\left\langle {m/{w_{{\sigma _t},out}}} \right\rangle  - \frac{{{h_{{\sigma _t},out}} - 1}}{2}}\\
							{\left\{ {m|{w_{{\sigma _t},out}}} \right\} - \frac{{{w_{{\sigma _t},out}} - 1}}{2}}
					\end{array}} \right] + \left[ {\begin{array}{*{20}{c}}
							{\frac{{{h_{out}} - 1}}{2}}\\
							{\frac{{{w_{out}} - 1}}{2}}
					\end{array}} \right]} \right).
			\end{array}
		\end{flalign}
		{\noindent}	 \rule[-10pt]{17.5cm}{0.05em}\\
	\end{figure*}
	\begin{equation}
		\label{eq22}
		\begin{aligned}
			&{[{F_{out}}]_t}(m) = {F_{out}}({R_{t,out}}(m))\\
			&= \sum\limits_{n = 0}^{{h_{out}}{w_{out}} - 1} {\Phi ({R_{t,out}}(m),n)} {F_{in}}(n)\\
			&= \sum\limits_{n = 0}^{{h_{out}}{w_{out}} - 1} {\Phi ({R_{t,out}}(m),{R_{t,in}}(n))} {F_{in}}({R_{t,in}}(n)),
		\end{aligned}
	\end{equation}
	where ${F_{in}}({R_{t,in}}(n)) = flatten({r_{{\sigma _t}}}f_{in}^{'})(n)$.
	
	Secondly, considering the left-hand side of Eq.(16),  i.e. ${[\phi ]_t}{[{F_{in}}]_t}$, it can be recast as Eq. (\ref{eq23}), where ${w_{{\sigma _t},*}}$ represents hyperparameters related to the size of feature map and the convolution under ${\sigma _t}$. (${w_{{\sigma _t},\phi}}$,${h_{{\sigma _t},\phi}}$) represents the width and height of the convolution kernal after rotating ${r_{\sigma _t}}$, respectively. Comparing Eqs. ($\ref{eq22}$) and ($\ref{eq23}$), we can easily get

	\begin{figure*}[ht]
		\begin{flalign}
			\label{eq23}
			\begin{aligned}
				\begin{array}{l}
					({[\phi ]_t}{[{F_{in}}]_t})(m) = \sum\limits_{n = 0}^{{h_{{\sigma _t},out}}{w_{{\sigma _t},out}} - 1} {{{[\phi ]}_t}(m,n){{[{F_{in}}]}_t}(n)}  = \sum\limits_{n = 0}^{{h_{{\sigma _t},out}}{w_{{\sigma _t},out}} - 1} {\Phi ({R_{t,\varphi }}(m,n))} {[{F_{in}}]_t}(n),\\[1mm]
					{R_{t,\varphi }}(m,n) = \left( {\left[ {\begin{array}{*{20}{c}}
								{\cos \frac{{\pi t}}{2}}&{\sin \frac{{\pi t}}{2}}\\[1mm]
								{ - \sin \frac{{\pi t}}{2}}&{\cos \frac{{\pi t}}{2}}
						\end{array}} \right]\left[ {\begin{array}{*{20}{c}}
								{\frac{1}{{{h_{{\sigma _t},d}}}}\left( {\left\langle {n/{w_{{\sigma _t},in}}} \right\rangle  - \left\langle {m/{w_{{\sigma _t},out}}} \right\rangle {h_{{\sigma _t},s}}} \right) - \frac{{{h_{{\sigma _t},\varphi }} - 1}}{2}}\\[1mm]
								{\frac{1}{{{w_{{\sigma _t},d}}}}\left( {\left\{ {n|{w_{{\sigma _t},in}}} \right\} - \left\{ {m|{w_{{\sigma _t},out}}} \right\}{w_{{\sigma _t},s}}} \right) - \frac{{{w_{{\sigma _t},\varphi }} - 1}}{2}}
						\end{array}} \right] + \left[ {\begin{array}{*{20}{c}}
								{\frac{{{h_\varphi } - 1}}{2}}\\[1mm]
								{\frac{{{w_\varphi } - 1}}{2}}
						\end{array}} \right]} \right).
				\end{array}
			\end{aligned}
		\end{flalign}
		{\noindent}	 \rule[-10pt]{17.5cm}{0.05em}\\
	\end{figure*}
	\begin{equation}
		\label{eq25}
		\begin{aligned}
			\begin{array}{l}
				{[{F_{in}}]_t}(n) = {F_{in}}({R_{t,in}}(n)),\\[1mm]
				{h_{out}}{w_{out}} = {h_{{\sigma _t},out}}{w_{{\sigma _t},out}},\\[1mm]
				\Phi ({R_{t,out}}(m),{R_{t,in}}(n)) = \Phi ({R_{t,\varphi }}(m,n)).
			\end{array}
		\end{aligned}
	\end{equation}
	After further derivations, the following conclusions can be attained as well, i.e.,
	\begin{subequations}
		\begin{equation}
			\label{21a}
			\begin{aligned}
				\begin{array}{l}
					{w_{out}} = ({p_l} + {p_r} + {w_{in}} - {w_d}({w_\varphi } - 1) - 1)/{w_s} + 1,\\
					{h_{out}} = ({p_a} + {p_b} + {h_{in}} - {h_d}({h_\varphi } - 1) - 1)/{h_s} + 1,\\
				\end{array}
			\end{aligned}
		\end{equation}
		\\
		\begin{equation}
			\begin{aligned}
				\label{21b}
				\begin{array}{l}
					\left[ {\begin{array}{*{20}{c}}
							{{p_{\sigma_t,a}}}\\
							{{p_{\sigma_t,l}}}\\
							{{p_{\sigma_t,b}}}\\
							{{p_{\sigma_t,r}}}
					\end{array}} \right] = {\left[ {\begin{array}{*{20}{c}}
								0&0&0&1\\
								1&0&0&0\\
								0&1&0&0\\
								0&0&1&0
						\end{array}} \right]^t}\left[ {\begin{array}{*{20}{c}}
							{{p_a}}\\
							{{p_l}}\\
							{{p_b}}\\
							{{p_r}}
					\end{array}} \right].
				\end{array}
			\end{aligned}
		\end{equation}
	\end{subequations}
	Note that, the detailed deductions of Eqs. (21a) and (21b) are respectively presented in  Appendixes C and D.
	
	So far, a concrete padding-based rotation equivariant mode has been given. Meanwhile, we emphasize that there are many ways to realize Eq. ($\ref{21a}$), and researchers are free to set its hyperparameters according to practical need, but here we recommend using Eq. ($\ref{22}$), which can avoid unnecessary padding as well as satisfy Eq. ($\ref{21a}$).
	\begin{equation}
		\label{22}
		\begin{aligned}
			\begin{array}{l}
				{p_a} = {p_{ab}} - {p_b},\\
				{p_r} = {p_{rl}} - {p_l},\\
				{p_b} = \left\langle {{p_{ab}}|2} \right\rangle,\\
				{p_l} = \left\langle {{p_{rl}}|2} \right\rangle,\\
				{p_{ab}} = \left( {{h_{out}} - 1} \right){h_s} + {h_d}({h_\varphi } - 1) + 1 - {h_{in}},\\
				{p_{rl}} = \left( {{w_{out}} - 1} \right){w_s} + {w_d}({w_\varphi } - 1) + 1 - {w_{in}}.
			\end{array}
		\end{aligned}
	\end{equation}
	
	To sum up, in this subsection, we use padding to design a specific convolutional mode to achieve rotation equivariance. The most significant advantage of this mode is that it can avoid the influence of network hyperparameters, image sizes, and convolution types on rotation equivariance, thereby extending the rotation equivariance to a wider range of applications.
	
	\subsection{Application of PreCM}
\begin{figure}[t!]
		\centering
		\includegraphics[width=3in]{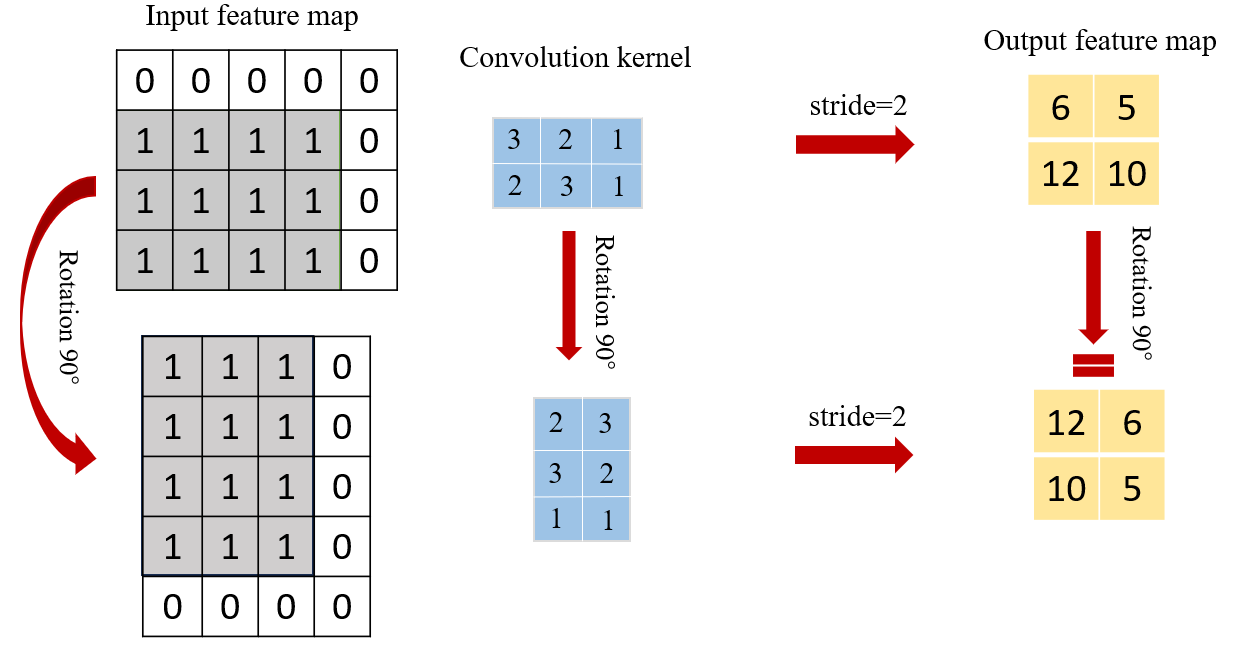}
		\caption{An example of the rotation equivariance of PreCM.}
		\label{fig_5}
	\end{figure}

	As mentioned earlier, to achieve rotation equivariance in networks, the convolution distributive law (Eq. (16)) must hold for images and convolutional kernels of arbitrary sizes. In terms of the previous studies, the design of equivariance is usually limited by convolution distributive law (as shown in Fig. (\ref{Fig_2})), thereby limiting their applications. However, for our proposed convolution mode, it solves this problem in a very convenient way. Let us illustrate this point with a specific example. Supposing $t=1$, then Eq. (16) can be written as:
	\begin{equation}
		\begin{aligned}
			{r_{{\sigma _1}}}\varphi { \otimes _{{\sigma _1}}}({r_{{\sigma _1}}}f)=
			{r_{{\sigma _1}}} (\varphi { \otimes _{{\sigma _0}}}f)
			={r_{{\sigma _1}}} (\varphi { \otimes [f \oplus {P_{{\sigma _0}}}])}.
		\end{aligned}
	\end{equation}
	From Eq. (22), researchers can calculate the required padding $P_{{\sigma _0}}$ under $\sigma_0$ mode. To be specific, when the image size ${w_{in}} \times {h_{in}}= 4\times3$, ${w_{out}} \times {h_{out}}= 2\times2$, convolution kernel size ${w_{\varphi}} \times {h_{\varphi}}= 3\times2$, and dilation ${w_{d}} \times {h_{d}}= 1\times1$, stride ${w_{s}} \times {h_{s}}= 2\times2$, we can easily get $P_{{\sigma _0}}= [1,0,0,1]$. With these values, once we change the convolution mode of Fig. \ref{Fig_2}, we can realize the convolution distributive law under any conditions, seeing Fig. \ref{fig_5}.
	
	Furthermore, in our network design, there are also a total of four distinct convolution modes, and their corresponding padding methods can be calculated using Eq. (21b). Specifically, when $P_{{\sigma _0}}= [1,0,0,1]$, we can get $P_{{\sigma _1}}= [1,1,0,0]$, $P_{{\sigma _2}}= [0,1,1,0]$, $P_{{\sigma _3}}= [0,0,1,1]$. To verify the feasibility and validity of the mentioned conclusions, we have constructed a simple three-layer network model in Appendix E as well as demonstrated specific application instances.
	
\begin{figure*}[ht!]
		\centering
		\includegraphics[width=7in]{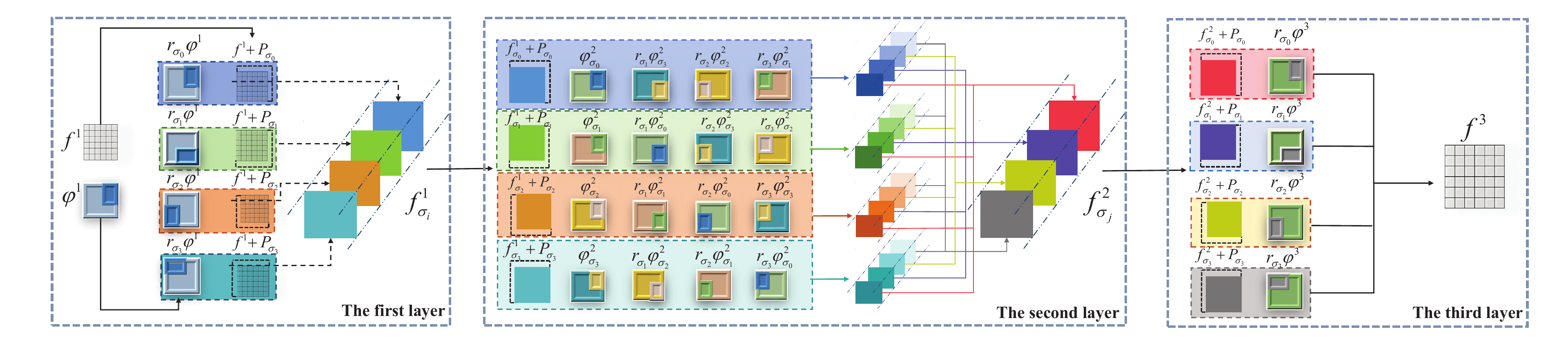}
		\captionsetup{justification=centering}
		\caption{The whole framework of PreCM.}
		\label{fig_6}
	\end{figure*}

	In conclusion, the proposed PreCM is regardless of the sizes and can be directly used as a replacement component to replace different types of convolution operations in existing networks. Therefore, to some extent, it can also be seen as a plug-and-play multi-scale rotation equivariant convolution. The whole framework of PreCM is displayed in Fig. \ref{fig_6}.
	
	\section{Experiments}
	In this section, we first introduce the training setup and the datasets Satellite Images of Water Bodies \cite{33}, DRIVE \cite{32}, and Floodnet \cite{31} that respectively represents the binary, small-sample, and multi-class segmentation tasks. Then, the evaluation metrics are introduced. Finally, we apply PreCM as a replacement component to different semantic segmentation networks for evaluating its performance. Simultaneously, we conducted a comprehensive comparison of the performance of PreCM-based networks, data augmentation-based networks, and specialized rotation equivariant networks using various evaluation metrics.
	
	\subsection{Training Setup and Datasets}
	1) \textit{Training Setup}:
	During the training phase, for all the networks, we uniformly adopted the Adam optimizer, with a batch size set to 4, a learning rate of 0.001, and a weight decay coefficient of 0.0005. Furthermore, during the training process, we employed Gaussian initialization for all networks, as this method enhances the stability, randomness, and efficiency of the training process of neural network, thereby facilitating more accurate convergence of the model to the global optimal solution. The only difference is that for the "Satellite Images of Water Bodies" and "Floodnet" datasets, we trained for 150 epochs, whereas for the "DRIVE" dataset, due to the limited information available from only 20 images used for training, we trained for 250 epochs to allow the model to more fully capture the data features and patterns. Besides, in order to reduce biases in the evaluation results, we put validation set and test set in one bucket and extract 70\% images randomly from this bucket as the test set for evaluating. This entire process is repeated five times, then the average values of results are recorded. All experiments were conducted on a desktop computer equipped with an NVIDIA GeForce RTX 4090 GPU and 64GB of RAM.
	
	2) \textit{Satellite Images of Water Bodies}:
	This dataset is a set of water body images taken by the Sentinel-2 satellite from any angle, which belongs to the category of binary semantic segmentation dataset. Each image is accompanied by a black and white mask, in which white represents water body and black represents something other than water. These masks are generated by calculating the normalized water difference index(NDWI)\cite{34}. This dataset contains 2328 images, of which 1,662 were used for training and 666 for testing and evaluating.
	
	3) \textit{DRIVE}:
	Within the binary semantic segmentation task, there also exist some other datasets, e.g., the medical image datasets that we often cannot obtain enough data for training. So, to test the performance of PreCM in the semantic segmentation of small-sample, the dataset DRIVE (Digital Retinal Images for Vessel Extraction) \cite{32} is chosen as the second dataset. It is attained from a diabetic retinopathy screening program in the Netherlands, composed of a total of JPEG 40 color fundus images. These images was equally divided into 20 images for training and 20 images for testing and evaluating.
	
	4) \textit{Floodnet}:
	In addition to binary semantic segmentation, multi-class semantic segmentation with relatively complex tasks is also one important branch in semantic segmentation family. To evaluate the performance of PreCM in this aspect, we here choose the dataset Floodnet \cite{31} as the final experimental data. It is divided into ten categories, including background, building-flooded, building-non-flooded, road-flooded, road-non-flooded, water, tree, vehicle, pool, grass, and contains 1445 images, of which 1120 were used for training and 325 for testing and evaluating.

	\subsection{Evaluation Metrics}
	In this paper, the metrics Intersection Over Union (IOU), Mean Intersection over Union (MIOU), and Dice Similarity Coefficient (DICE) are adopted to evaluate the segmentation results of networks \cite{35}, which are respectively defined as
	\begin{equation}
		\begin{aligned}
			&IOU = \frac{{TP}}{{TP + FN + FP}},\\[0.5mm]
			&MIOU = \frac{1}{2}(\frac{{TP}}{{TP + FN + FP}} + \frac{{TN}}{{TN + FN + FP}}),\\[0.5mm]
			&DICE = \frac{{2TP}}{{2TP + FP + FN}},\\[0.5mm]
		\end{aligned}
	\end{equation}
	where TP is the number of positive examples predicted correctly, FP means the number of positive examples prediction errors, TN represents  the number of negative examples predicted correctly, and FN denotes  the number of negative examples prediction errors.
	
	In addition, in practical application scenarios such as drone surveillance and satellite remote sensing imaging, image acquisition often encounters changes in perspective. Therefore, assessing the robustness of models against rotation becomes particularly important. For this goal, the RD metric is here designed for the first time to quantitatively assess the impact of image rotation, which can be described as
	\begin{equation}
		RD = \frac{{|{f_{{r_{{\sigma _i}}}}} - {f_{{r_{{\sigma _0}}}}}|}}{{{w_{out}} \times {h_{out}}}},
	\end{equation}
	where ${w_{out}},{h_{out}}$ represent the width and height of the output feature map respectively. $f_{{r_{{\sigma _i}}}}$ means the corresponding output feature map when the input image is rotated $r_{{\sigma _i}}$. This metric explicitly demonstrates the percentage of changed pixels relative to the total number of pixels in the output before and after the rotation operation, thereby revealing the sensitivity of model to image rotation. Specifically, a lower RD value indicates a stronger ability of model to resist rotation interference.

	\subsection{Quantitative Analysis}
	\begin{figure}[!t]
		\centering
		\includegraphics[width=3.5in]{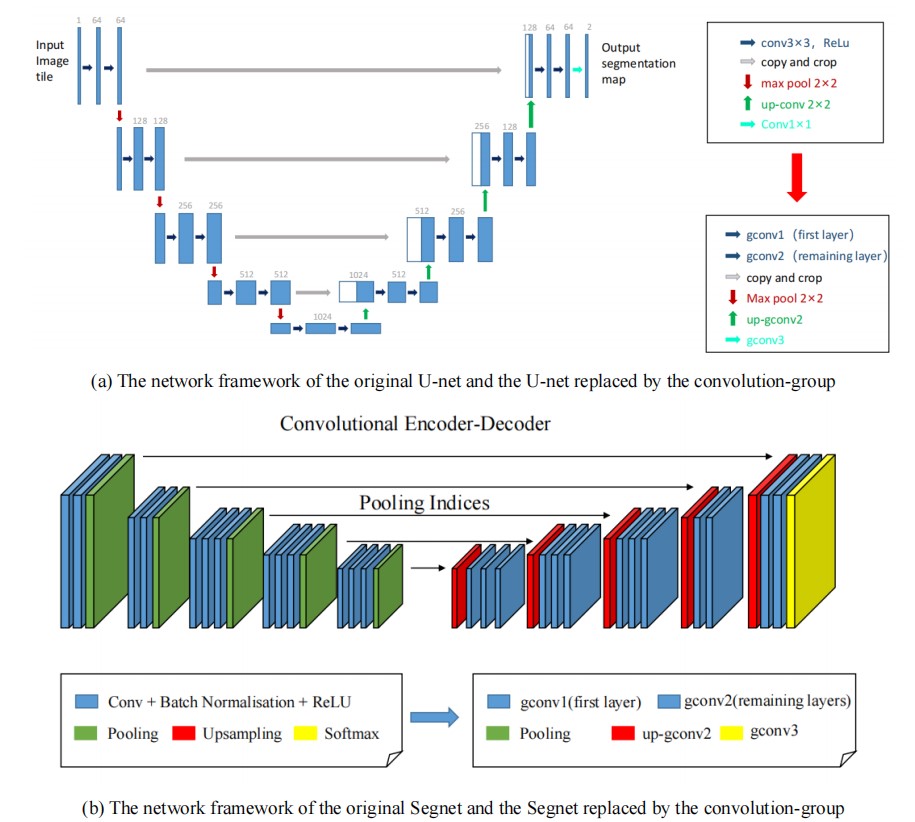}
		\caption{The replacement process of U-net with PreCM. PreCM1 is the first layer of PreCM, PreCM2 represents the second layer of PreCM, and PreCM3 denotes the third layer of PreCM.}
		\label{fig_7}
	\end{figure}
	
	\textbf{Experimental Method } Our experiment is conducted in three parts. Firstly, since our group theory is entirely built upon convolutional operations, we are able to replace the convolutional layers in rotation-sensitive networks to achieve rotation equivariance for these networks. In order to verify this point, we select six networks for substitution experiments, specifically three classical semantic segmentation networks: U-net \cite{12}, SegNet \cite{23}, PSPNet \cite{24}, and three lightweight semantic segmentation networks: Enet \cite{13}, Erfnet \cite{25}, BiSeNetV2 \cite{26}. It should be pointed out that, these networks are original networks without pre-training. Taking U-net as an example, the PreCM replacement process is illustrated in Fig. \ref{fig_7}. Specifically, we integrate PreCM into a rotation equivariant convolutional function, which comprises three parts: PreCM1, PreCM2, and PreCM3, corresponding to the three-layer structure in our method. During the replacement process, we substitute the first convolutional layer of U-net with PreCM1, the last convolutional layer with PreCM3, and the remaining convolutional layers with PreCM2, thereby transforming U-net into a new rotation equivariant U-net architecture. It is noteworthy that, although our theory only achieves zero-error rotation equivariance at the four angles 0$^\circ$, 90$^\circ$, 180$^\circ$, and 270$^\circ$, significant performance improvements can still be observed at other rotation angles, as demonstrated in the following experiments.
	Besides, during the specific training process, our proposed PreCM halves the number of channels. This is because our convolutional kernels are four times larger, and the number of parameters required for the convolutional kernels is proportional to the product of the input and output channels. Therefore, we halve the input and output channels of intermediate layers to ensure a consistent number of parameters. It is worth noting that this step is set only for the purpose of a fairer experimental comparison. In practice, this operation is not necessary, and there is no need to adjust the number of channels if sufficient computational resources are available.
	
	In the second part, given that data augmentation techniques can significantly enhance the network's ability to learn rotation features from multi-angle objects, we compare PreCM-based networks with data augmentation-based networks. Specifically, we expanded the dataset size by three times, covering rotation angles of {90$^\circ$}, {180$^\circ$}, and {270$^\circ$}. Thirdly, to thoroughly validate the performance of our method, we compare PreCM with three specialized rotation equivariant networks: RIC-CNN \cite{mo2024ric}, H-Net \cite{worrall2017harmonic}, and E2CNN \cite{weiler2019general}. Among them, RIC-CNN achieves rotation robustness by incorporating a coordinate-related design into the convolutional process; H-Net is a rotation equivariant network designed according to circular harmonics; and E2CNN is a network that achieves rotation and reflection equivariance based on group theory. Specifically, we use the codes and protocols provided by their authors and integrate these rotation equivariant methods into the Encoder-Decoder architecture with U-net as the backbone, thereby ensuring the parameters remain basically consistent.
	
	\begin{table*}[!h]
		\renewcommand\arraystretch{1.4}
		\begin{center}
			\caption{Evaluation results(\%) of original networks and PreCM replacement networks on Satellite Images of Water Bodies, FloodNet datasets and DRIVE.}
			\begin{tabular}{|c|c|c|c|c|c|c|c|c|c|c|c|c|c|} 
				\hline
				\textbf{\multirow{2}*{Network}} & \textbf{\multirow{2}*{Rotation}} & \multicolumn{4}{c|}{\textbf{Satellite Images of Water Bodies}} &\multicolumn{4}{c|}{\textbf{DRIVE}} & \multicolumn{4}{c|}{\textbf{Floodnet}}
				\\ \cline{3-14}
				~& &\textbf{IOU} & \textbf{MIOU} & \textbf{DICE} & \textbf{RD}&\textbf{IOU} & \textbf{MIOU} & \textbf{DICE} & \textbf{RD} &\textbf{IOU} & \textbf{MIOU} & \textbf{DICE} & \textbf{RD}
				\\
				\hline
				\multirow{5}*{U-net} & 0$^\circ$ &80.23  &86.54 &88.65 &\diagbox[height=1.6em,width=3em]{}{}& 65.97 &81.16 &79.69&\diagbox[height=1.6em,width=3em]{}{}&76.94 & 87.20 & 85.28 &\diagbox[height=1.6em,width=3em]{}{}
				\\ \cline{2-14}
				~&90$^\circ$ &79.39 &85.12 &87.64 &3.89 & 63.84 &79.86 &77.68 &2.18& 72.19 & 83.09 & 81.54 & 14.09
				\\ \cline{2-14}
				~& 180$^\circ$ &79.67 &86.01 &87.86 &3.46 &63.91 &80.04 &77.92 &2.11& 73.86 & 84.32 & 83.10 &12.74
				\\ \cline{2-14}
				~& 270$^\circ$ &79.25 &85.32 &87.56 &3.94 &63.54 &79.82 &77.71 &2.16 & 71.92 & 83.41 & 81.36 &14.53
				\\ \cline{2-14}
				~&random &76.34 &83.39 &84.42 & 4.17 &63.16 &79.38 &77.17 &2.82 &71.02 &83.61  &80.85  &15.24
				\\ \cline{2-14}
				\hline
				\multirow{2}*{U-net(PreCM)}&0$^\circ$/90$^\circ$/180$^\circ$/270$^\circ$&\pmb{86.66}&\pmb{90.79}&	\pmb{92.27}&	\pmb{0.00} &\pmb{69.72} &\pmb{83.65} &\pmb{82.11} &\pmb{0.00}& \pmb{79.06} & \pmb{88.52} & \pmb{87.07} &\pmb{0.00}
				\\ \cline{2-14}
				~&random & \pmb{85.19} & \pmb{89.61} &\pmb{91.32} & \pmb{1.62} &\pmb{69.22} &\pmb{82.91} & \pmb{81.78}&\pmb{1.46} &\pmb{76.85} &\pmb{86.87} &\pmb{84.64} &  \pmb{8.41}
				\\
				\hline
				\multirow{5}*{SegNet} & 0$^\circ$ &76.99 &84.56&	85.92&\diagbox[height=1.6em,width=3em]{}{}& 49.93 &72.57 &66.80 &\diagbox[height=1.6em,width=3em]{}{}& 68.73 & 82.00 & 79.86 & \diagbox[height=1.6em,width=3em]{}{}
				\\ \cline{2-14}
				~&90$^\circ$ &75.95&83.88&	84.89&	3.96& 46.26&70.58 &63.23 &3.25 & 62.93 & 78.79 & 74.61 & 21.70
				\\ \cline{2-14}
				~& 180$^\circ$ &76.37&	84.12&	85.08&	3.63&47.13 &70.92 &64.18 &3.13& 66.09 & 80.57 & 77.02 &19.52
				\\ \cline{2-14}
				~& 270$^\circ$ &75.74&	83.80&	84.92&	3.89&46.05 &70.91 &64.03 &3.28& 62.74 & 78.69 & 74.75 &21.86
				\\ \cline{2-14}
				~&random &74.64 &82.30 &83.47 & 5.11 &45.47 &70.14 &62.06 &3.90 & 62.97 &79.02  &75.19  &21.65
				\\ \cline{2-14}
				\hline
				\multirow{2}*{SegNet(PreCM)}&0$^\circ$/90$^\circ$/180$^\circ$/270$^\circ$&\pmb{84.34}&\pmb{89.38}&	\pmb{91.19}&	\pmb{0.00}& \pmb{62.63}&\pmb{79.38} &\pmb{77.01} &\pmb{0.00}& \pmb{74.29} & \pmb{85.26} & \pmb{83.39}& \pmb{0.00}
				\\ \cline{2-14}
				~&random & \pmb{82.41} & \pmb{87.83} &\pmb{89.46} & \pmb{2.30} & \pmb{61.33} &\pmb{78.64} & \pmb{75.98}& \pmb{2.38} &\pmb{71.24} &\pmb{83.01} &\pmb{81.24} & \pmb{12.29}
				\\
				\hline
				\multirow{5}*{PSPNet} & 0$^\circ$ &71.56 &81.00  &81.98 &\diagbox[height=1.6em,width=3em]{}{}&36.39 &64.70 &53.38 &\diagbox[height=1.6em,width=3em]{}{}&76.82 &86.74 &84.91 &\diagbox[height=1.6em,width=3em]{}{}
				\\ \cline{2-14}
				~&90$^\circ$  &71.09 &80.38 &81.50 &2.53  &35.01 &64.04 &52.07 &2.30&71.91 &83.91 &81.59 &15.18
				\\ \cline{2-14}
				~& 180$^\circ$ &71.54 &81.27 &81.74 &2.28  &35.28 &64.29 &52.18 &2.24 & 73.87 &85.02 &82.30 &14.08
				\\ \cline{2-14}
				~& 270$^\circ$ &71.17 &80.39 &81.46 &2.48  &35.30 &64.09 &52.11 &2.39 & 72.08 &84.07 &81.95 &15.02
				\\ \cline{2-14}
				~&random &74.87 &82.54 &83.82 & 3.92 & 35.05 &64.13 &52.15 & 4.19 &70.93 &83.01  &80.72  &15.14
				\\ \cline{2-14}
				\hline
				\multirow{2}*{PSPNet(PreCM)}&0$^\circ$/90$^\circ$/180$^\circ$/270$^\circ$&\pmb{78.23} &\pmb{85.19} &\pmb{86.65} &\pmb{0.00}	 &\pmb{40.58} &\pmb{66.87} &\pmb{57.30} &\pmb{0.00}&\pmb{77.34} &\pmb{87.07} &\pmb{85.19}  &\pmb{0.00}
				\\ \cline{2-14}
				~&random & \pmb{80.16} & \pmb{85.86} &\pmb{87.45} & \pmb{2.18} & \pmb{39.22}& \pmb{65.92} &\pmb{55.47} & \pmb{2.14} &\pmb{75.06} &\pmb{85.20} &\pmb{82.96} &\pmb{8.53}
				\\
				\hline
				\multirow{5}*{ENet} & 0$^\circ$ &83.48&	89.09&	90.42&\diagbox[height=1.6em,width=3em]{}{}&59.93 &77.66 &74.80 &\diagbox[height=1.6em,width=3em]{}{}& 78.94 & 87.95 & 86.87 & \diagbox[height=1.6em,width=3em]{}{}
				\\ \cline{2-14}
				~&90$^\circ$ &82.52&	88.29&	89.77&	2.79 &57.91 &76.48 &72.98 &3.06& 73.88 & 85.02 & 82.85 & 14.13
				\\ \cline{2-14}
				~& 180$^\circ$ &82.84&	88.49&	90.11&	2.67 &58.12 &76.87 &73.55 &2.95& 75.05 & 85.68 & 83.89 &13.28
				\\ \cline{2-14}
				~& 270$^\circ$ &82.49 &	88.13&	89.74&	2.82&57.32 &76.54 &72.94 &3.02& 74.08 & 85.27 & 83.09 &13.81
				\\ \cline{2-14}
				~&random &79.58 &86.04 &87.71 & 4.26 & 58.11 & 76.70 & 73.45 & 3.86 & 73.15& 84.67 &82.24  &14.54
				\\
				\hline
				\multirow{2}*{ENet(PreCM)}&0$^\circ$/90$^\circ$/180$^\circ$/270$^\circ$&\pmb{88.23}&\pmb{92.24}&	\pmb{93.12}&\pmb{0.00}&\pmb{64.76} &\pmb{80.66} &\pmb{78.58} &\pmb{0.00}& \pmb{80.38} & \pmb{88.91} & \pmb{87.37} &\pmb{0.00}
				\\ \cline{2-14}
				~&random & \pmb{87.00} & \pmb{91.26} &\pmb{92.52} & \pmb{1.58} & \pmb{63.35} & \pmb{79.64} & \pmb{77.53} & \pmb{2.05} &\pmb{78.29} &\pmb{86.87} &\pmb{85.36} &\pmb{8.06}
				\\
				\hline
				\multirow{5}*{ERFNet} & 0$^\circ$ &82.97&	88.35&	90.01&\diagbox[height=1.6em,width=3em]{}{}& 55.30 &75.29 &71.12 &\diagbox[height=1.6em,width=3em]{}{}& 79.18 &88.09  & 86.98 & \diagbox[height=1.6em,width=3em]{}{}
				\\ \cline{2-14}
				~&90$^\circ$ &81.51&	87.38&	88.92&	3.03  & 52.48&73.60 &68.74 &3.09& 72.90 & 84.13 & 82.20 &14.68
				\\ \cline{2-14}
				~& 180$^\circ$ &82.55&	88.39&	89.66&	2.81&54.02 &74.36 &70.37 &2.93& 76.27 & 86.38 & 84.51 &12.09
				\\ \cline{2-14}
				~& 270$^\circ$ &81.88&	87.76&	89.30&	2.98&52.59 &73.62 &68.88 &3.05& 73.11 & 84.58 & 82.35 &14.16
				\\ \cline{2-14}
				~&random &80.79 &86.39 &87.99 & 3.23 & 53.09 &73.89 &69.28 &3.26 & 72.75& 84.45 &82.05  &14.52
				\\
				\hline
				\multirow{2}*{ERFNet(PreCM)}&0$^\circ$/90$^\circ$/180$^\circ$/270$^\circ$&\pmb{86.82}&\pmb{91.24}&	\pmb{92.30}&	\pmb{0.00} &\pmb{65.57} &\pmb{81.01} &\pmb{79.32} &\pmb{0.00}& \pmb{81.02} & \pmb{89.35} & \pmb{88.06} &\pmb{0.00}
				\\ \cline{2-14}
				~&random & \pmb{86.05} & \pmb{90.66} &\pmb{91.93} & \pmb{1.30} & \pmb{64.58} & \pmb{80.33} & \pmb{78.45} & \pmb{2.16} &\pmb{78.45} &\pmb{87.74} &\pmb{86.19} &\pmb{7.15}
				\\
				\hline
				\multirow{5}*{BiSeNetV2} & 0$^\circ$ & 79.11 & 86.09  & 87.38 & \diagbox[height=1.6em,width=3em]{}{} &35.28 &63.99 &52.06 &\diagbox[height=1.6em,width=3em]{}{}& 76.90 & 86.83 & 85.32 & \diagbox[height=1.6em,width=3em]{}{}
				\\ \cline{2-14}
				~&90$^\circ$ & 78.05 & 85.33 & 86.65 &	3.19  &30.17 &60.38 &46.49 &2.61& 69.96 & 82.02 & 79.97 & 16.14
				\\ \cline{2-14}
				~& 180$^\circ$ & 77.70 & 85.14 & 86.46 & 3.01 & 32.06&61.46 &48.60 &2.51& 71.86 & 83.18 & 81.32 &14.43
				\\ \cline{2-14}
				~& 270$^\circ$ & 78.15 & 85.34 & 86.69 & 3.13 &29.97 &60.42 &45.85 &2.58 & 70.39 & 81.88 & 80.26 &16.11
				\\ \cline{2-14}
				~&random &74.36 &81.23 &83.41 & 3.94 &28.91 &60.01 &42.73 & 3.53 &69.00 &82.29  &79.39  &16.95
				\\
				\hline
				\multirow{2}*{BiSeNetV2(PreCM)}&0$^\circ$/90$^\circ$/180$^\circ$/270$^\circ$&\pmb{82.74}&\pmb{88.35}&	\pmb{89.93}&	\pmb{0.00}&\pmb{40.66} &\pmb{66.90} &\pmb{57.64} &\pmb{0.00}& \pmb{79.99} & \pmb{88.45} & \pmb{87.32} &\pmb{0.00}
				\\ \cline{2-14}
				~&random & \pmb{80.72} & \pmb{87.07} &\pmb{88.26} & \pmb{2.51} & \pmb{39.20} & \pmb{65.97} & \pmb{56.24} & \pmb{2.17} &\pmb{77.34} &\pmb{86.82} &\pmb{85.61} & \pmb{9.45}
				\\
				\hline
			\end{tabular}
		\end{center}
	\end{table*}
	
	\begin{table*}[!h]
		\renewcommand\arraystretch{1.3}
		\begin{center}
			\caption{Comparison Results(\%) of Data Augmentation and PreCM Replacement.}
			\begin{tabular}{|c|c|c|c|c|c|c|c|c|c|c|c|c|} 
				\hline
				\textbf{\multirow{2}*{Network}}& \multicolumn{4}{c|}{\textbf{Satellite Images of Water Bodies}} &\multicolumn{4}{c|}{\textbf{DRIVE}} & \multicolumn{4}{c|}{\textbf{Floodnet}}
				\\ \cline{2-13}
				~&\textbf{IOU} & \textbf{MIOU} & \textbf{DICE} & \textbf{RD}&\textbf{IOU} & \textbf{MIOU} & \textbf{DICE} & \textbf{RD} &\textbf{IOU} & \textbf{MIOU} & \textbf{DICE} & \textbf{RD}
				\\
				\hline
				U-net+aug &84.22  &89.15  &90.63 & 2.64   & 69.05 &82.71 &81.62 &1.71 &75.51  &86.06 &84.26 & 9.33
				\\
				\hline
				U-net(PreCM) &\pmb{85.19}&\pmb{89.61}&	\pmb{91.32}&	\pmb{1.62} & \pmb{69.22} &\pmb{82.91} &\pmb{81.78} &\pmb{1.45} & \pmb{76.85} & \pmb{86.87} & \pmb{84.64} &\pmb{8.41}
				\\
				\hline
				SegNet+aug & 82.23 & 87.35 & 89.35 & 2.85  & 56.38& 75.88&71.88 &3.53 & \pmb{71.29} &\pmb{83.20}  &\pmb{81.31}  & 13.25
				\\
				\hline
				SegNet(PreCM)&\pmb{82.41}&\pmb{87.83}&	\pmb{89.46}&	\pmb{2.30}& \pmb{61.33}&\pmb{78.64} &\pmb{75.98} &\pmb{2.37} & 71.24 & 83.01 & 81.24& \pmb{12.29}
				\\
				\hline
				PSPNet+aug &\pmb{80.84}  & \pmb{86.15} & \pmb{87.75} &  2.34 & 35.96 &64.78 &52.83 &3.42 & \pmb{75.65} &\pmb{85.28} &\pmb{83.57} & 9.13
				\\
				\hline
				PSPNet(PreCM)&80.16 &85.86 &87.45 &\pmb{2.17}	 &\pmb{39.22} &\pmb{65.92} &\pmb{55.47} &\pmb{2.14} & 75.06&85.20 &82.96  &\pmb{8.53}
				\\
				\hline
				ENet+aug & 86.43 & 90.93 & 92.17& 1.83 &62.68 &79.24 &77.01 &3.06 & 77.68 &86.29 &85.09 &8.78
				\\
				\hline
				ENet(PreCM)&\pmb{87.00}&\pmb{91.26}&	\pmb{92.52}&\pmb{1.58}&\pmb{63.35} &\pmb{79.64} &\pmb{77.53} &\pmb{2.05}& \pmb{78.29} & \pmb{86.87} & \pmb{85.36} &\pmb{8.06}
				\\
				\hline
				ERFNet+aug &85.54 & 90.22 &91.56 & 1.89  &63.10 &79.49 &77.34 &3.11 &76.81  &86.77 &84.72 &9.24
				\\
				\hline
				ERFNet(PreCM)&\pmb{86.05}&\pmb{90.66}&	\pmb{91.93}&	\pmb{1.30} &\pmb{64.58} &\pmb{80.33} &\pmb{78.45} &\pmb{2.16}& \pmb{78.45} & \pmb{87.74} & \pmb{86.19} &\pmb{7.15}
				\\
				\hline
				BiSeNetV2+aug & 79.40 & 86.27 & 87.24 &  3.56  &37.81&65.41 &54.81 & 4.98& 76.15  &85.42 &84.57 &10.04
				\\
				\hline
				BiSeNetV2(PreCM)&\pmb{80.72}&\pmb{87.07}&	\pmb{88.26}&	\pmb{2.51}&\pmb{39.20} &\pmb{65.97} &\pmb{56.24} &\pmb{2.17}& \pmb{77.34} & \pmb{86.82} & \pmb{85.61} &\pmb{9.45}
				\\
				\hline
			\end{tabular}
		\end{center}
	\end{table*}

	\begin{table*}[!h]
		\renewcommand\arraystretch{1.4}
		\begin{center}
			\caption{Evaluation results(\%) of rotation equivariant networks on Satellite Images of Water Bodies, FloodNet datasets and DRIVE. Data for 0$^\circ$/90$^\circ$/180$^\circ$/270$^\circ$ are expressed as mean±standard deviation.}
			\begin{tabular}{|c|c|c|c|c|c|c|c|} 
				\hline
				\textbf{\multirow{2}*{Network}} & \textbf{\multirow{2}*{Rotation}} & \multicolumn{2}{c|}{\textbf{Satellite Images of Water Bodies}} &\multicolumn{2}{c|}{\textbf{DRIVE}} & \multicolumn{2}{c|}{\textbf{Floodnet}}
				\\ \cline{3-8}
				~& &\textbf{IOU} & \textbf{RD} & \textbf{IOU} &  \textbf{RD} &\textbf{IOU} & \textbf{RD}
				\\
				\hline
				\multirow{2}*{RIC-CNN}&0$^\circ$/90$^\circ$/180$^\circ$/270$^\circ$&82.45$\pm$0.14& 1.36$\pm$0.23&	66.86$\pm$0.03&1.25$\pm$0.16	& 78.15$\pm$0.01&2.44$\pm$0.48
				\\ \cline{2-8}
				~&random & 78.85 & 2.68 & 67.07 & 2.03 & 76.46 & 8.93
				\\
				\hline
				\multirow{2}*{H-Net}&0$^\circ$/90$^\circ$/180$^\circ$/270$^\circ$&81.68$\pm$0.47 & 1.42$\pm$0.29 &65.94$\pm$0.17 &1.37$\pm$0.21 &77.21$\pm$0.10 & 3.64$\pm$0.62
				\\ \cline{2-8}
				~&random & 76.84 & 3.51 & 64.48 & 2.19 &73.41 & 10.28
				\\
				\hline
				\multirow{2}*{E2CNN}&0$^\circ$/90$^\circ$/180$^\circ$/270$^\circ$&85.38$\pm$0.54 &0.94$\pm$0.19 &66.81$\pm$0.02	& 1.19$\pm$0.20 & 77.97$\pm$0.02& 3.20$\pm$0.80
				\\ \cline{2-8}
				~&random &81.59  & 3.35 & 67.35 & 2.08 &74.33 &10.86
				\\
				\hline
				\multirow{2}*{U-net(PreCM)}&0$^\circ$/90$^\circ$/180$^\circ$/270$^\circ$&\pmb{86.66$\pm$0.00}&\pmb{0.00$\pm$0.00}&	\pmb{69.72$\pm$0.00}&	\pmb{0.00$\pm$0.00} &\pmb{79.06$\pm$0.00} &\pmb{0.00$\pm$0.00}
				\\ \cline{2-8}
				~&random & \pmb{85.19} & \pmb{1.62} &\pmb{69.22} & \pmb{1.46} &\pmb{76.85} &\pmb{8.41}
				\\
				\hline
			\end{tabular}
		\end{center}
	\end{table*}

	\begin{figure*}[!ht]
		\footnotesize
		\centering
		\captionsetup[subfloat]{position=bottom,labelformat=empty}	
		\subfloat[\footnotesize ]{\includegraphics[height=1.4cm,width=1.3cm]{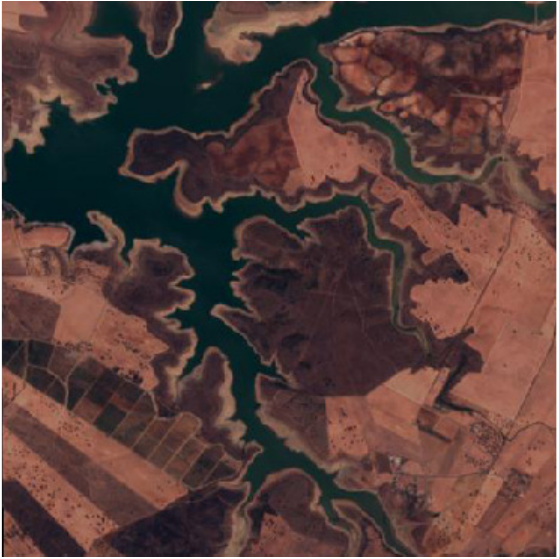}}\hspace{0.05cm}
		\subfloat[\footnotesize ]{\includegraphics[height=1.4cm,width=1.3cm]{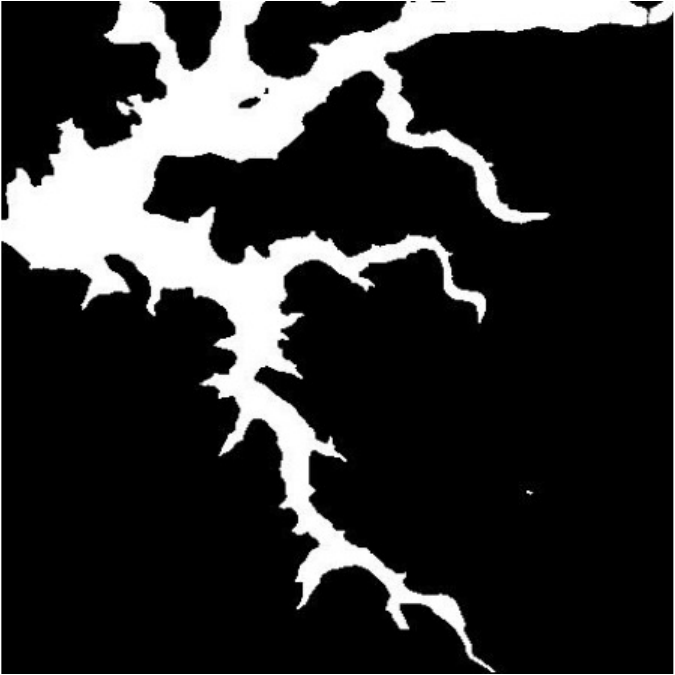}}\hspace{0.05cm}
		\subfloat[\footnotesize ]{\includegraphics[height=1.4cm,width=1.3cm]{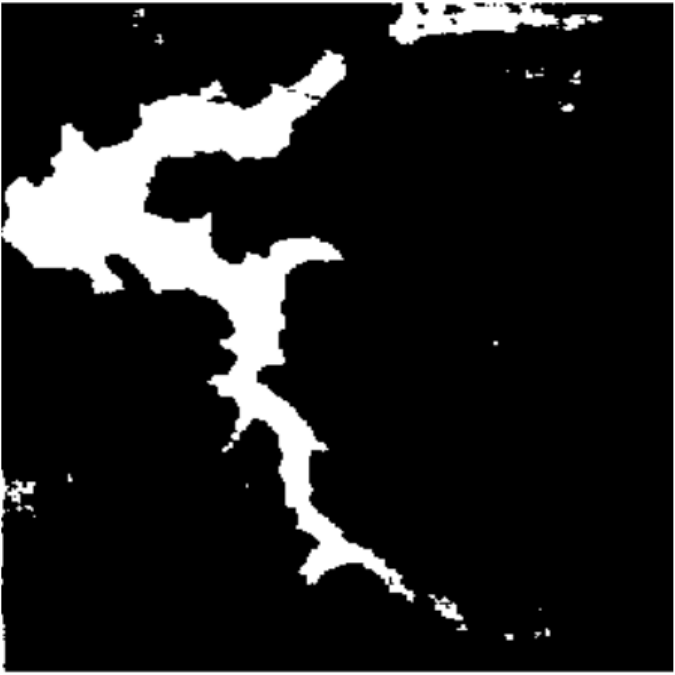}}\hspace{0.05cm}
		\subfloat[\footnotesize ]{\includegraphics[height=1.4cm,width=1.3cm]{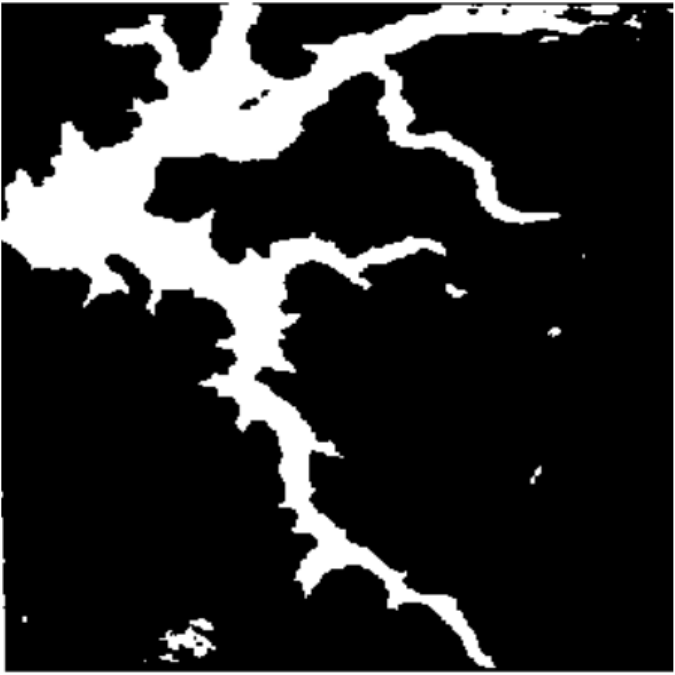}}\hspace{0.05cm}
		\subfloat[\footnotesize ]{\includegraphics[height=1.4cm,width=1.3cm]{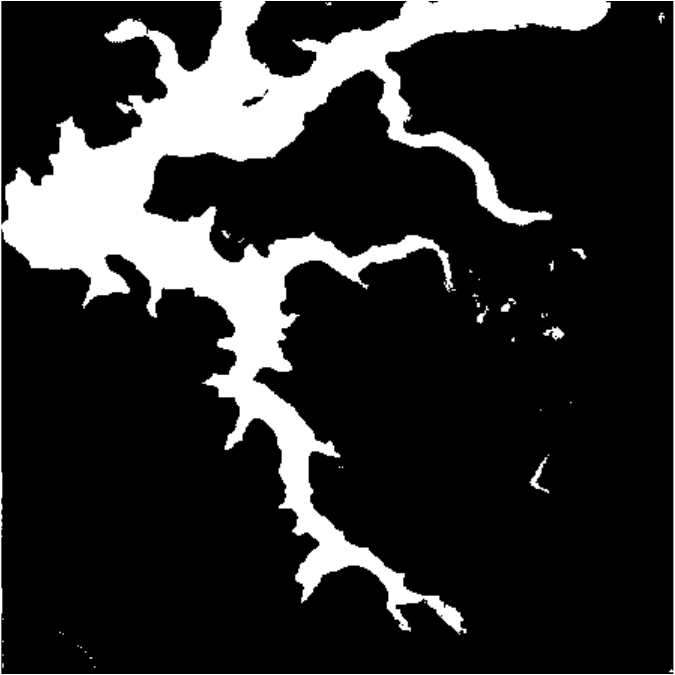}}\hspace{0.05cm}
		\subfloat[\footnotesize ]{\includegraphics[height=1.4cm,width=1.3cm]{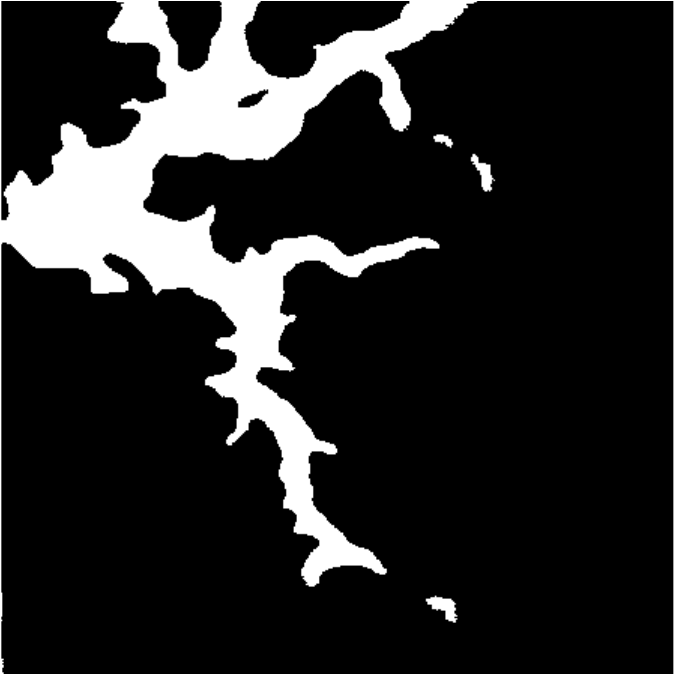}}\hspace{0.05cm}
		\subfloat[\footnotesize ]{\includegraphics[height=1.4cm,width=1.3cm]{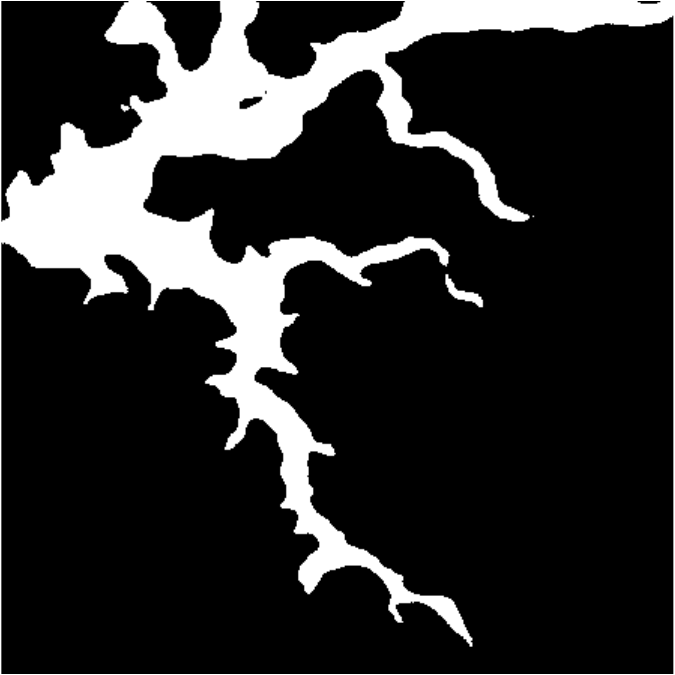}}\hspace{0.05cm}
		\subfloat[\footnotesize ]{\includegraphics[height=1.4cm,width=1.3cm]{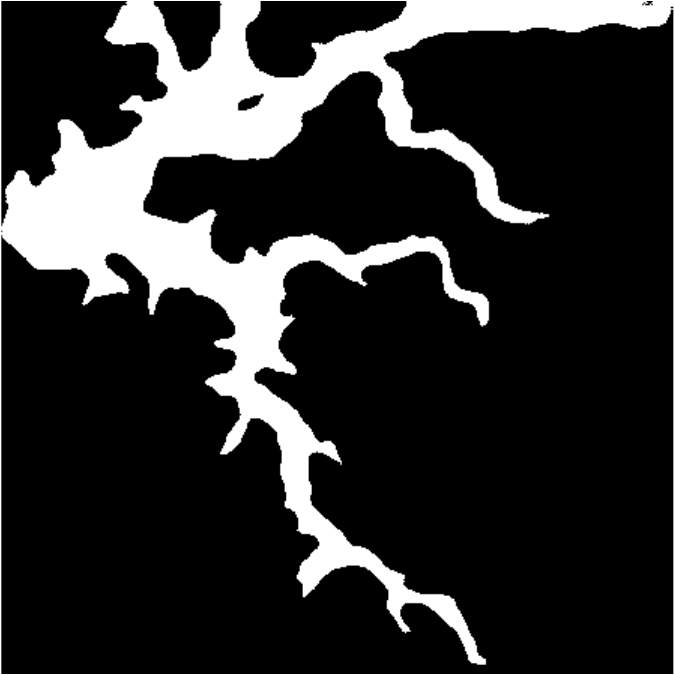}}\hspace{0.05cm}
		\subfloat[\footnotesize ]{\includegraphics[height=1.4cm,width=1.3cm]{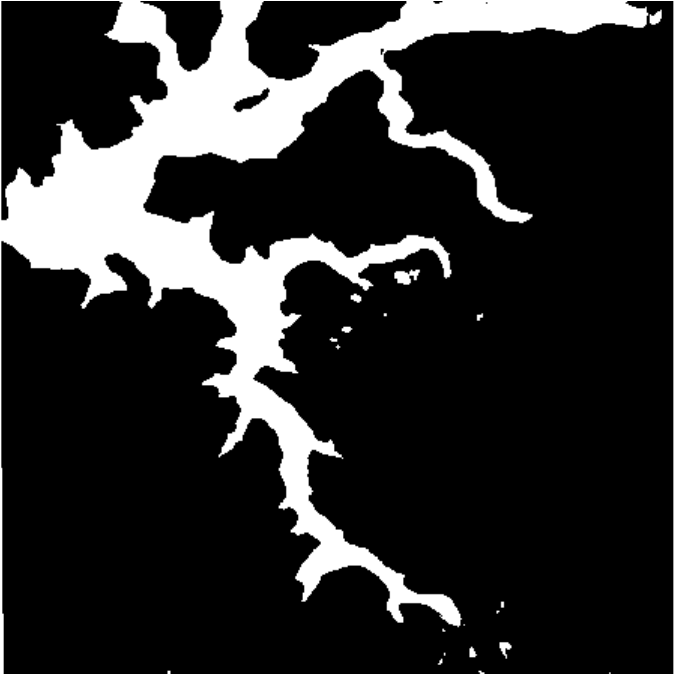}}\hspace{0.05cm}
		\subfloat[\footnotesize ]{\includegraphics[height=1.4cm,width=1.3cm]{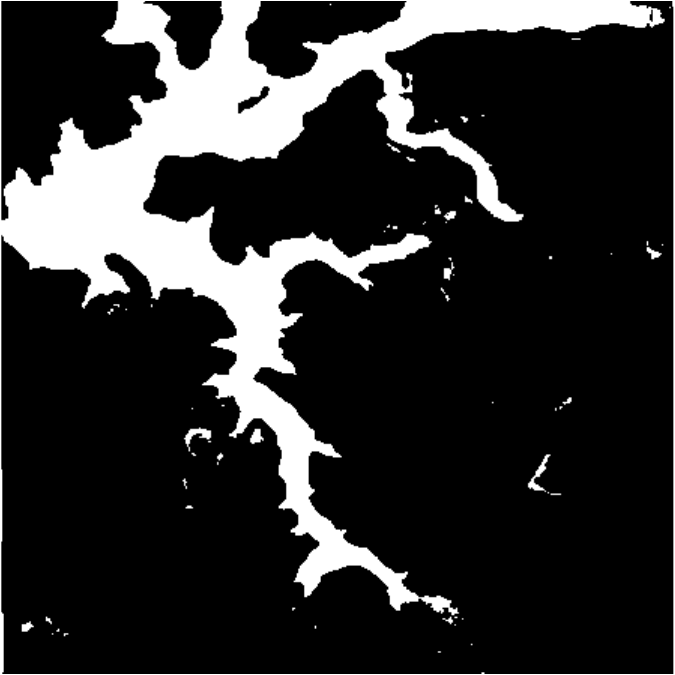}}\hspace{0.05cm}
		\subfloat[\footnotesize ]{\includegraphics[height=1.4cm,width=1.3cm]{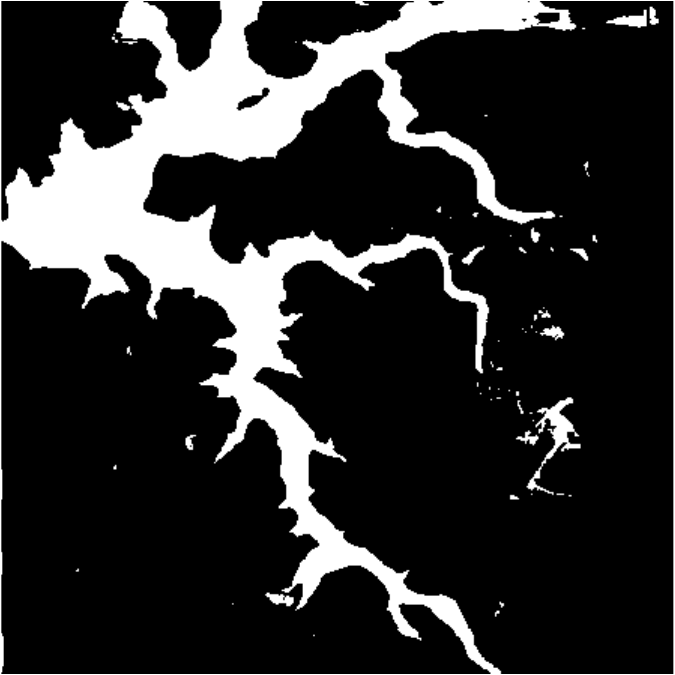}}\hspace{0.05cm}\\
		\vspace{-0.7cm}
		\centering
		\subfloat[\footnotesize ]{\includegraphics[height=1.4cm,width=1.3cm]{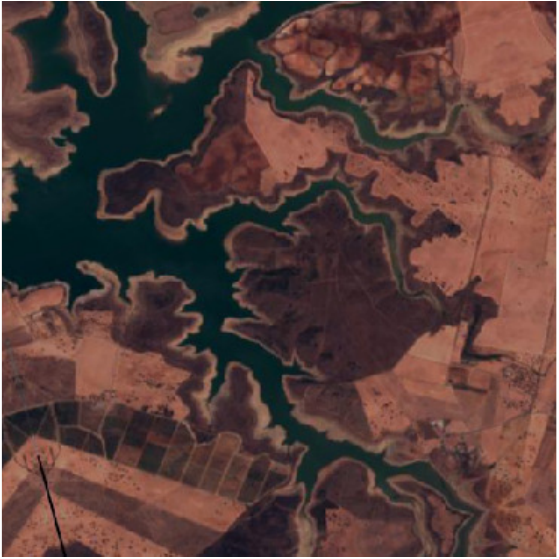}}\hspace{0.05cm}
		\subfloat[\footnotesize ]{\includegraphics[height=1.4cm,width=1.3cm]{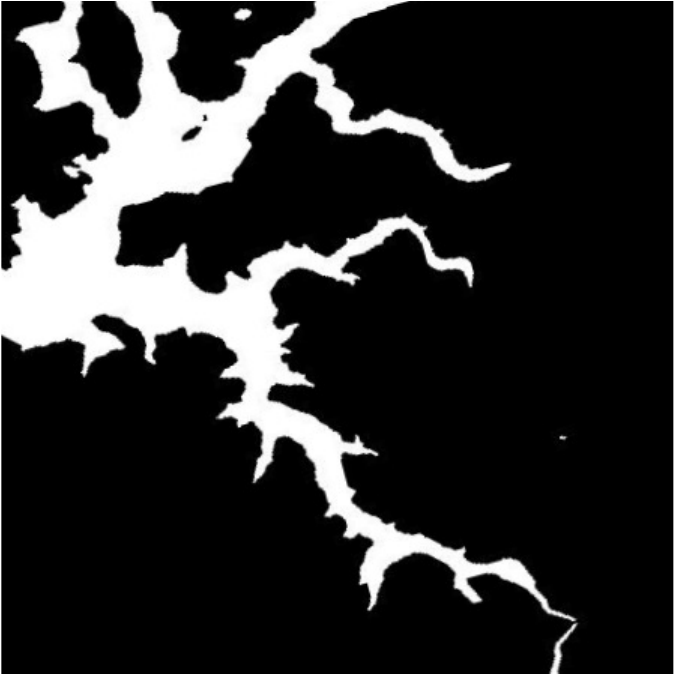}}\hspace{0.05cm}
		\subfloat[\footnotesize ]{\includegraphics[height=1.4cm,width=1.3cm]{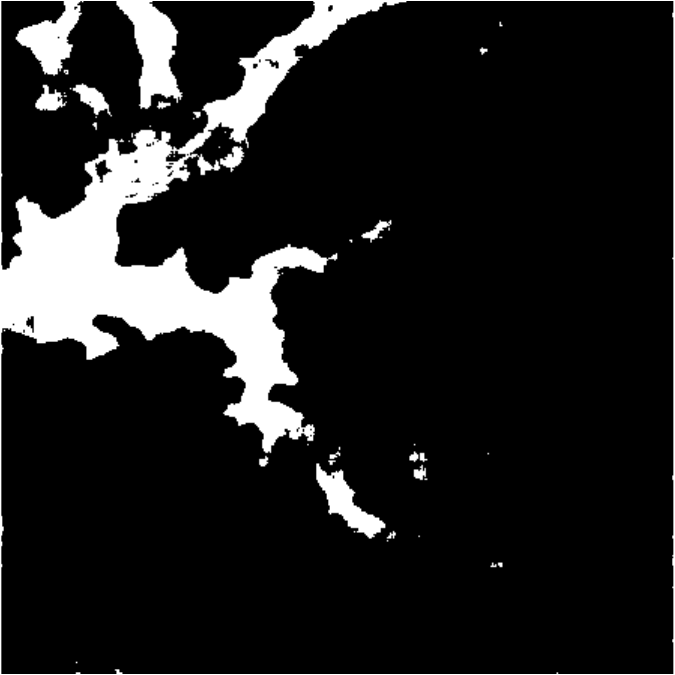}}\hspace{0.05cm}
		\subfloat[\footnotesize ]{\includegraphics[height=1.4cm,width=1.3cm]{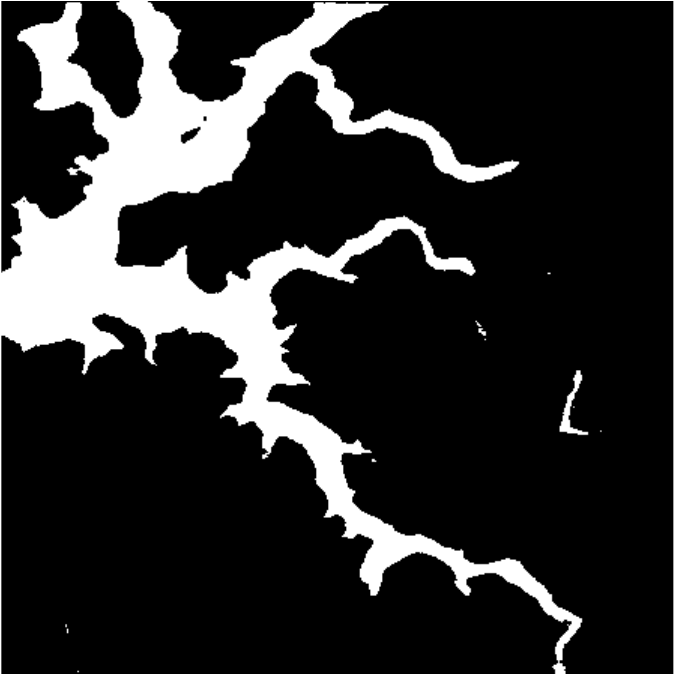}}\hspace{0.05cm}
		\subfloat[\footnotesize ]{\includegraphics[height=1.4cm,width=1.3cm]{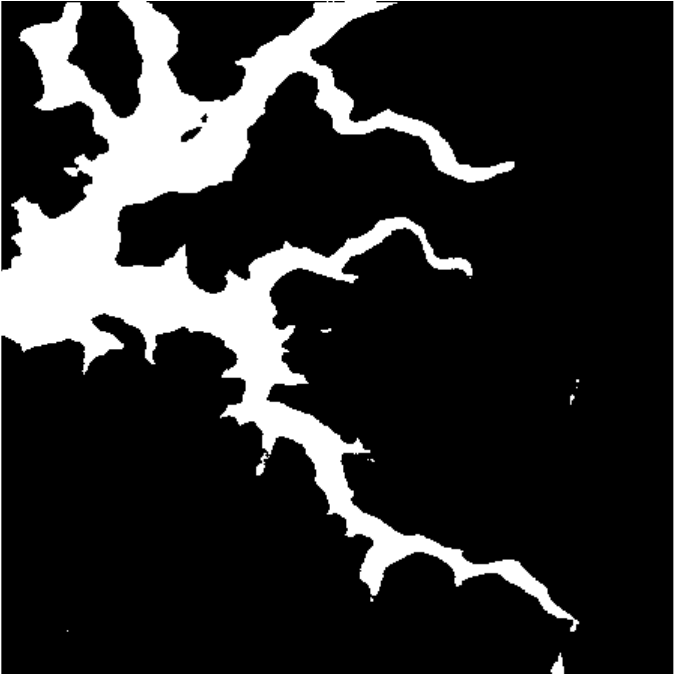}}\hspace{0.05cm}
		\subfloat[\footnotesize ]{\includegraphics[height=1.4cm,width=1.3cm]{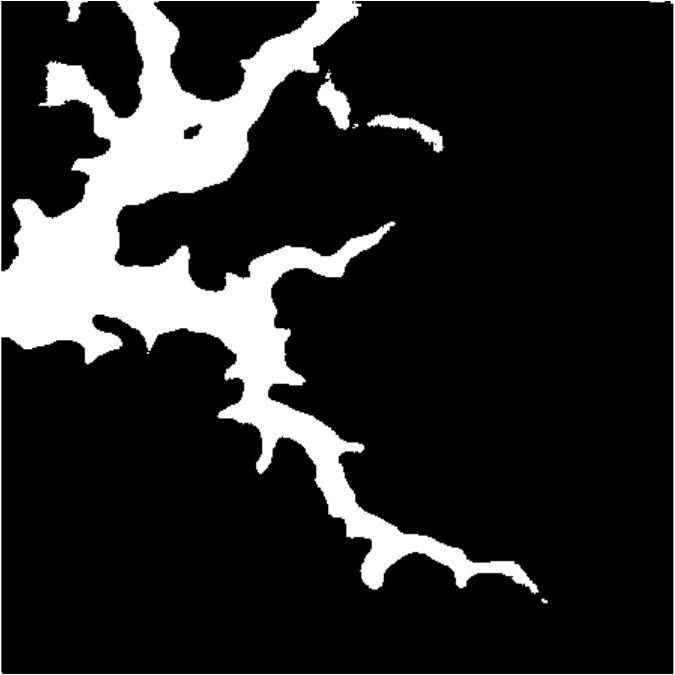}}\hspace{0.05cm}
		\subfloat[\footnotesize ]{\includegraphics[height=1.4cm,width=1.3cm]{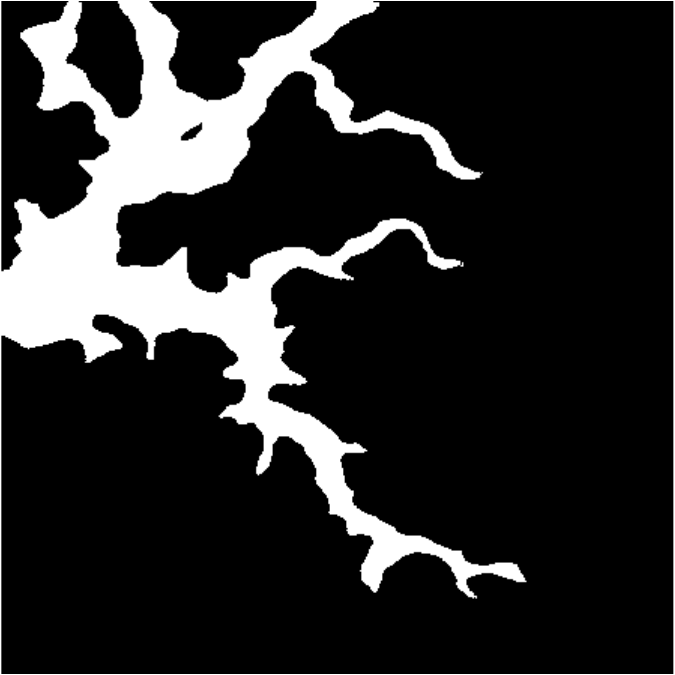}}\hspace{0.05cm}
		\subfloat[\footnotesize ]{\includegraphics[height=1.4cm,width=1.3cm]{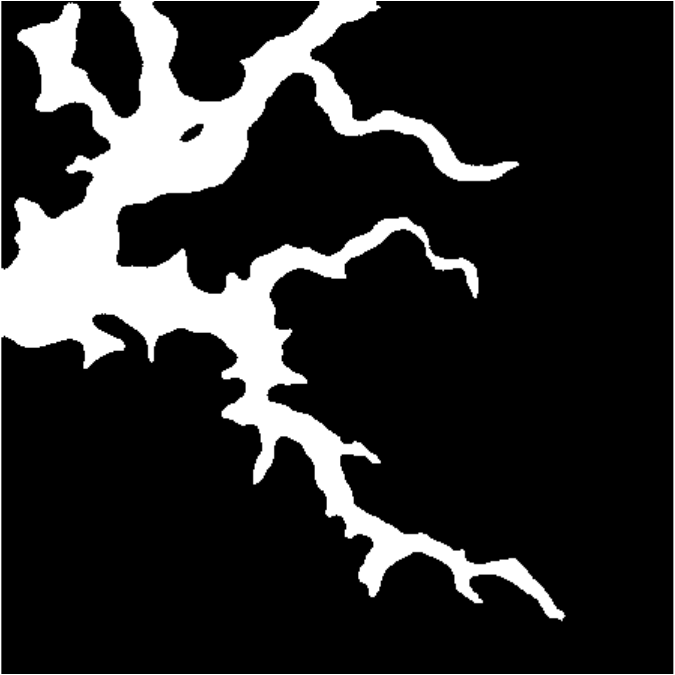}}\hspace{0.05cm}
		\subfloat[\footnotesize ]{\includegraphics[height=1.4cm,width=1.3cm]{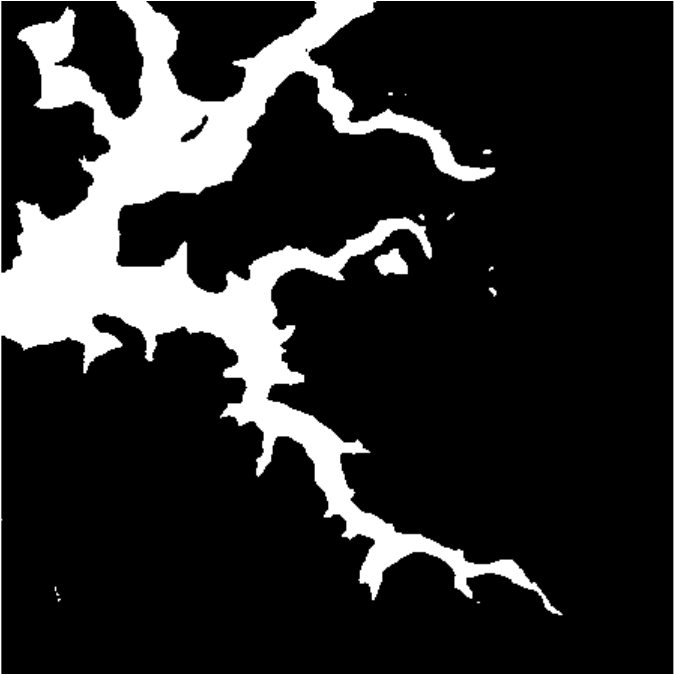}}\hspace{0.05cm}
		\subfloat[\footnotesize ]{\includegraphics[height=1.4cm,width=1.3cm]{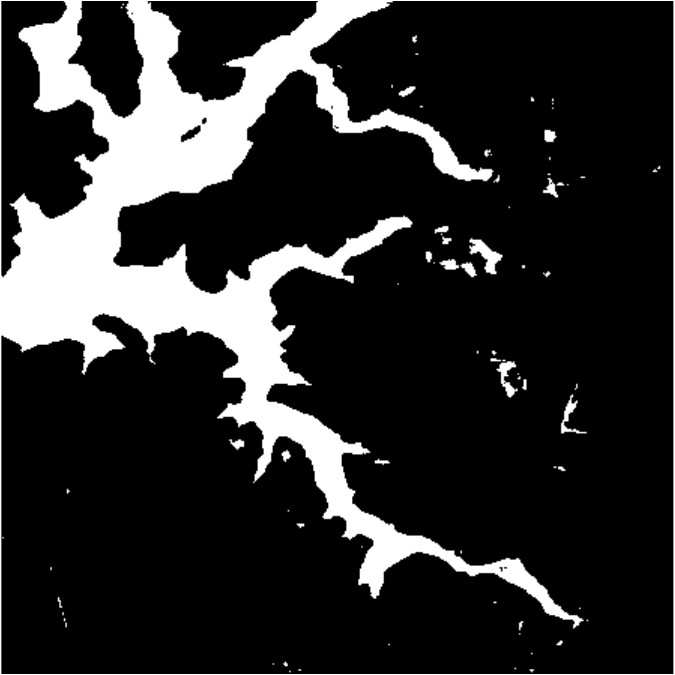}}\hspace{0.05cm}
		\subfloat[\footnotesize ]{\includegraphics[height=1.4cm,width=1.3cm]{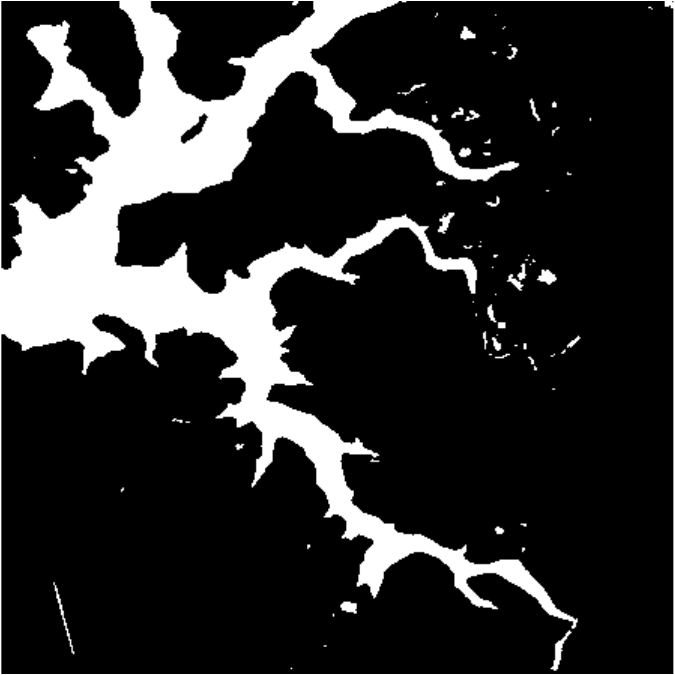}}\hspace{0.05cm}\\
		\vspace{-0.7cm}
		\centering
		\subfloat[\footnotesize ]{\includegraphics[height=1.4cm,width=1.3cm]{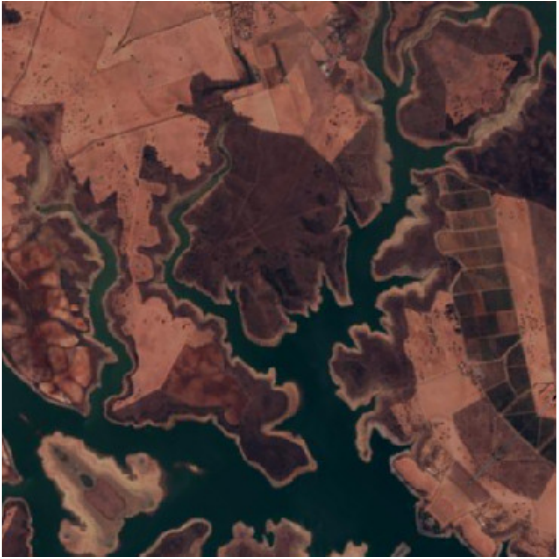}}\hspace{0.05cm}
		\subfloat[\footnotesize ]{\includegraphics[height=1.4cm,width=1.3cm]{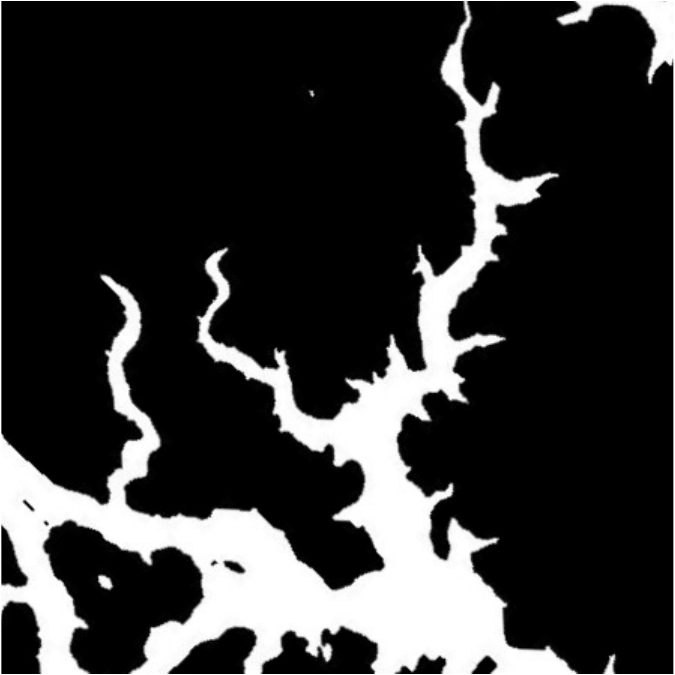}}\hspace{0.05cm}
		\subfloat[\footnotesize ]{\includegraphics[height=1.4cm,width=1.3cm]{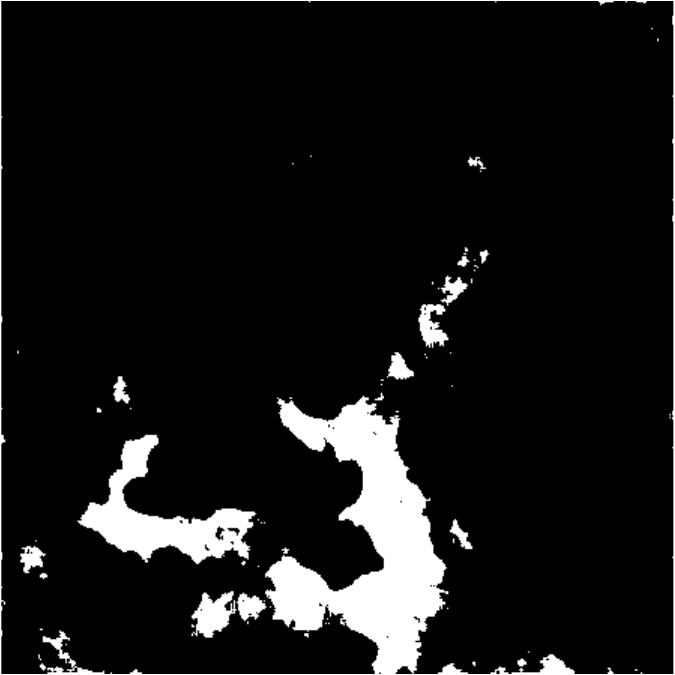}}\hspace{0.05cm}
		\subfloat[\footnotesize ]{\includegraphics[height=1.4cm,width=1.3cm]{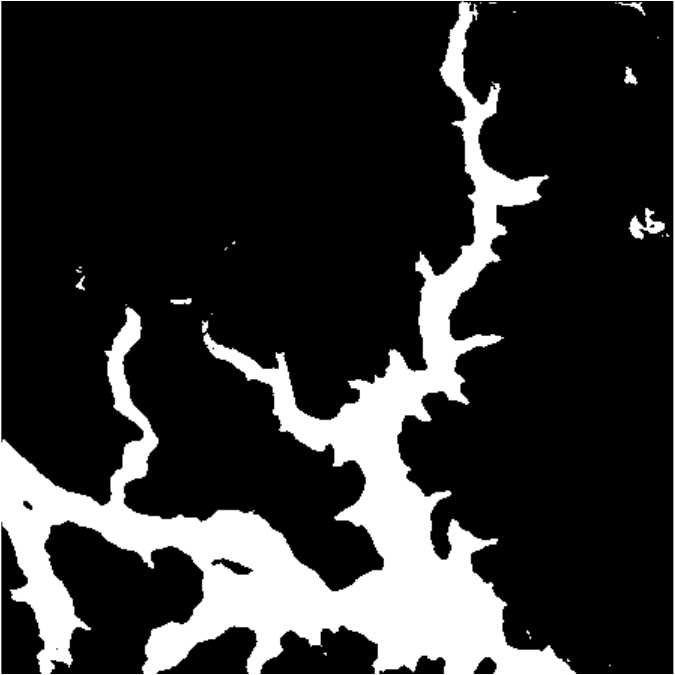}}\hspace{0.05cm}
		\subfloat[\footnotesize ]{\includegraphics[height=1.4cm,width=1.3cm]{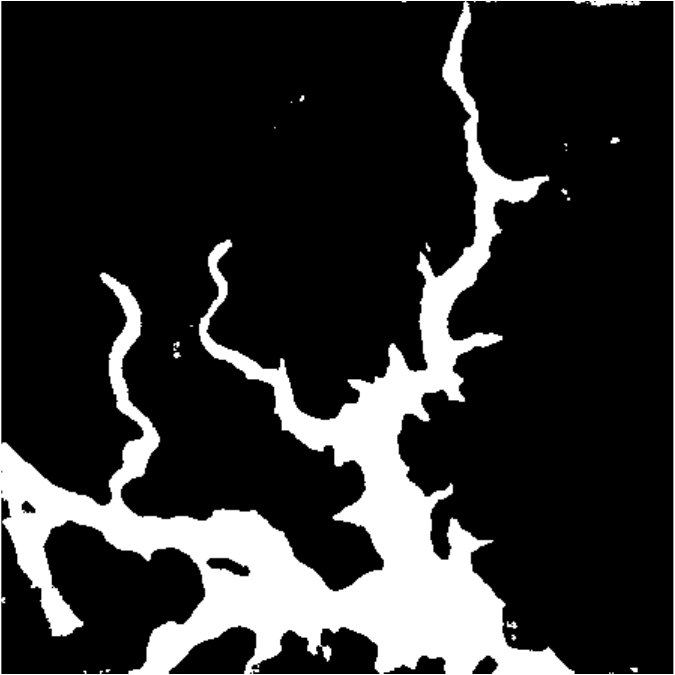}}\hspace{0.05cm}
		\subfloat[\footnotesize ]{\includegraphics[height=1.4cm,width=1.3cm]{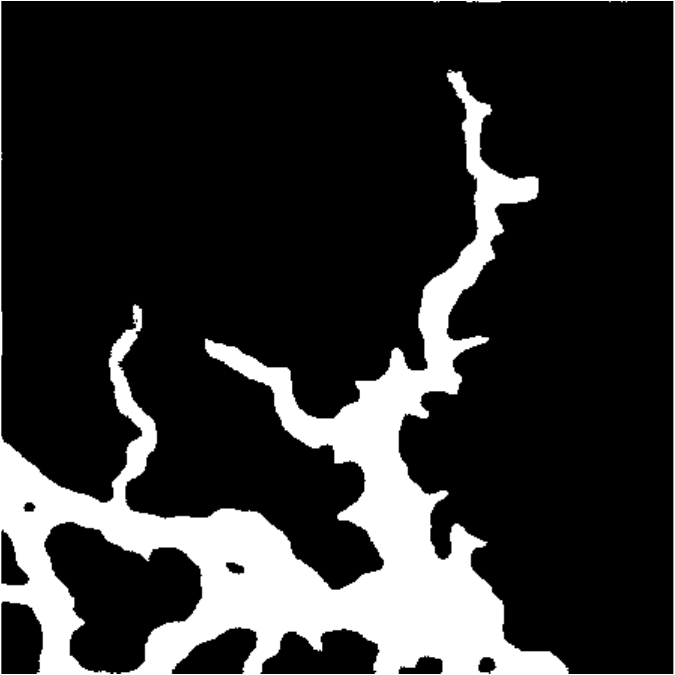}}\hspace{0.05cm}
		\subfloat[\footnotesize ]{\includegraphics[height=1.4cm,width=1.3cm]{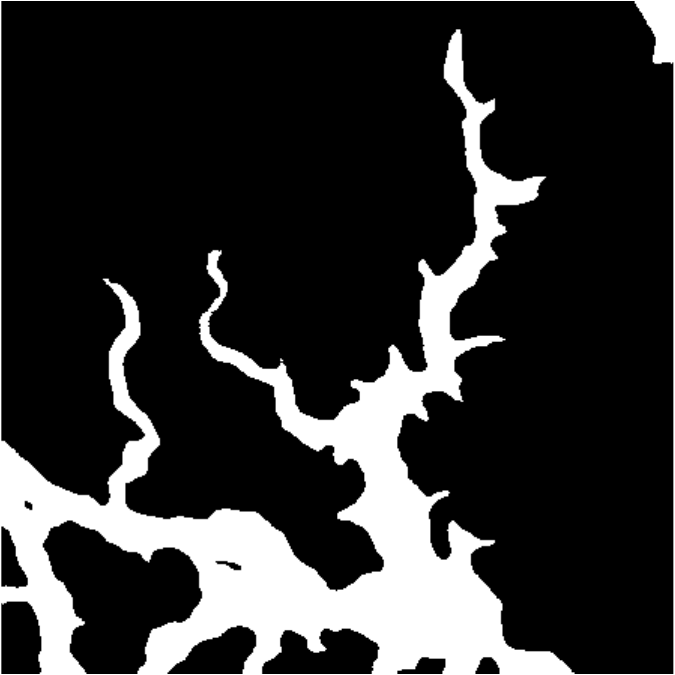}}\hspace{0.05cm}
		\subfloat[\footnotesize ]{\includegraphics[height=1.4cm,width=1.3cm]{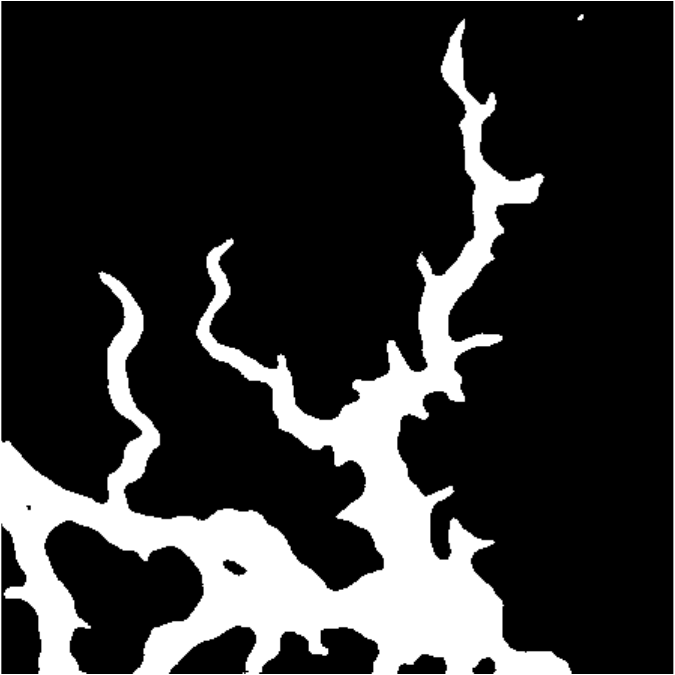}}\hspace{0.05cm}
		\subfloat[\footnotesize ]{\includegraphics[height=1.4cm,width=1.3cm]{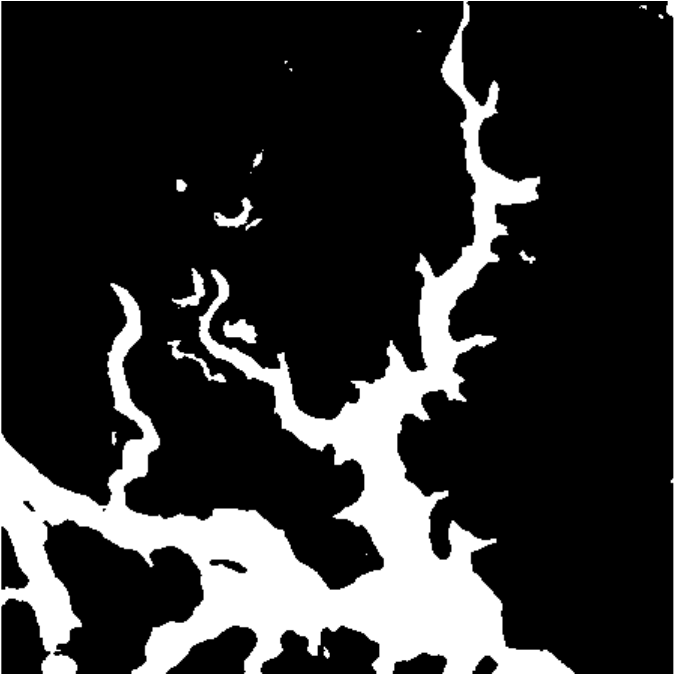}}\hspace{0.05cm}
		\subfloat[\footnotesize ]{\includegraphics[height=1.4cm,width=1.3cm]{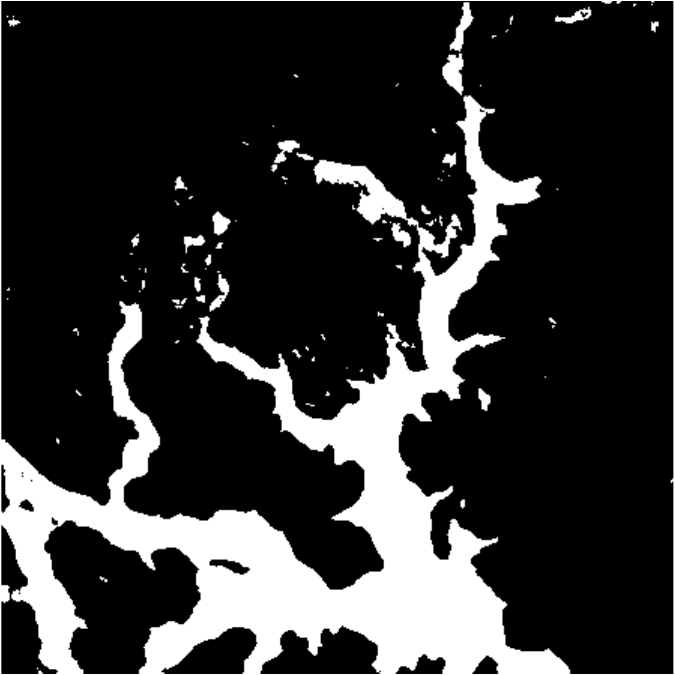}}\hspace{0.05cm}
		\subfloat[\footnotesize ]{\includegraphics[height=1.4cm,width=1.3cm]{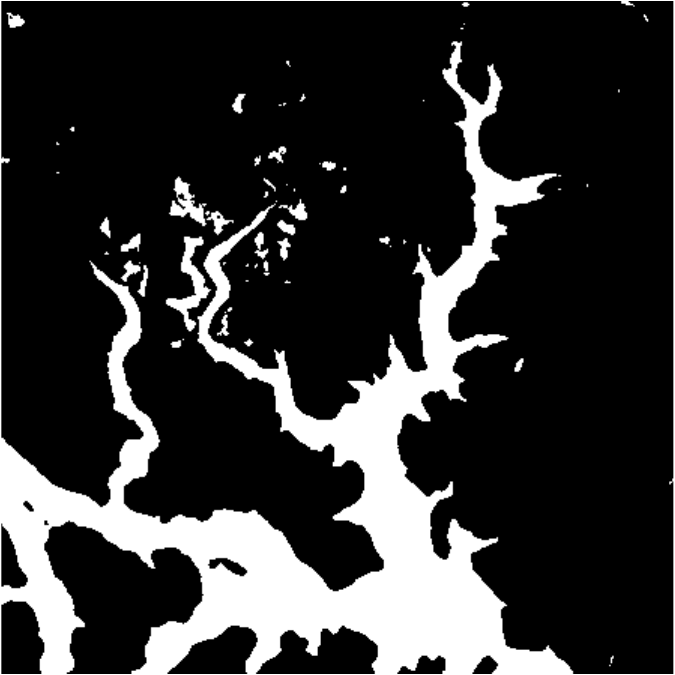}}\hspace{0.05cm}\\
		\vspace{-0.7cm}
		\centering
		\subfloat[\footnotesize ]{\includegraphics[height=1.4cm,width=1.3cm]{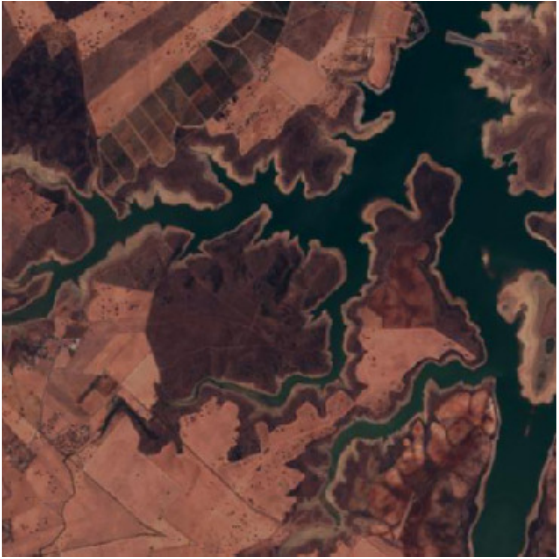}}\hspace{0.05cm}
		\subfloat[\footnotesize ]{\includegraphics[height=1.4cm,width=1.3cm]{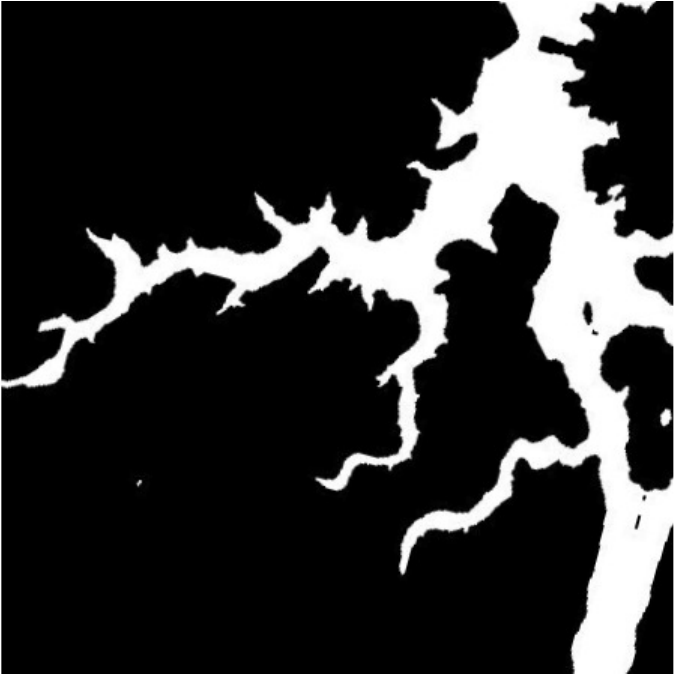}}\hspace{0.05cm}
		\subfloat[\footnotesize ]{\includegraphics[height=1.4cm,width=1.3cm]{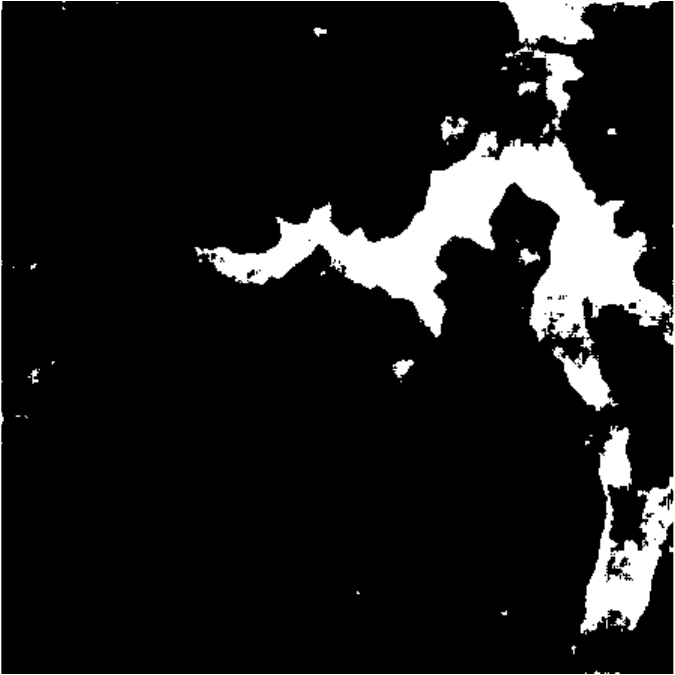}}\hspace{0.05cm}
		\subfloat[\footnotesize ]{\includegraphics[height=1.4cm,width=1.3cm]{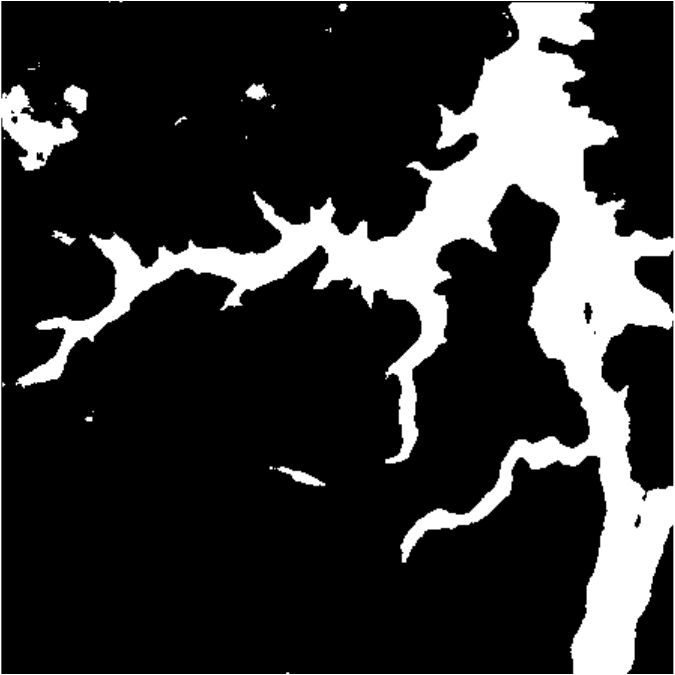}}\hspace{0.05cm}
		\subfloat[\footnotesize ]{\includegraphics[height=1.4cm,width=1.3cm]{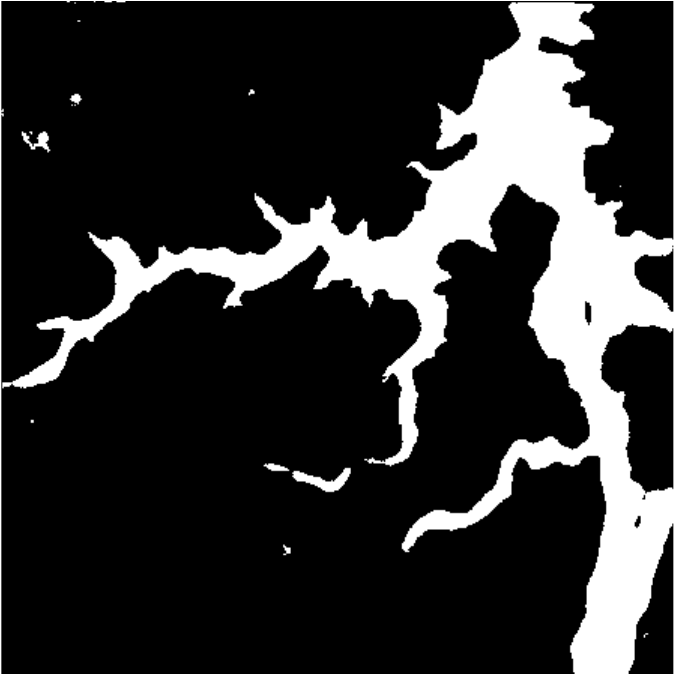}}\hspace{0.05cm}
		\subfloat[\footnotesize ]{\includegraphics[height=1.4cm,width=1.3cm]{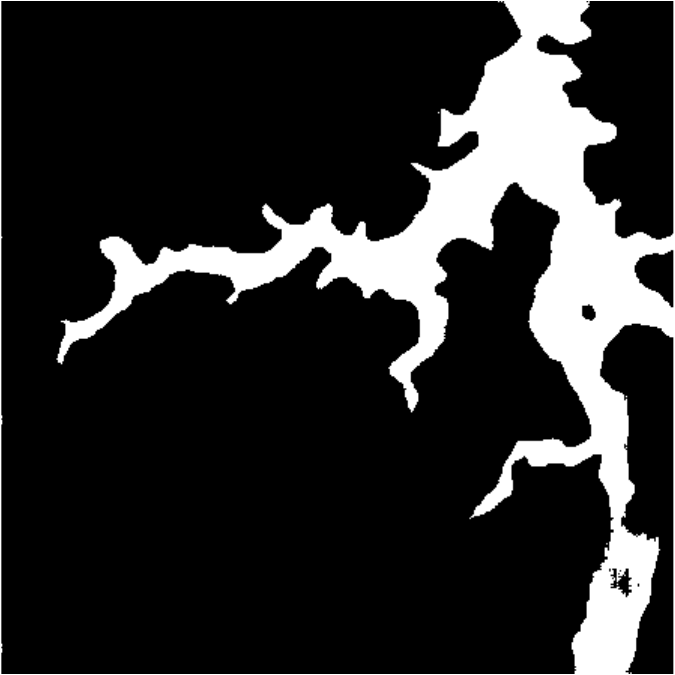}}\hspace{0.05cm}
		\subfloat[\footnotesize ]{\includegraphics[height=1.4cm,width=1.3cm]{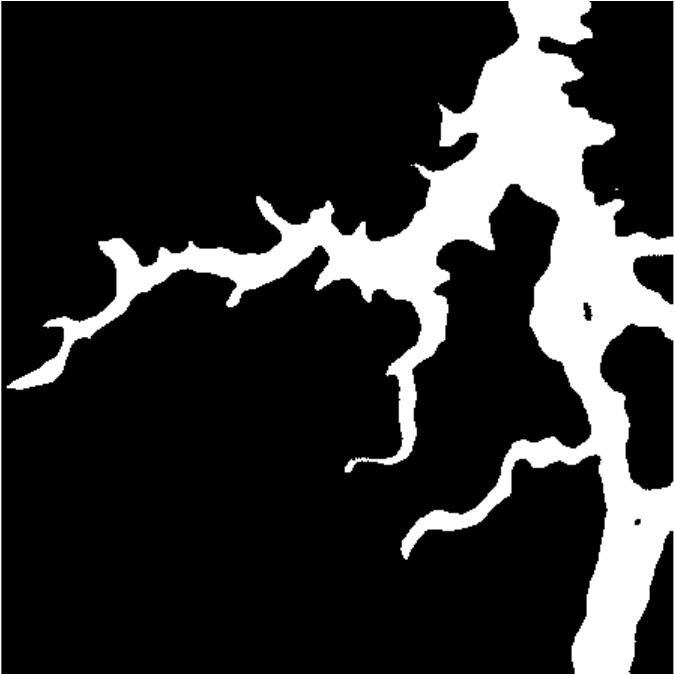}}\hspace{0.05cm}
		\subfloat[\footnotesize ]{\includegraphics[height=1.4cm,width=1.3cm]{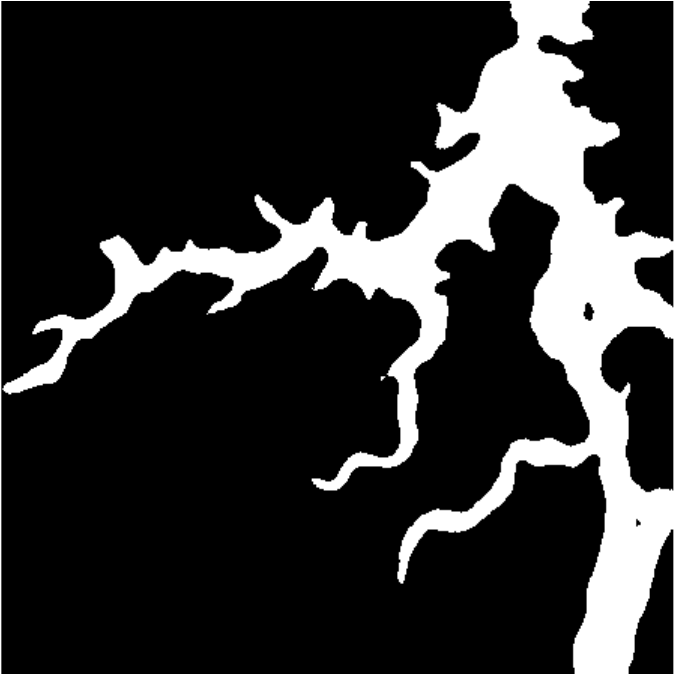}}\hspace{0.05cm}
		\subfloat[\footnotesize ]{\includegraphics[height=1.4cm,width=1.3cm]{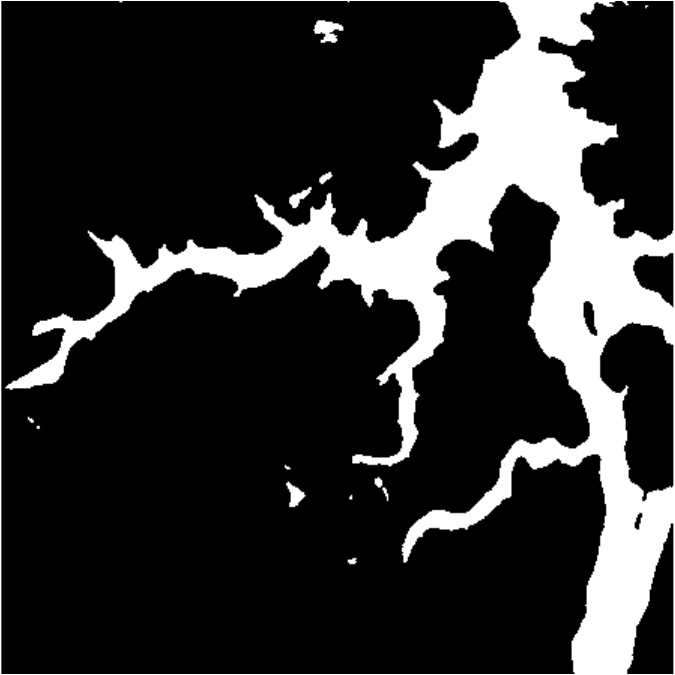}}\hspace{0.05cm}
		\subfloat[\footnotesize ]{\includegraphics[height=1.4cm,width=1.3cm]{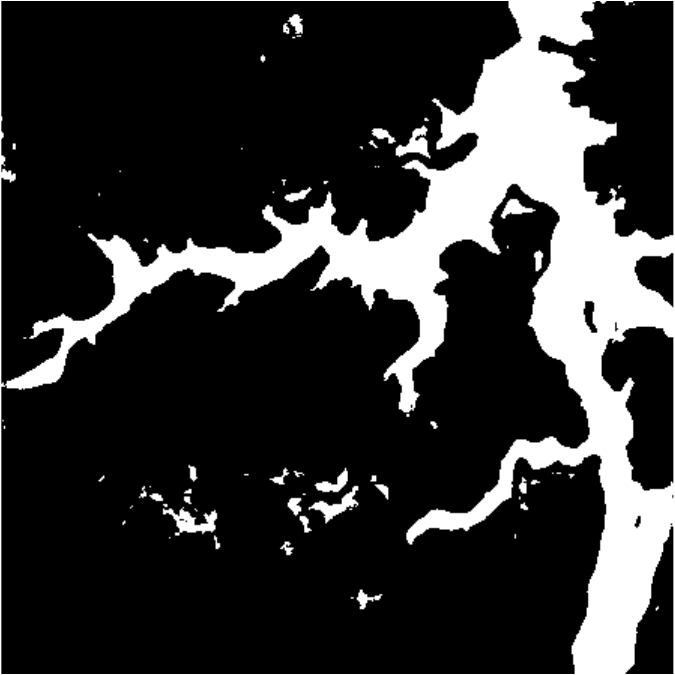}}\hspace{0.05cm}
		\subfloat[\footnotesize ]{\includegraphics[height=1.4cm,width=1.3cm]{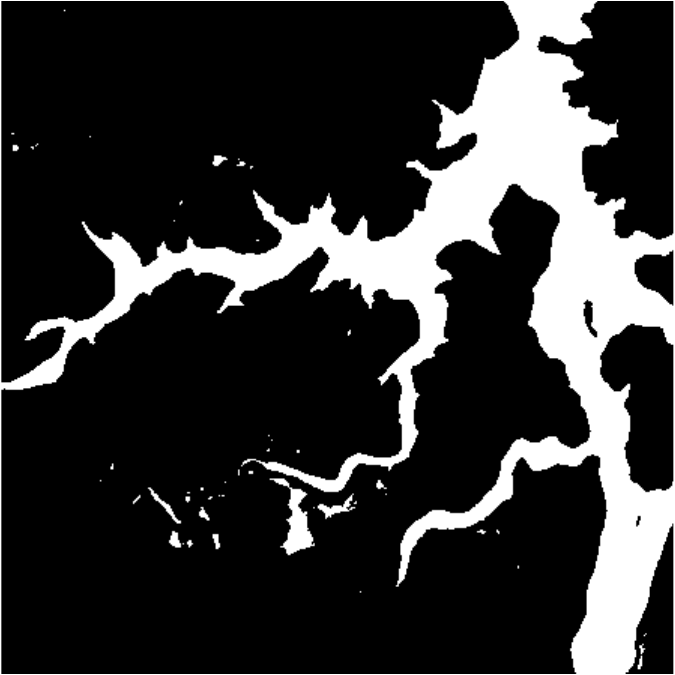}}\hspace{0.05cm}\\
		\vspace{-0.7cm}
		\centering
		\subfloat[\footnotesize (a) ]{\includegraphics[height=1.4cm,width=1.45cm]{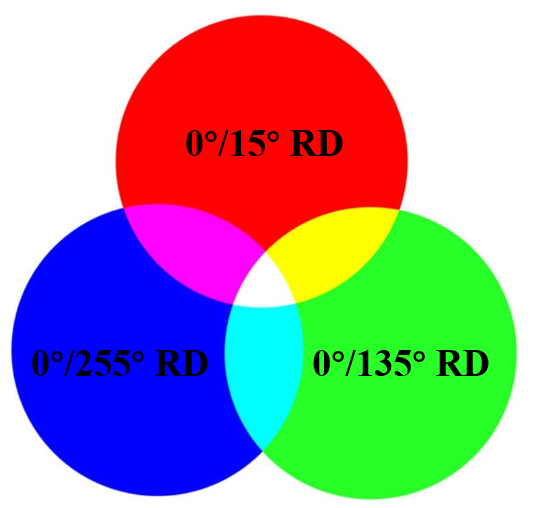}}\hspace{0.05cm}
		\subfloat[\footnotesize (b) ]{\includegraphics[height=1.4cm,width=1.3cm]{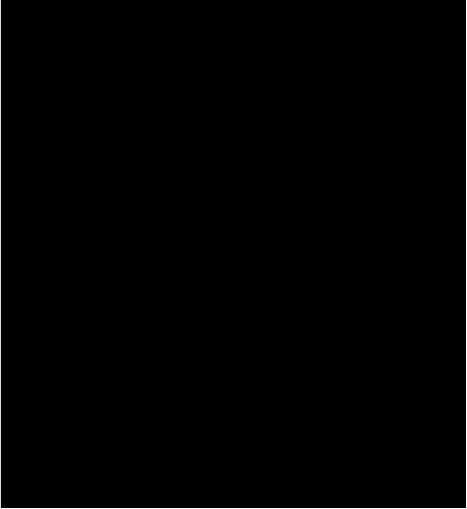}}\hspace{0.05cm}
		\subfloat[\footnotesize (c) ]{\includegraphics[height=1.4cm,width=1.3cm]{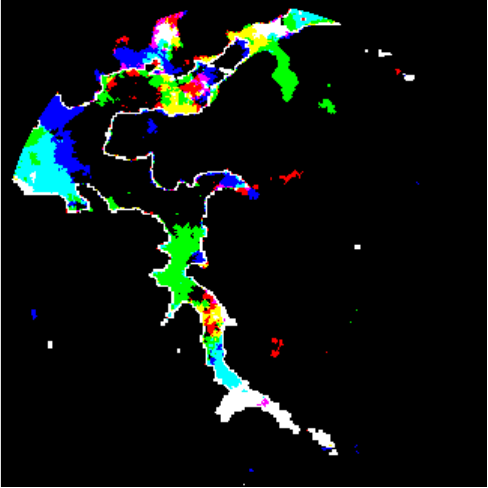}}\hspace{0.05cm}
		\subfloat[\footnotesize (d) ]{\includegraphics[height=1.4cm,width=1.3cm]{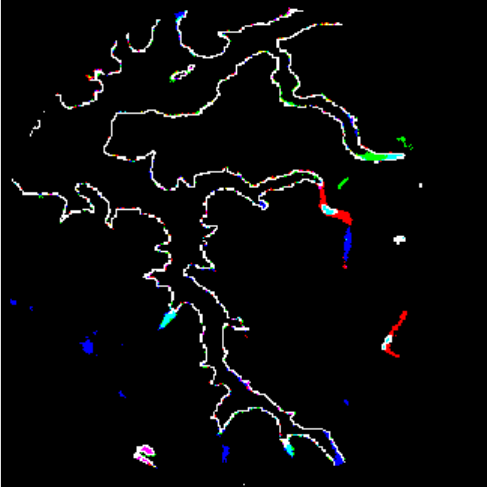}}\hspace{0.05cm}
		\subfloat[\footnotesize (e) ]{\includegraphics[height=1.4cm,width=1.3cm]{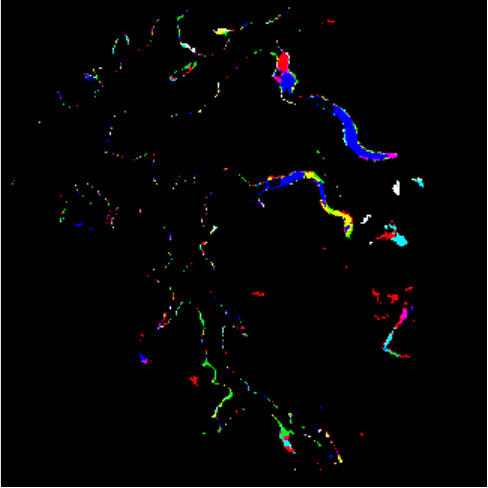}}\hspace{0.05cm}
		\subfloat[\footnotesize (f) ]{\includegraphics[height=1.4cm,width=1.3cm]{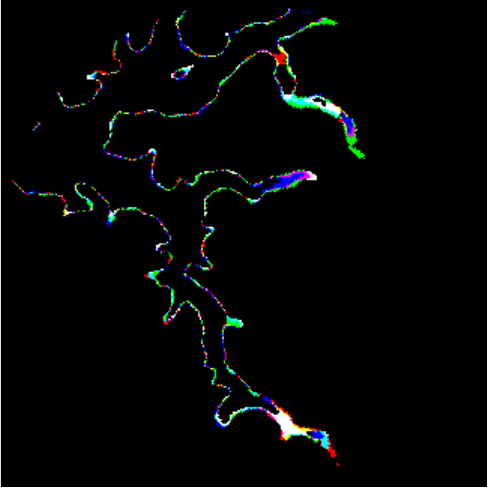}}\hspace{0.05cm}
		\subfloat[\footnotesize (g) ]{\includegraphics[height=1.4cm,width=1.3cm]{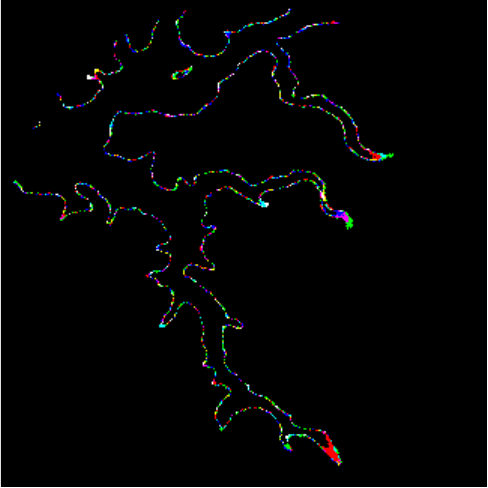}}\hspace{0.05cm}
		\subfloat[\footnotesize (h) ]{\includegraphics[height=1.4cm,width=1.3cm]{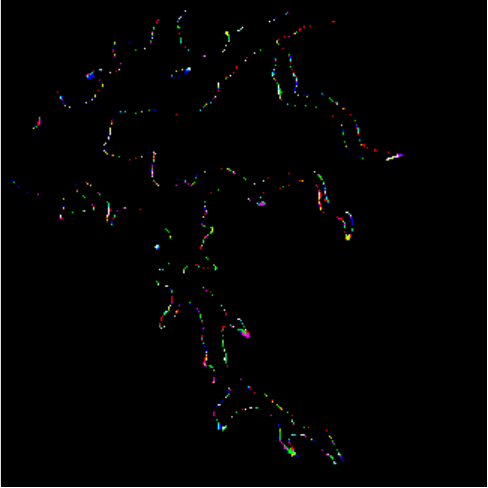}}\hspace{0.05cm}
		\subfloat[\footnotesize (i) ]{\includegraphics[height=1.4cm,width=1.3cm]{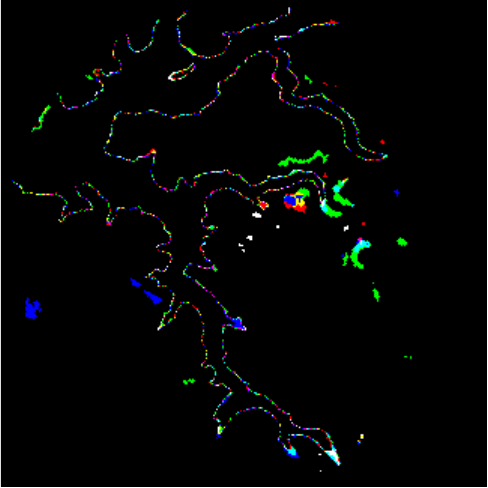}}\hspace{0.05cm}
		\subfloat[\footnotesize (j) ]{\includegraphics[height=1.4cm,width=1.3cm]{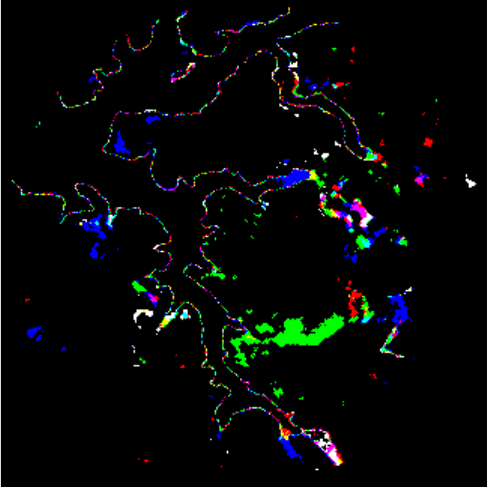}}\hspace{0.05cm}
		\subfloat[\footnotesize (k) ]{\includegraphics[height=1.4cm,width=1.3cm]{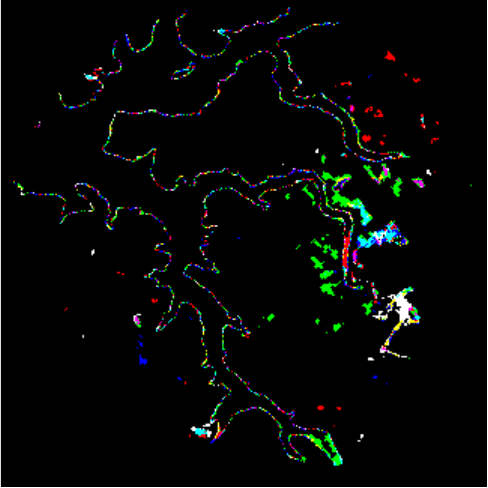}}\hspace{0.05cm}
		\label{fig_8}
		\caption{Visualization results of classical semantic segmentation networks on Satellite Images of Water Bodies.  (a) Raw optical image; (b) Ground Truth; (c) SegNet; (d) SegNet+aug; (e) SegNet(PreCM); (f) ERFNet; (g) ERFNet+aug; (h) ERFNet(PreCM); (i) RIC-CNN; (j) H-Net; (k) E2CNN. The last row are the difference maps, wherein red, green, and blue respectively denote the rotation difference between 0$^\circ$ and 15$^\circ$,  0$^\circ$ and 135$^\circ$, 0$^\circ$ and 255$^\circ$.}
	\end{figure*}
	
	\textbf{Experimental Results }
	1) \textit{Satellite Images of Water Bodies}: Table I presents the evaluation results of six networks before and after the replacement of PreCM on Satellite Images of Water Bodies dataset. During the test process, we assess the networks using 0$^\circ$, 90$^\circ$, 180$^\circ$, 270$^\circ$, and random angles. For the random angles test, we construct an orientation set spanning from 0$^\circ$ to 360$^\circ$, and randomly assign an angle from this set to each image. From Table I, four points can be easily obtained. The first one is that, the proposed PreCM takes into account different orientation information of images, so when the test images are not rotated, the networks with PreCM still have significant improvements in the evaluation metrics. The second one is that, because the PreCM-based networks have absolute rotation equivariance at 0$^\circ$/90$^\circ$/180$^\circ$/270$^\circ$, the segmentation accuracy errors caused by rotations at these special angles have been reduced to the computer floating-point precision, as evidenced by the consistent RD value of 0.00. In contrast, the orientation information of original networks can only be obtained from the dataset itself. So, without the information of different orientation angles, the IOU, MIOU, and DICE values of original networks decrease after rotating the test images. The value of RD also shows that when the images are rotated, the segmentation results are different from the original output feature maps, resulting in poor resistance to rotational interference. Thirdly, the test results at random angles demonstrate more significant performance improvements compared to the results of the test at four specific angles. Furthermore, in terms of the RD value, the network after replacement exhibits a lower RD value, indicating its superior rotational robustness. The last point is that, all the original six networks contain up-sampling and down-sampling as well as some special design of convolution, such as dilated convolution, transposed convolution, asymmetric convolution, and flattened convolution. Therefore, the successful application directly proves the universality of PreCM for arbitrarily sized images and convolution kernels. Indirectly, this also verifies the correctness of our mathematical deductions.
	
	Tables \uppercase\expandafter{\romannumeral2} and \uppercase\expandafter{\romannumeral3} present the quantitative comparison results of our method with data augmentation techniques and existing rotation equivariant networks. In Table \uppercase\expandafter{\romannumeral2}, to comprehensively evaluate the effectiveness of our method, we also conduct tests using random angles. The results indicate that although the data augmentation methods adopt four times more data for training than PreCM, the comparative results can still be obtained from PreCM. Therefore, we believe that our method can achieve comparable results with a smaller dataset, demonstrating substantial application potential. Furthermore, compared to the networks with data augmentation, our method exhibits a lower RD value, which strongly proves that our network has superior performance in resisting rotational disturbances. In Table \uppercase\expandafter{\romannumeral3}, we present detailed segmentation performance for four specific angles and random angles. The results indicate that, at four special angles, all the networks with rotation equivariance exhibit good performance, with our method standing out and demonstrating superior results. However, when facing random angle rotation, despite incorporating rotation equivariant design, the performance of all these networks decreases. Nevertheless, under such unfavorable conditions, our network still maintains the smallest accuracy loss and RD values.

	2) \textit{DRIVE}: The quantitative analysis results of this dataset have been detailed in Tables \uppercase\expandafter{\romannumeral1}, \uppercase\expandafter{\romannumeral2} and \uppercase\expandafter{\romannumeral3}. As clearly shown in Table I, even with only 20 images in the training process, the adoption of a PreCM-based replacement network significantly enhances performance. Furthermore, we have observed that when the test images are rotated, the performance of the original network undergoes a substantial decline, particularly evident in the cases of random rotation. In contrast, the segmentation network with PreCM replacement not only maintains exceptional performance stability at four specific angles, but also demonstrates far superior segmentation results and rotation robustness at random angles compared to the original network.
	
	From Table \uppercase\expandafter{\romannumeral2}, under most networks, PreCM has successfully improved segmentation performance even with a smaller number of training samples, particularly achieving an improvement of approximately 5\% in the SegNet and PSPNet. Additionally, in terms of RD values, networks based on PreCM also show a advantage compared to those based on data augmentation.
	Besides, as demonstrated in the comparison with other equivariant networks in Table \uppercase\expandafter{\romannumeral3}, our method exhibits significantly superior segmentation performance (IOU) and consistency (RD), both under random angles and the four specific angles. This fully validates the effectiveness of our method.
	
	\begin{figure*}[!ht]
		\footnotesize
		\centering
		\captionsetup[subfloat]{position=bottom,labelformat=empty}	
		\subfloat[\footnotesize ]{\includegraphics[height=1.3cm,width=1.3cm]{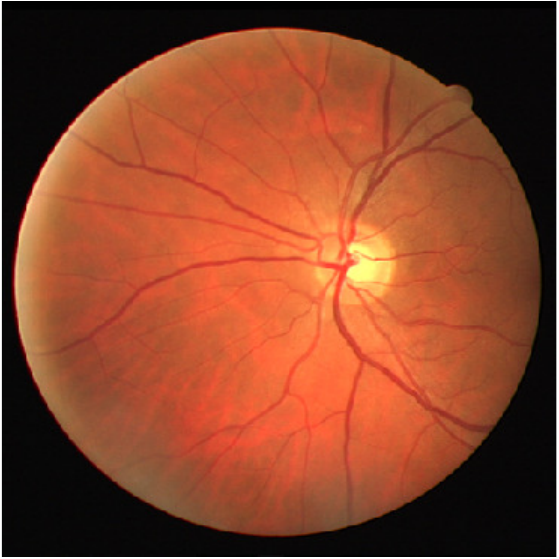}}\hspace{0.05cm}
		\subfloat[\footnotesize ]{\includegraphics[height=1.3cm,width=1.3cm]{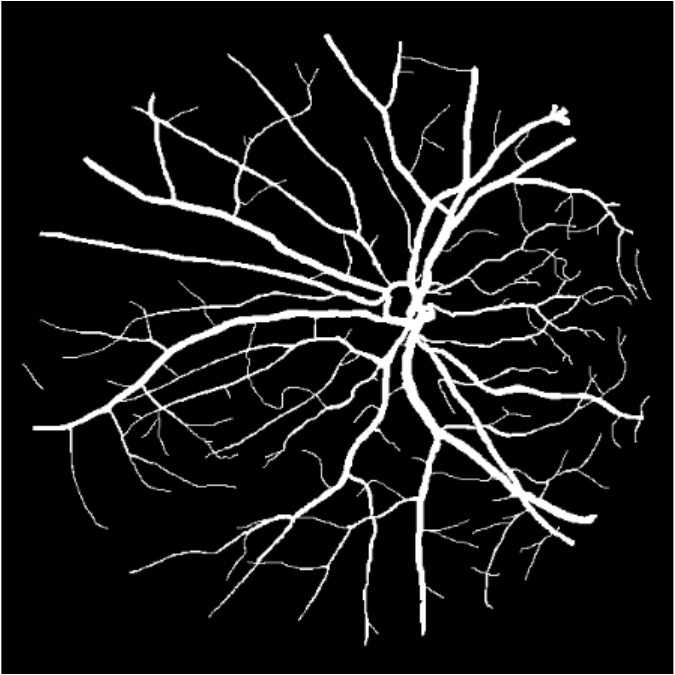}}\hspace{0.05cm}
		\subfloat[\footnotesize ]{\includegraphics[height=1.3cm,width=1.3cm]{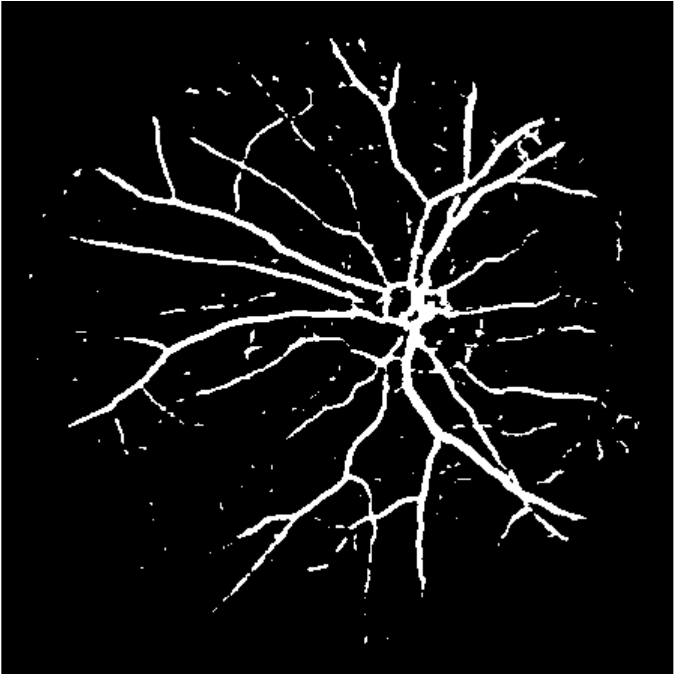}}\hspace{0.05cm}
		\subfloat[\footnotesize ]{\includegraphics[height=1.3cm,width=1.3cm]{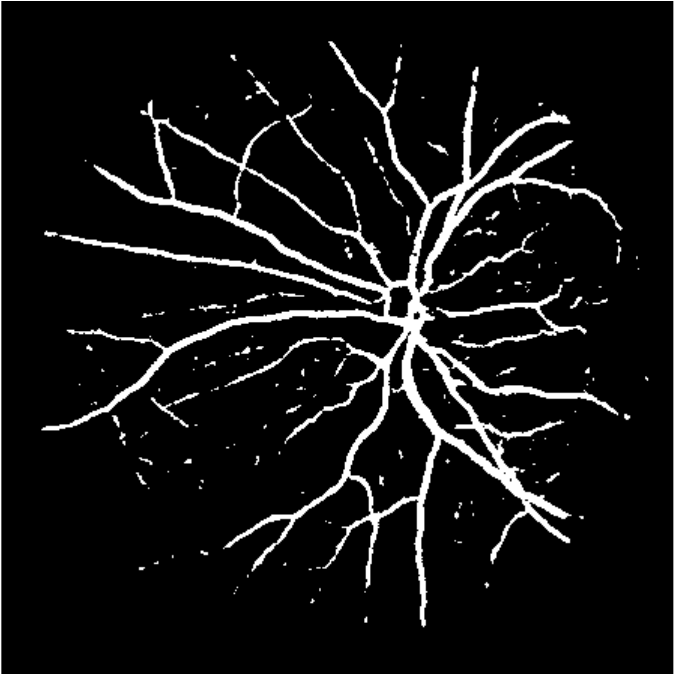}}\hspace{0.05cm}
		\subfloat[\footnotesize ]{\includegraphics[height=1.3cm,width=1.3cm]{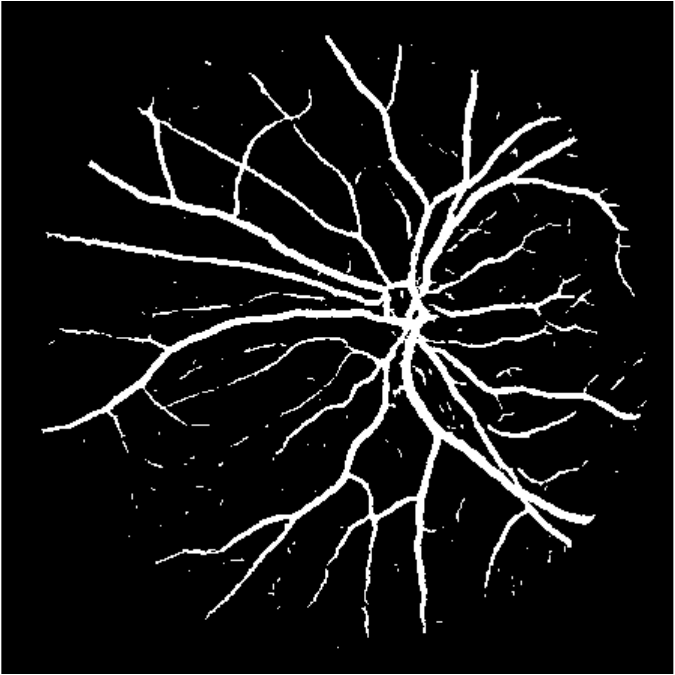}}\hspace{0.05cm}
		\subfloat[\footnotesize ]{\includegraphics[height=1.3cm,width=1.3cm]{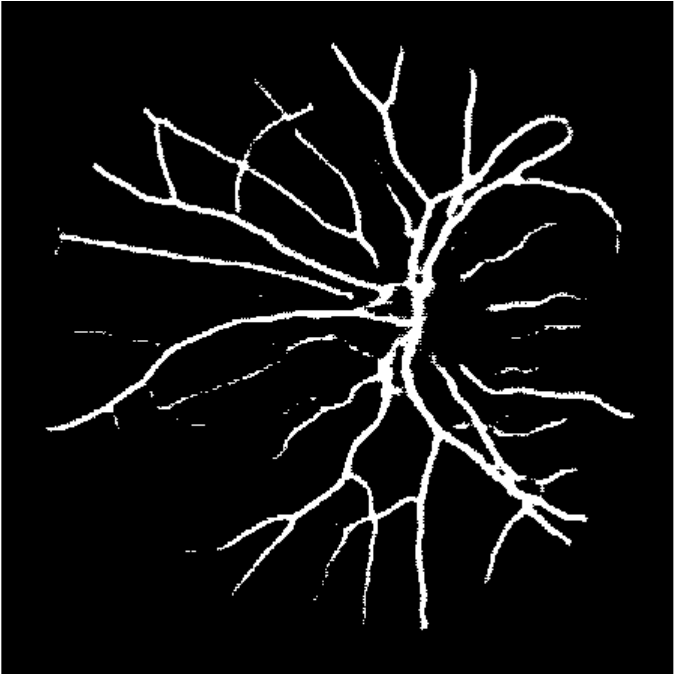}}\hspace{0.05cm}
		\subfloat[\footnotesize ]{\includegraphics[height=1.3cm,width=1.3cm]{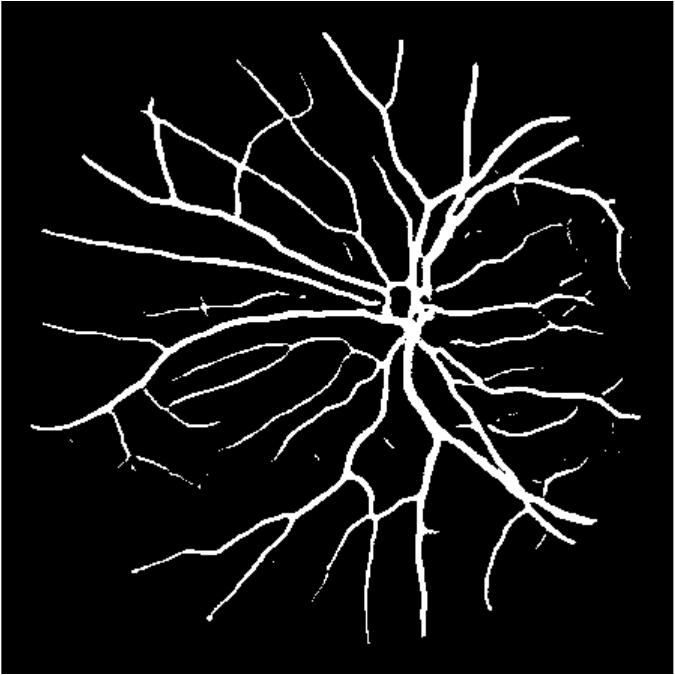}}\hspace{0.05cm}
		\subfloat[\footnotesize ]{\includegraphics[height=1.3cm,width=1.3cm]{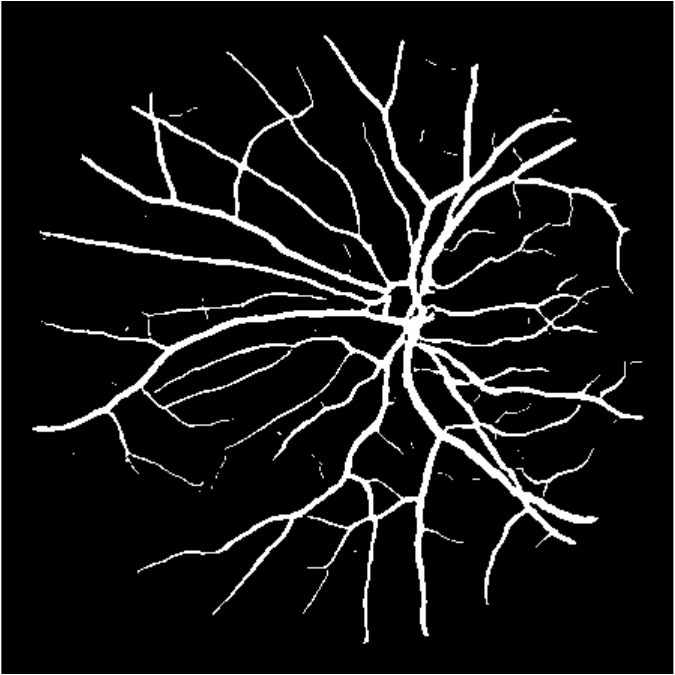}}\hspace{0.05cm}
		\subfloat[\footnotesize ]{\includegraphics[height=1.3cm,width=1.3cm]{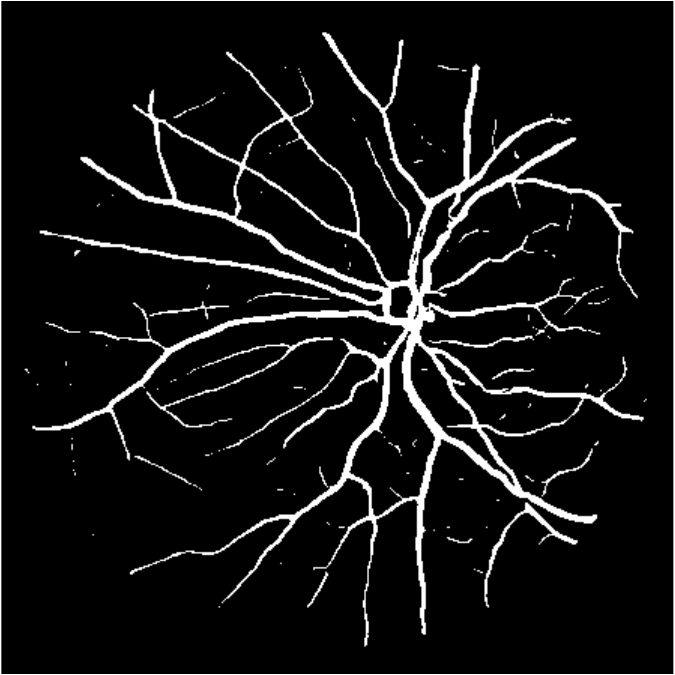}}\hspace{0.05cm}
		\subfloat[\footnotesize ]{\includegraphics[height=1.3cm,width=1.3cm]{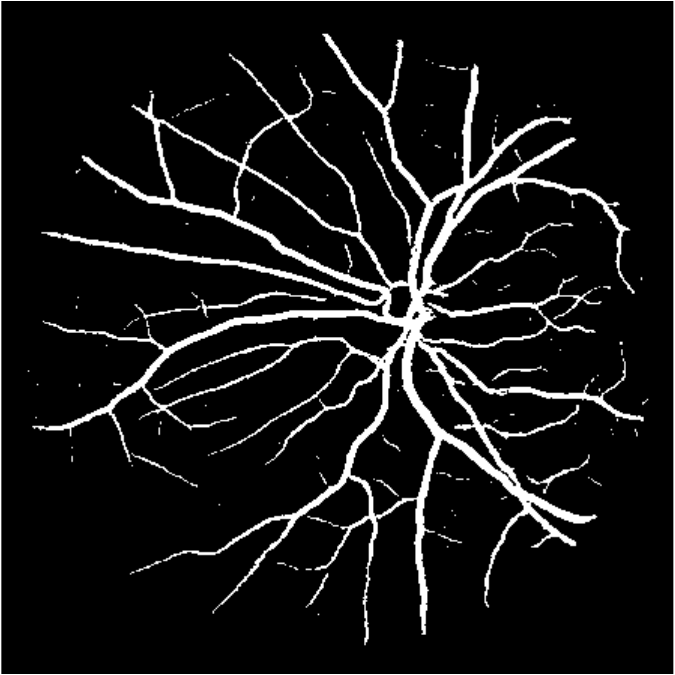}}\hspace{0.05cm}
		\subfloat[\footnotesize ]{\includegraphics[height=1.3cm,width=1.3cm]{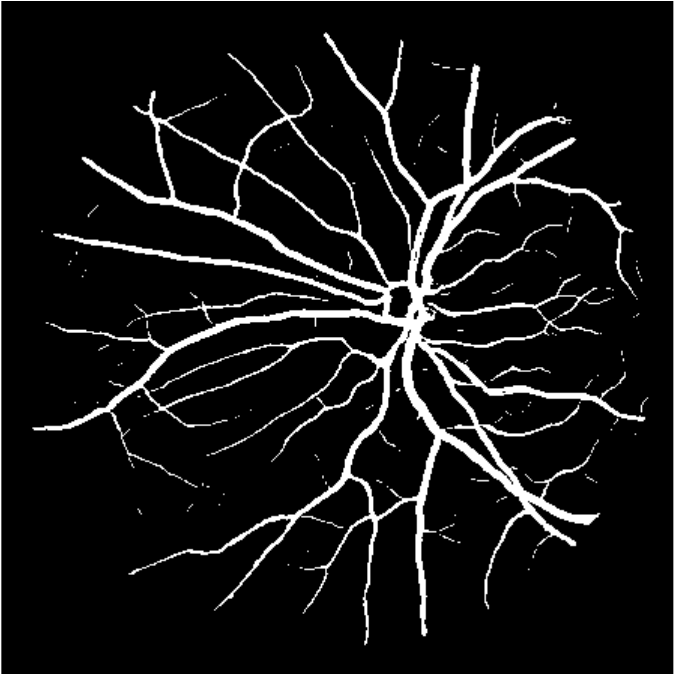}}\hspace{0.05cm}\\
		\vspace{-0.7cm}
		\centering
		\subfloat[\footnotesize ]{\includegraphics[height=1.3cm,width=1.3cm]{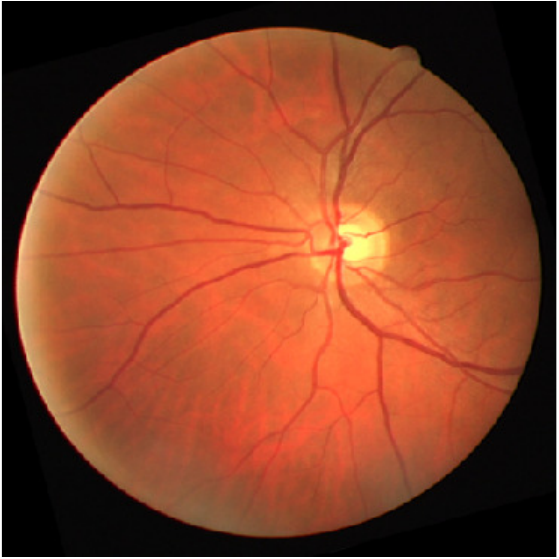}}\hspace{0.05cm}
		\subfloat[\footnotesize ]{\includegraphics[height=1.3cm,width=1.3cm]{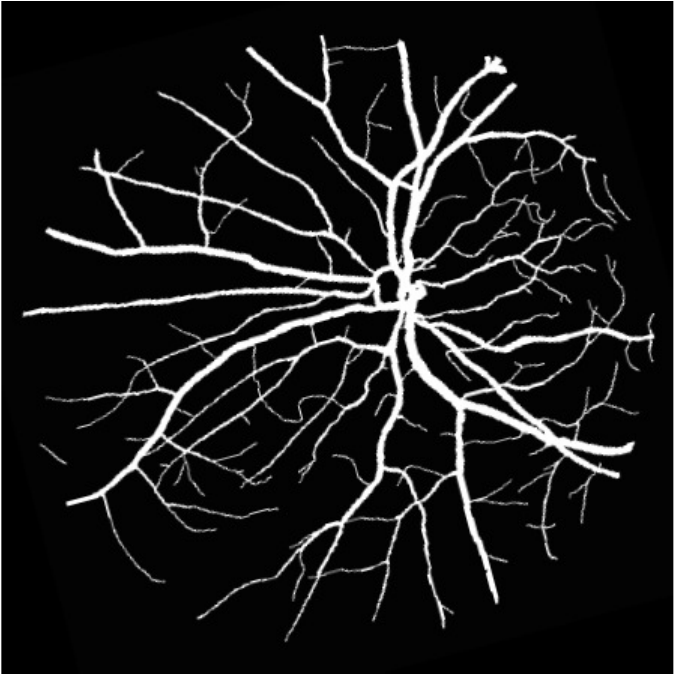}}\hspace{0.05cm}
		\subfloat[\footnotesize ]{\includegraphics[height=1.3cm,width=1.3cm]{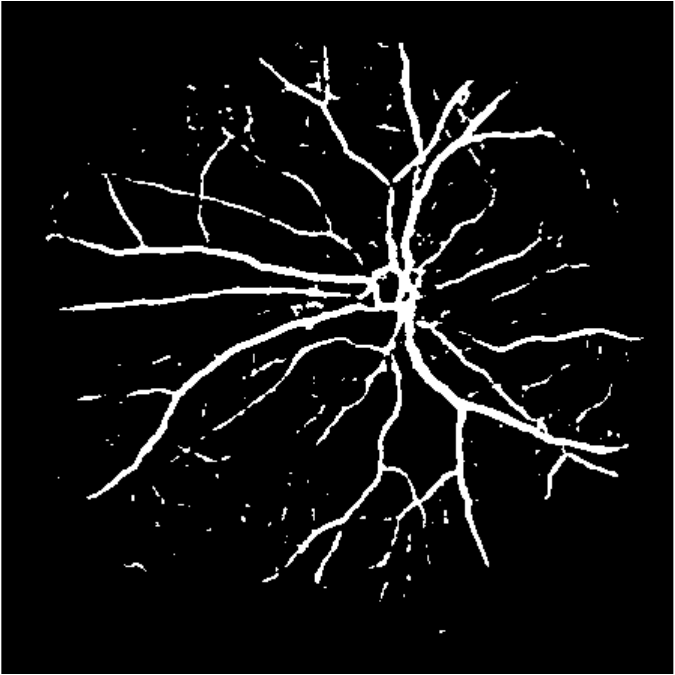}}\hspace{0.05cm}
		\subfloat[\footnotesize ]{\includegraphics[height=1.3cm,width=1.3cm]{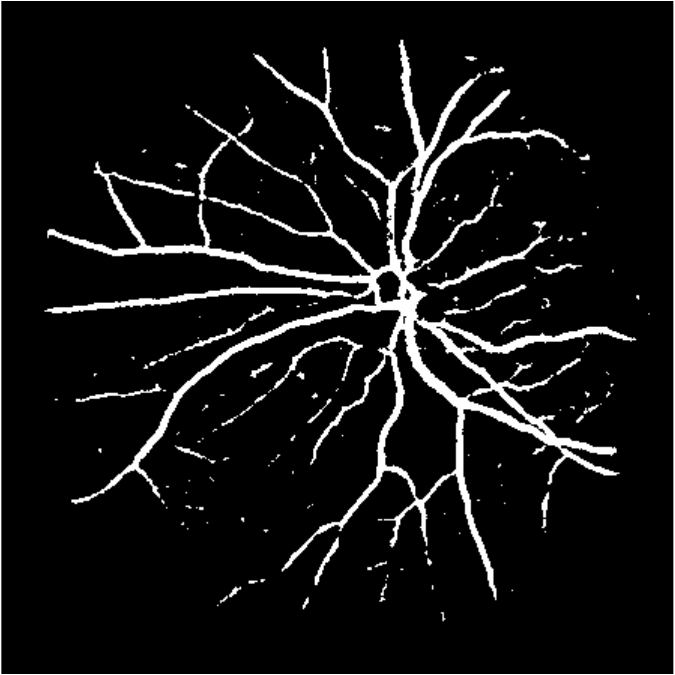}}\hspace{0.05cm}
		\subfloat[\footnotesize ]{\includegraphics[height=1.3cm,width=1.3cm]{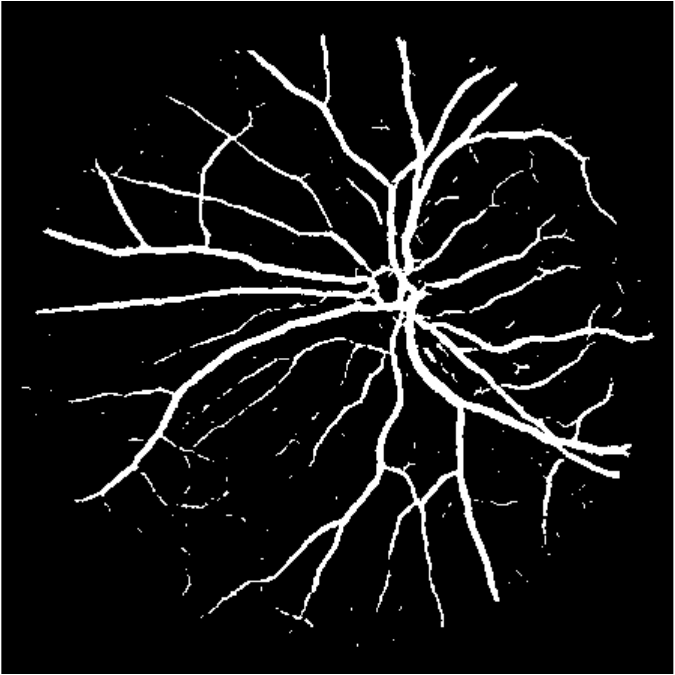}}\hspace{0.05cm}
		\subfloat[\footnotesize ]{\includegraphics[height=1.3cm,width=1.3cm]{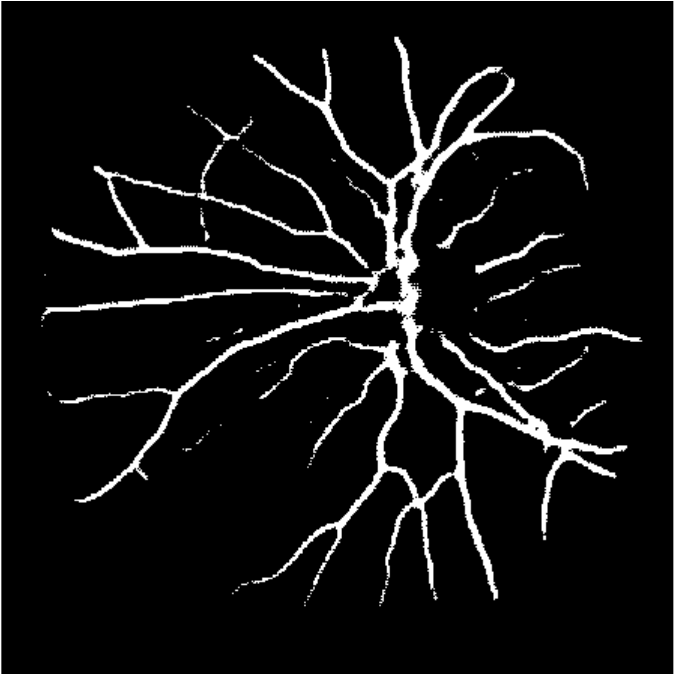}}\hspace{0.05cm}
		\subfloat[\footnotesize ]{\includegraphics[height=1.3cm,width=1.3cm]{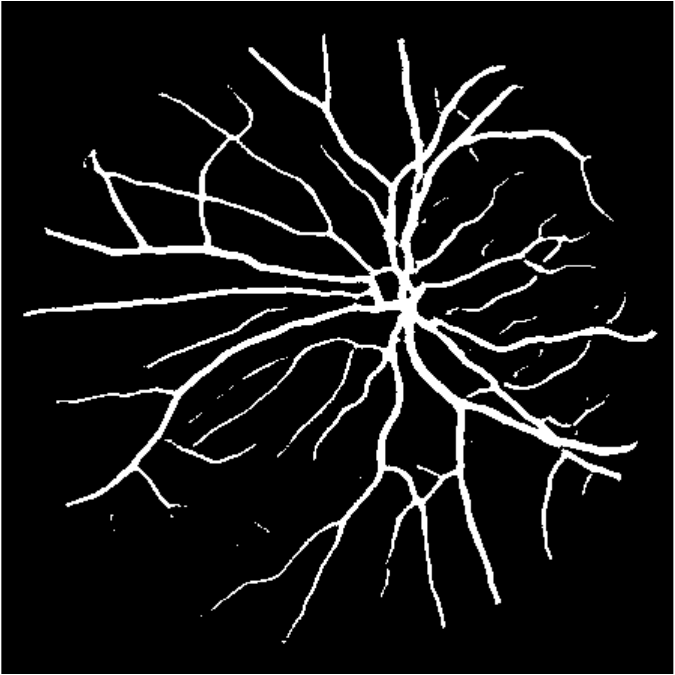}}\hspace{0.05cm}
		\subfloat[\footnotesize ]{\includegraphics[height=1.3cm,width=1.3cm]{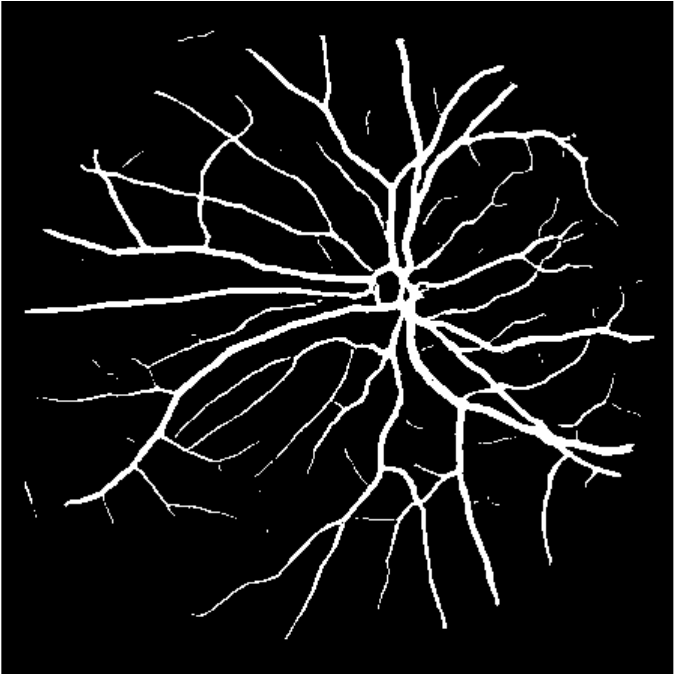}}\hspace{0.05cm}
		\subfloat[\footnotesize ]{\includegraphics[height=1.3cm,width=1.3cm]{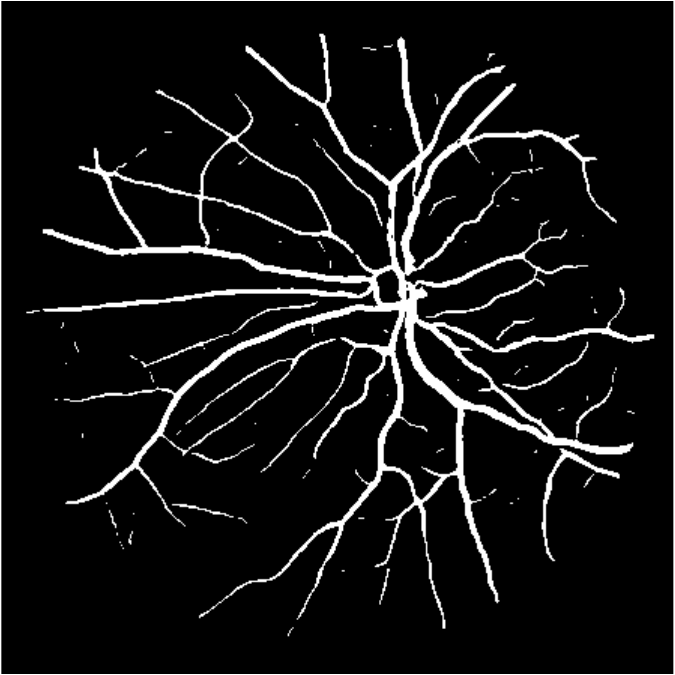}}\hspace{0.05cm}
		\subfloat[\footnotesize ]{\includegraphics[height=1.3cm,width=1.3cm]{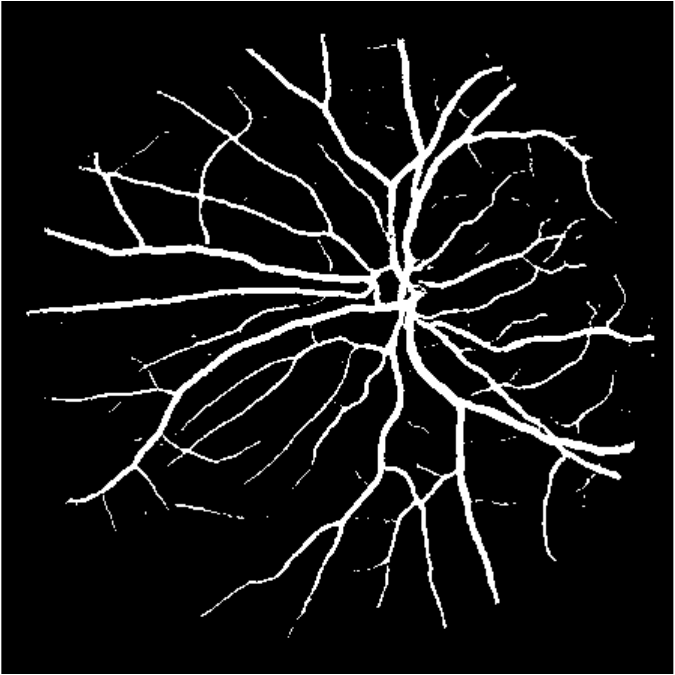}}\hspace{0.05cm}
		\subfloat[\footnotesize ]{\includegraphics[height=1.3cm,width=1.3cm]{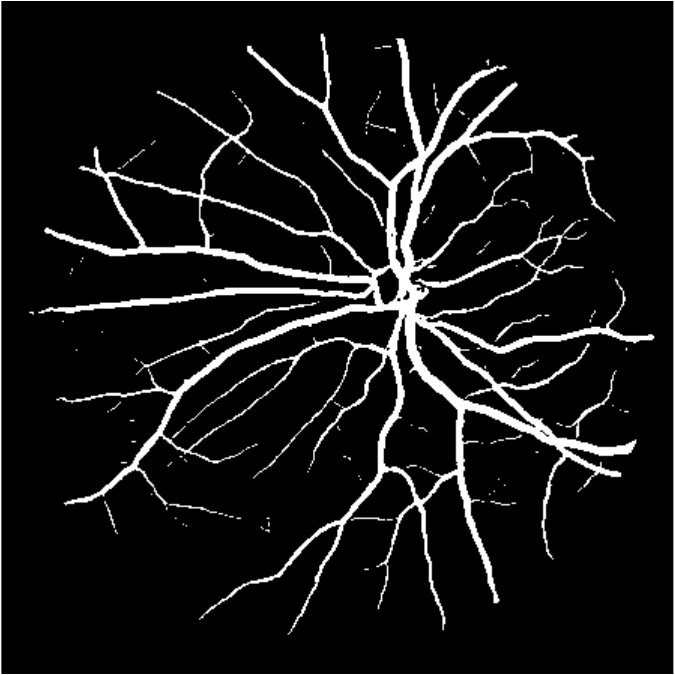}}\hspace{0.05cm}\\
		\vspace{-0.7cm}
		\centering
		\subfloat[\footnotesize ]{\includegraphics[height=1.3cm,width=1.3cm]{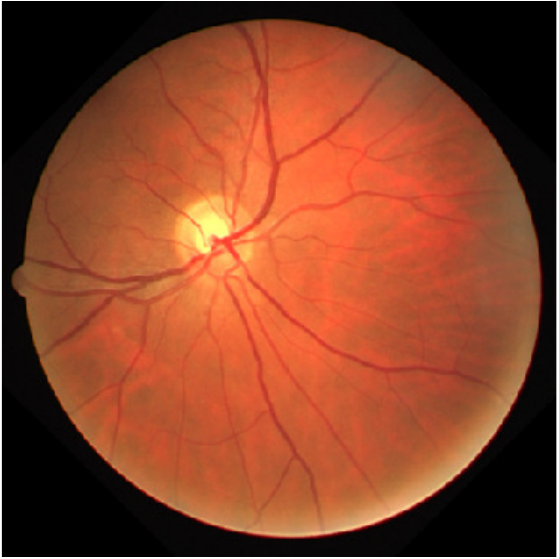}}\hspace{0.05cm}
		\subfloat[\footnotesize ]{\includegraphics[height=1.3cm,width=1.3cm]{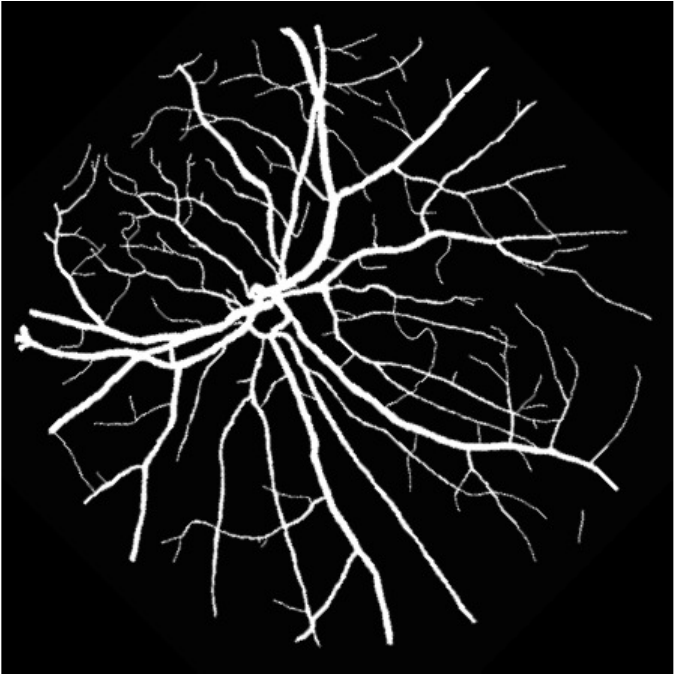}}\hspace{0.05cm}
		\subfloat[\footnotesize ]{\includegraphics[height=1.3cm,width=1.3cm]{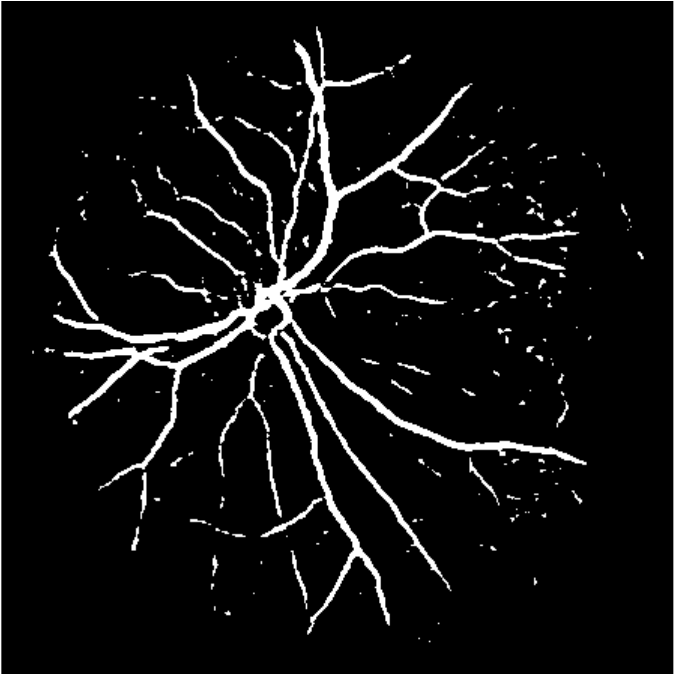}}\hspace{0.05cm}
		\subfloat[\footnotesize ]{\includegraphics[height=1.3cm,width=1.3cm]{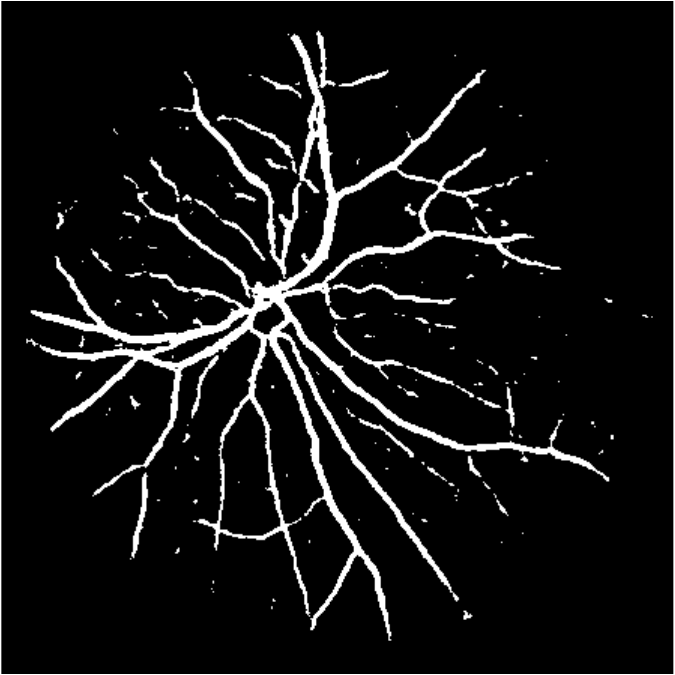}}\hspace{0.05cm}
		\subfloat[\footnotesize ]{\includegraphics[height=1.3cm,width=1.3cm]{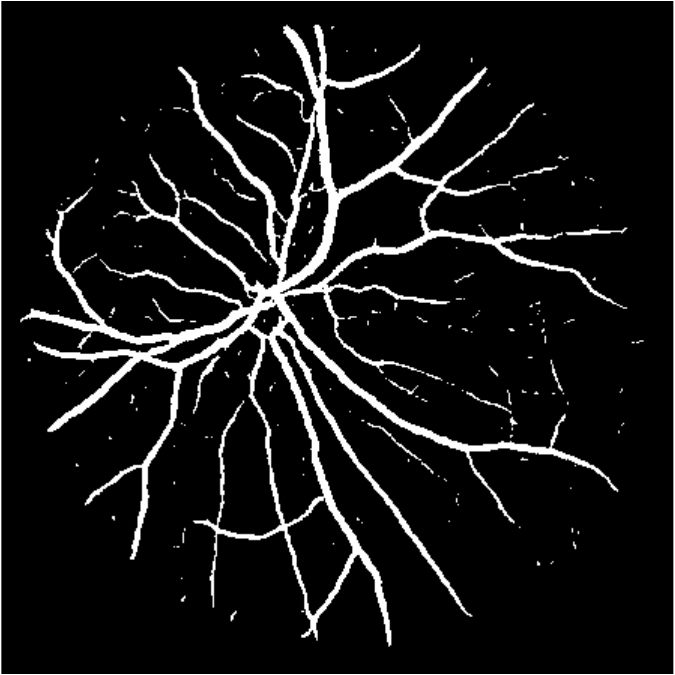}}\hspace{0.05cm}
		\subfloat[\footnotesize ]{\includegraphics[height=1.3cm,width=1.3cm]{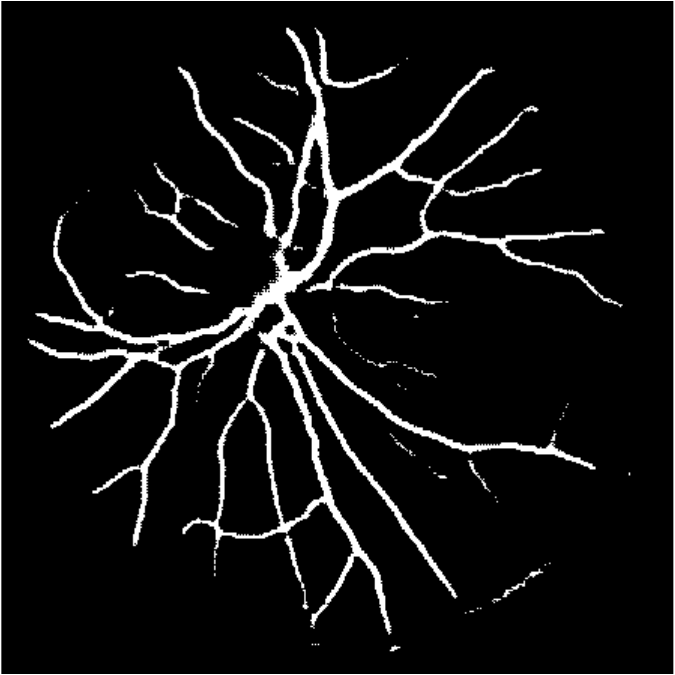}}\hspace{0.05cm}
		\subfloat[\footnotesize ]{\includegraphics[height=1.3cm,width=1.3cm]{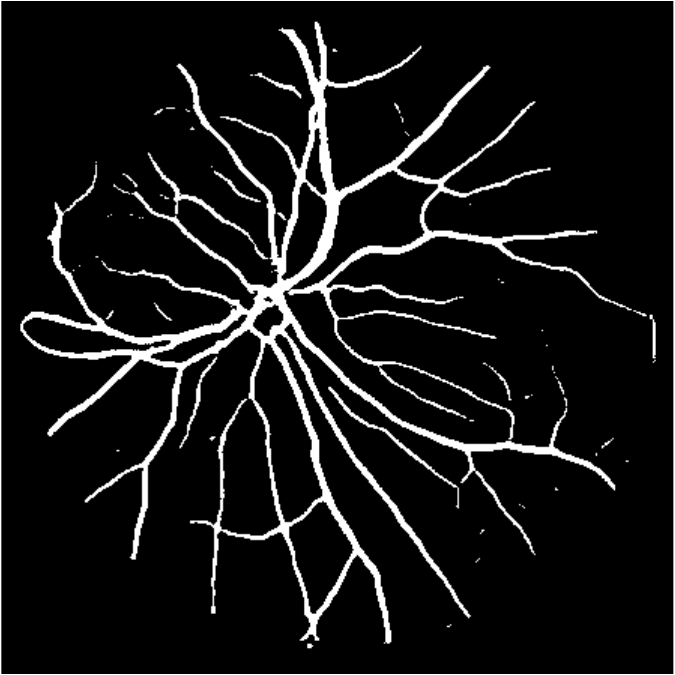}}\hspace{0.05cm}
		\subfloat[\footnotesize ]{\includegraphics[height=1.3cm,width=1.3cm]{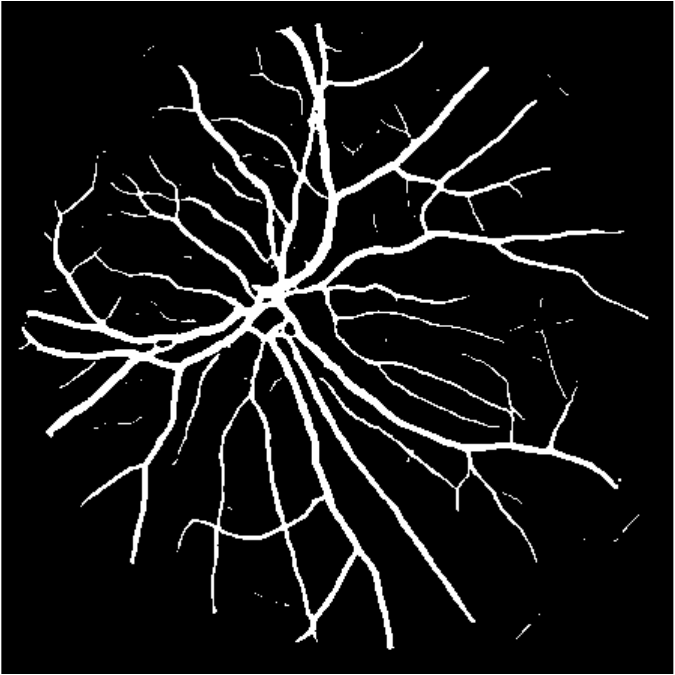}}\hspace{0.05cm}
		\subfloat[\footnotesize ]{\includegraphics[height=1.3cm,width=1.3cm]{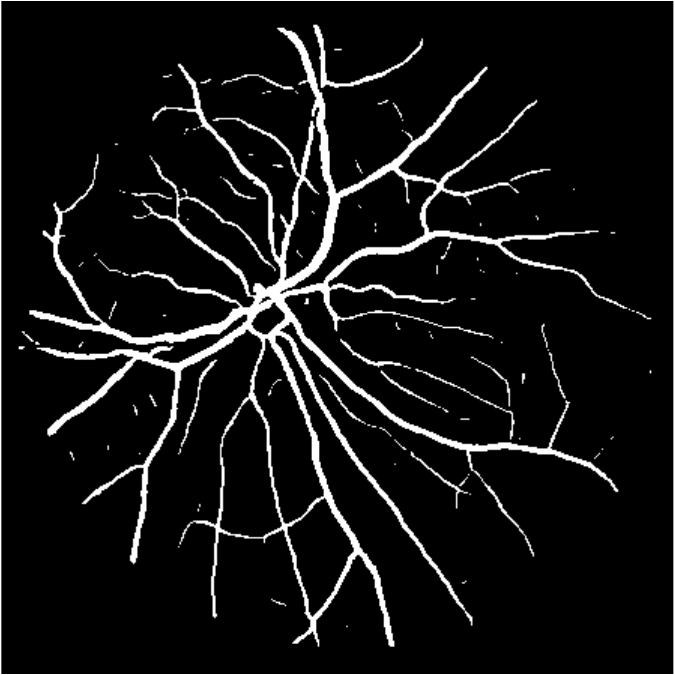}}\hspace{0.05cm}
		\subfloat[\footnotesize ]{\includegraphics[height=1.3cm,width=1.3cm]{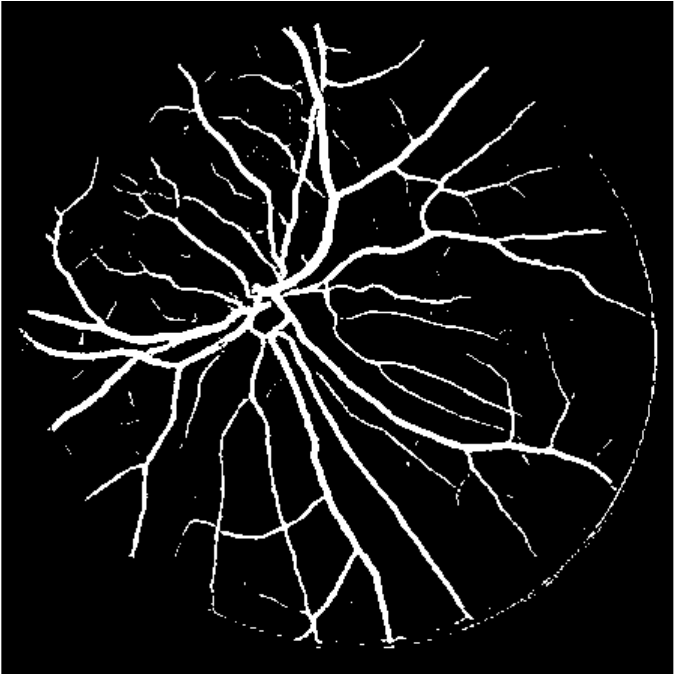}}\hspace{0.05cm}
		\subfloat[\footnotesize ]{\includegraphics[height=1.3cm,width=1.3cm]{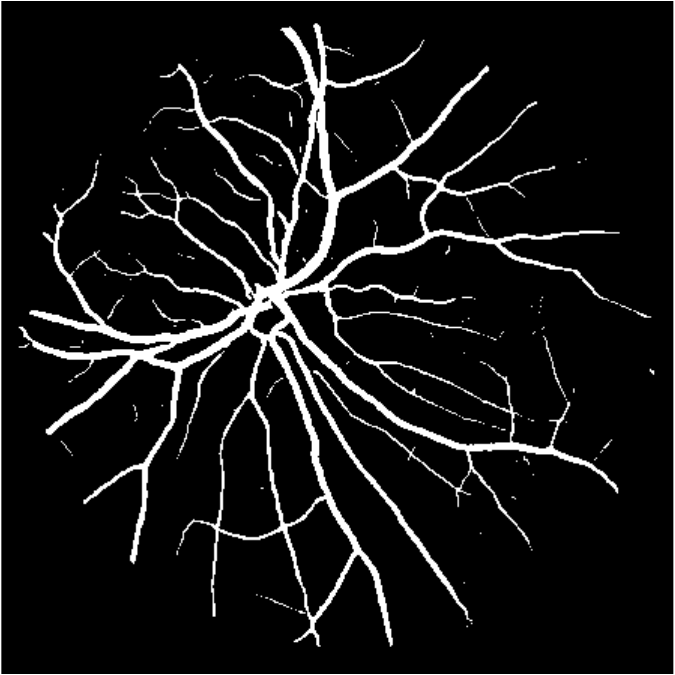}}\hspace{0.05cm}\\
		\vspace{-0.7cm}
		\centering
		\subfloat[\footnotesize ]{\includegraphics[height=1.3cm,width=1.3cm]{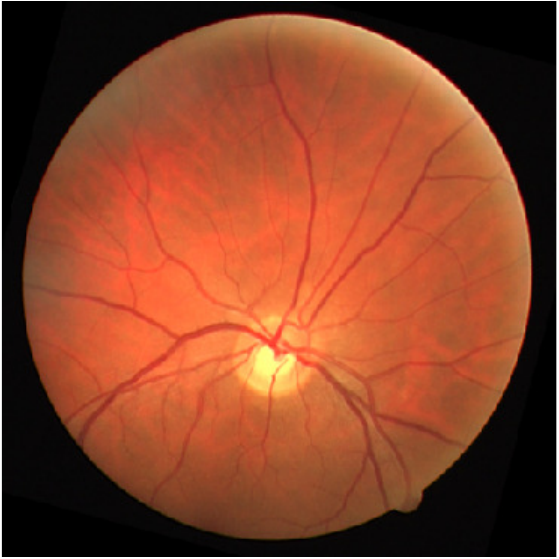}}\hspace{0.05cm}
		\subfloat[\footnotesize ]{\includegraphics[height=1.3cm,width=1.3cm]{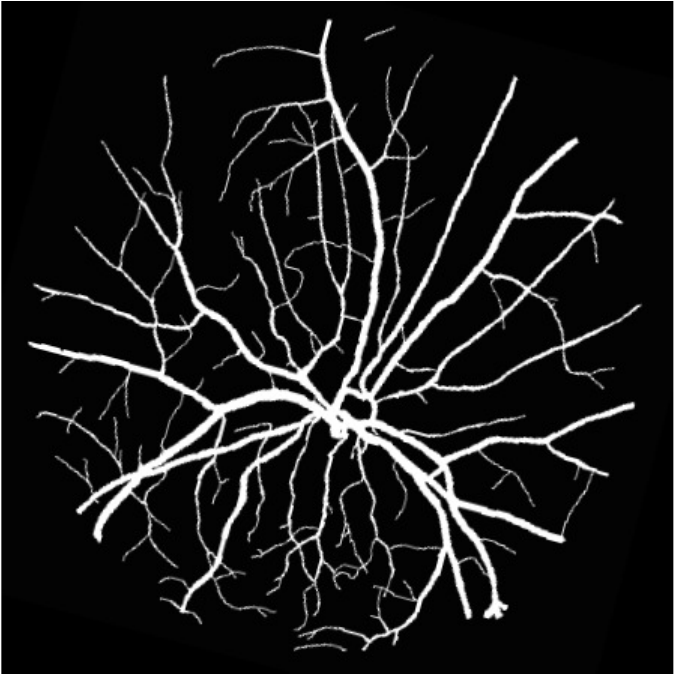}}\hspace{0.05cm}
		\subfloat[\footnotesize ]{\includegraphics[height=1.3cm,width=1.3cm]{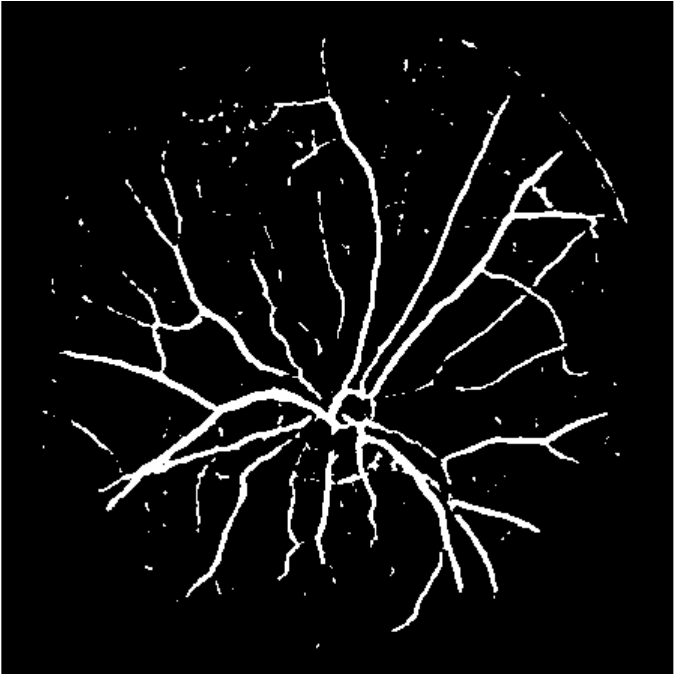}}\hspace{0.05cm}
		\subfloat[\footnotesize ]{\includegraphics[height=1.3cm,width=1.3cm]{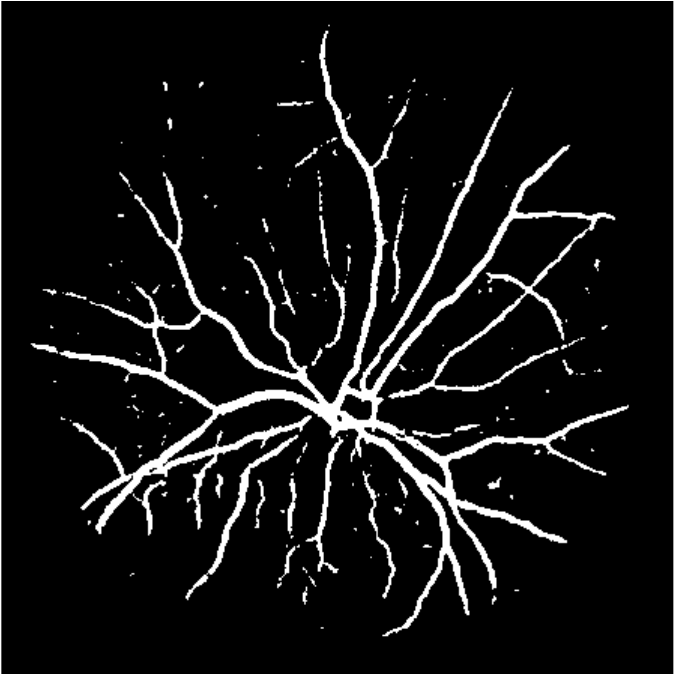}}\hspace{0.05cm}
		\subfloat[\footnotesize ]{\includegraphics[height=1.3cm,width=1.3cm]{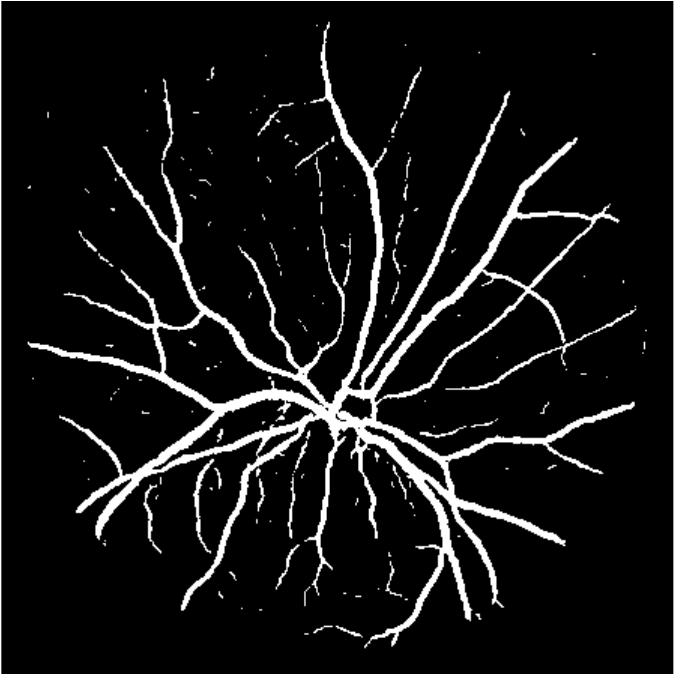}}\hspace{0.05cm}
		\subfloat[\footnotesize ]{\includegraphics[height=1.3cm,width=1.3cm]{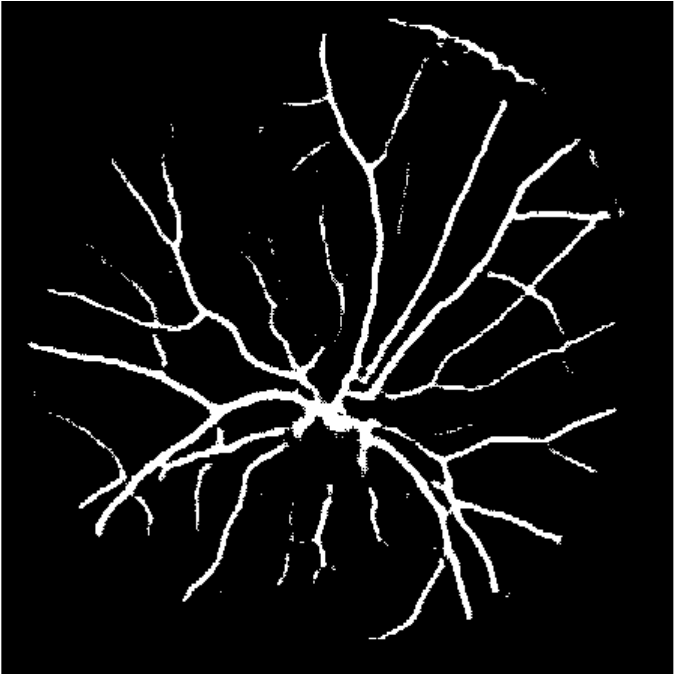}}\hspace{0.05cm}
		\subfloat[\footnotesize ]{\includegraphics[height=1.3cm,width=1.3cm]{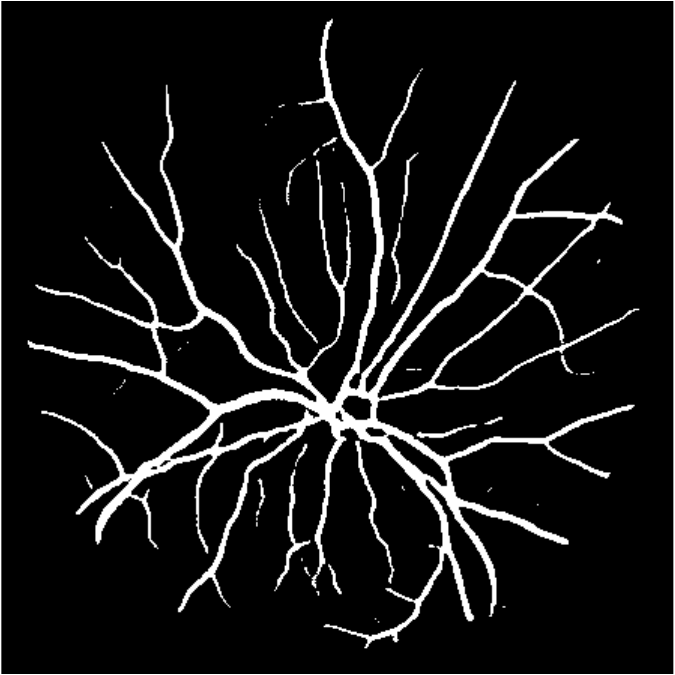}}\hspace{0.05cm}
		\subfloat[\footnotesize ]{\includegraphics[height=1.3cm,width=1.3cm]{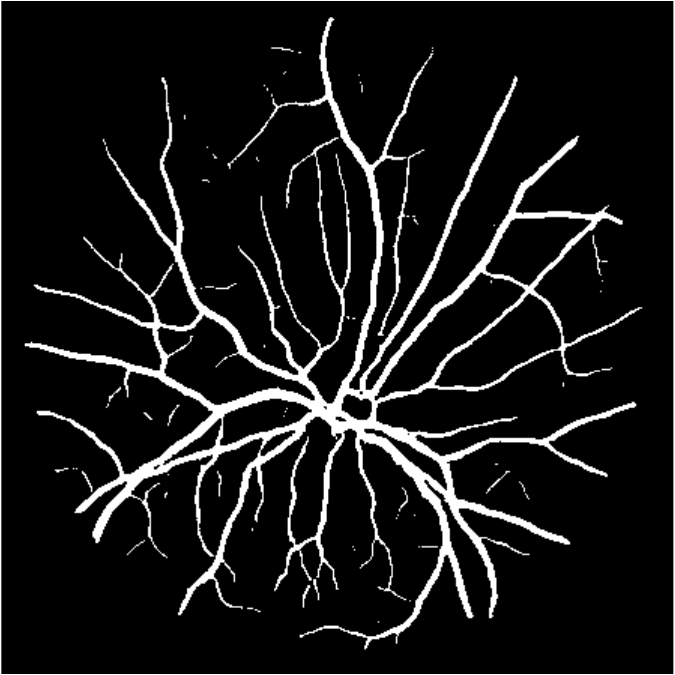}}\hspace{0.05cm}
		\subfloat[\footnotesize ]{\includegraphics[height=1.3cm,width=1.3cm]{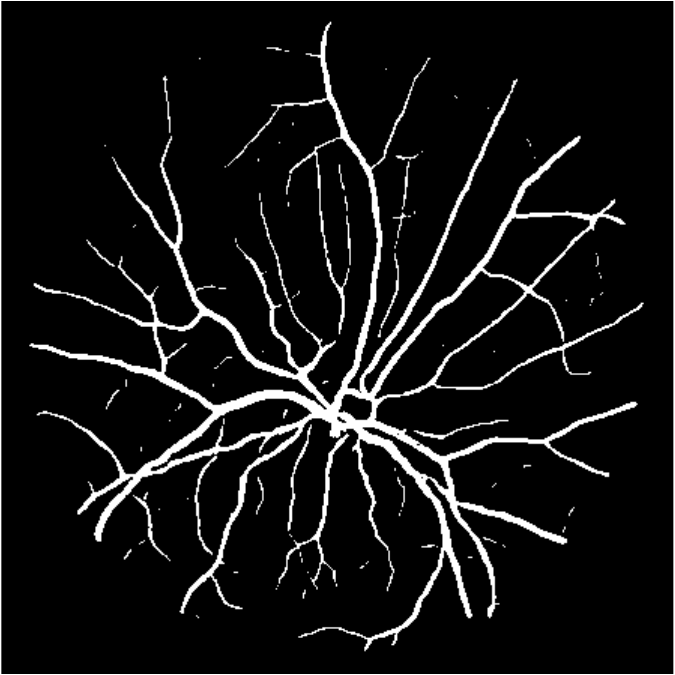}}\hspace{0.05cm}
		\subfloat[\footnotesize ]{\includegraphics[height=1.3cm,width=1.3cm]{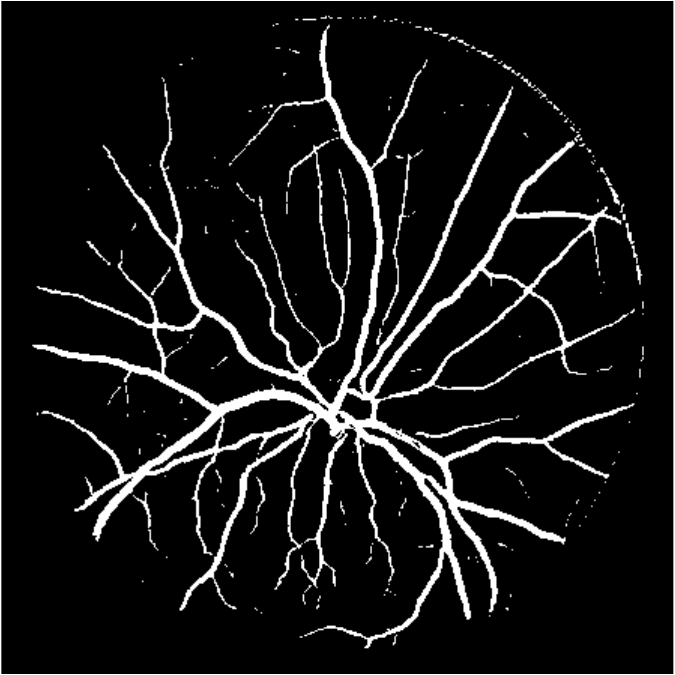}}\hspace{0.05cm}
		\subfloat[\footnotesize ]{\includegraphics[height=1.3cm,width=1.3cm]{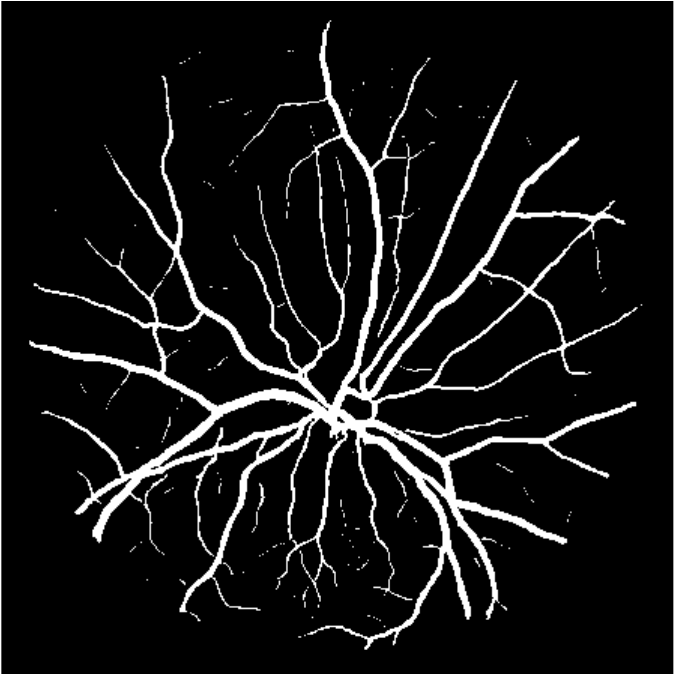}}\hspace{0.05cm}\\
		\vspace{-0.7cm}
		\centering
		\subfloat[\footnotesize (a) ]{\includegraphics[height=1.3cm,width=1.4cm]{DRA_map}}\hspace{0.05cm}
		\subfloat[\footnotesize (b) ]{\includegraphics[height=1.3cm,width=1.3cm]{gray.eps}}\hspace{0.05cm}
		\subfloat[\footnotesize (c) ]{\includegraphics[height=1.3cm,width=1.3cm]{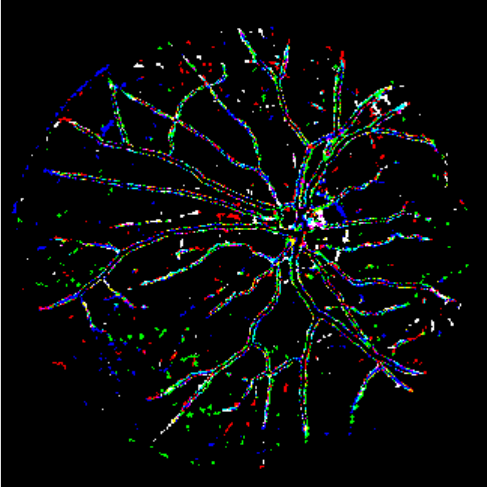}}\hspace{0.05cm}
		\subfloat[\footnotesize (d) ]{\includegraphics[height=1.3cm,width=1.3cm]{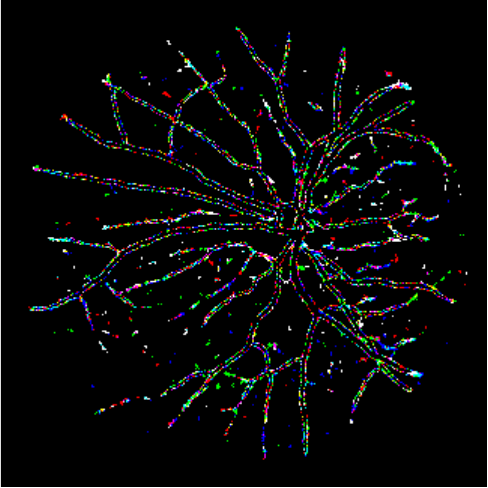}}\hspace{0.05cm}
		\subfloat[\footnotesize (e) ]{\includegraphics[height=1.3cm,width=1.3cm]{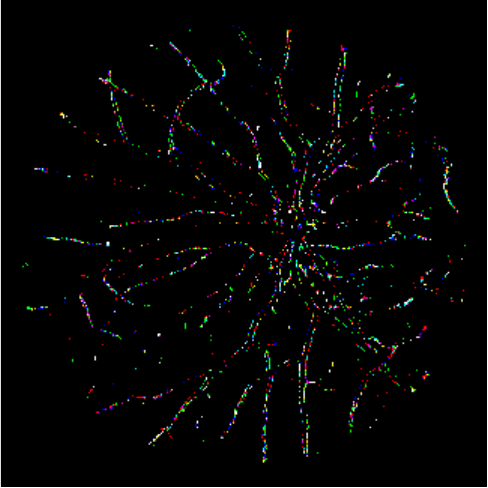}}\hspace{0.05cm}
		\subfloat[\footnotesize (f) ]{\includegraphics[height=1.3cm,width=1.3cm]{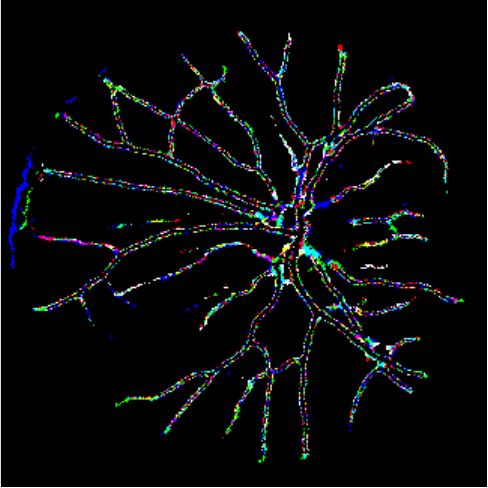}}\hspace{0.05cm}
		\subfloat[\footnotesize (g) ]{\includegraphics[height=1.3cm,width=1.3cm]{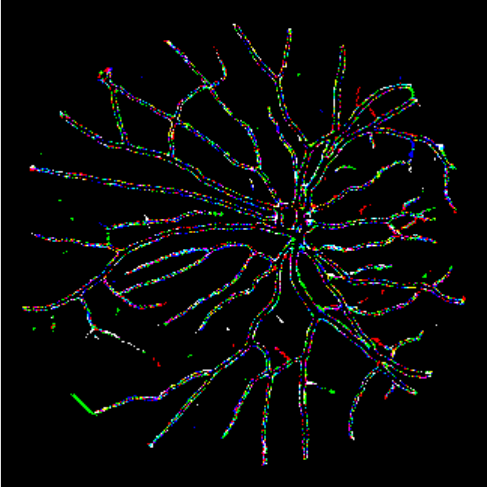}}\hspace{0.05cm}
		\subfloat[\footnotesize (h) ]{\includegraphics[height=1.3cm,width=1.3cm]{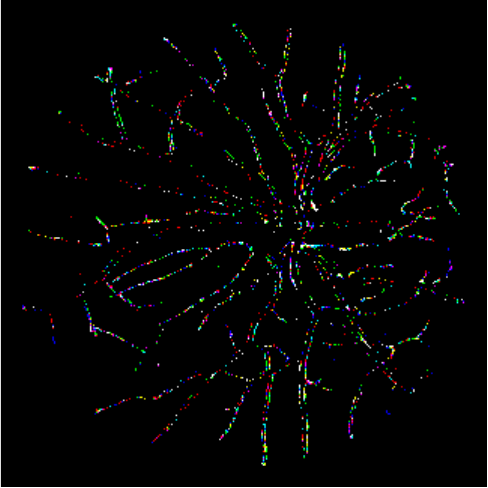}}\hspace{0.05cm}
		\subfloat[\footnotesize (i) ]{\includegraphics[height=1.3cm,width=1.3cm]{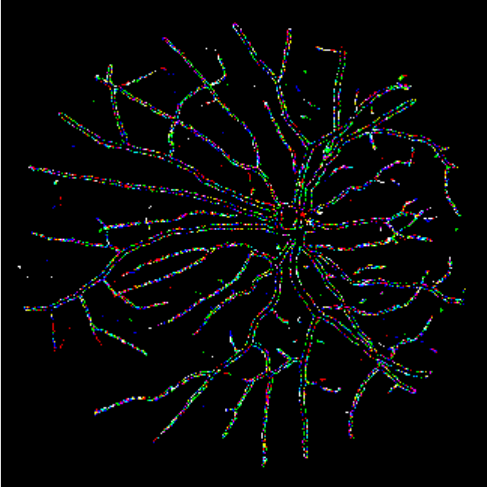}}\hspace{0.05cm}
		\subfloat[\footnotesize (j) ]{\includegraphics[height=1.3cm,width=1.3cm]{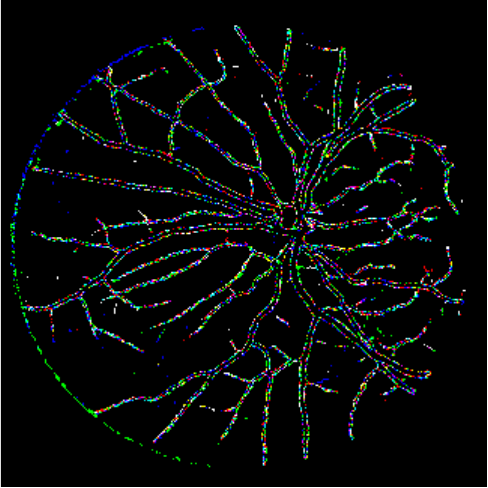}}\hspace{0.05cm}
		\subfloat[\footnotesize (k) ]{\includegraphics[height=1.3cm,width=1.3cm]{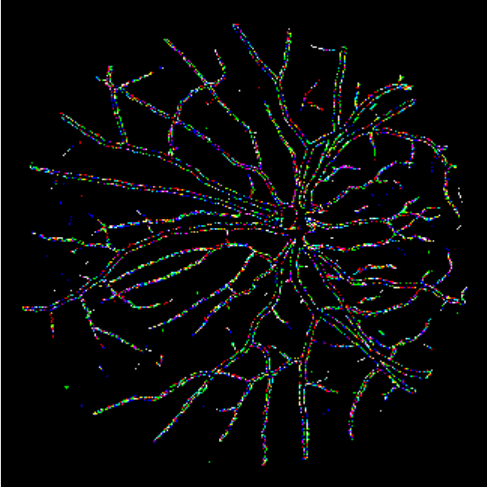}}\hspace{0.05cm}
		\label{fig_9}
		\caption{Visualization results of classical semantic segmentation networks on DRIVE. (a) Raw optical image; (b) Ground Truth; (c) SegNet; (d) SegNet+aug; (e) SegNet(PreCM); (f) ERFNet; (g) ERFNet+aug; (h) ERFNet(PreCM); (i) RIC-CNN; (j) H-Net; (k) E2CNN. The last row are the difference maps, wherein red, green, and blue respectively denote the rotation difference between 0$^\circ$ and 15$^\circ$,  0$^\circ$ and 135$^\circ$, 0$^\circ$ and 255$^\circ$.}
	\end{figure*}
	\begin{figure*}[h!]
		\footnotesize
		\centering
		\captionsetup[subfloat]{position=bottom,labelformat=empty}	
		\subfloat[\footnotesize ]{\includegraphics[height=1.3cm,width=1.3cm]{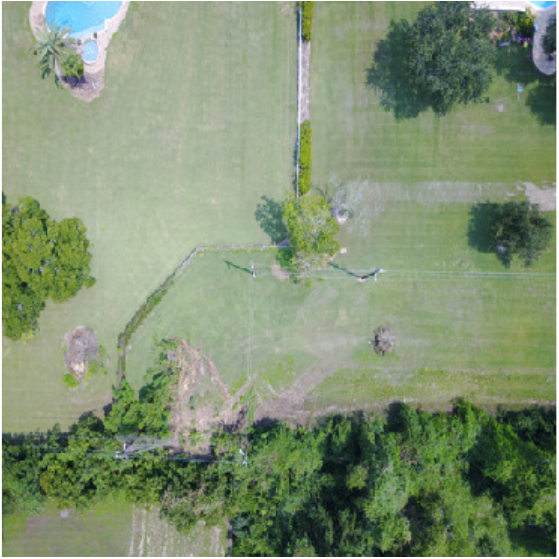}}\hspace{0.05cm}
		\subfloat[\footnotesize ]{\includegraphics[height=1.3cm,width=1.3cm]{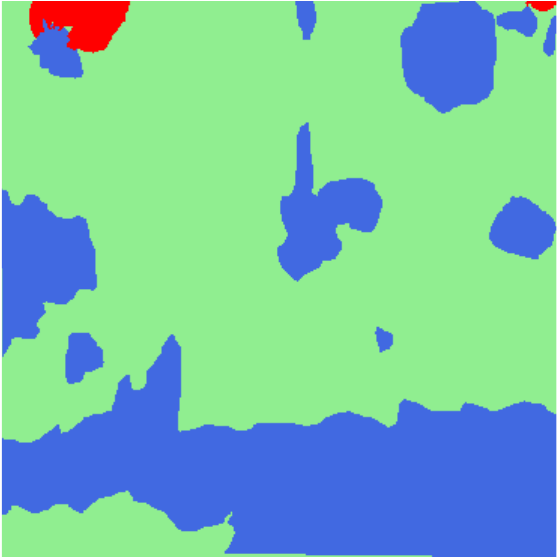}}\hspace{0.05cm}
		\subfloat[\footnotesize ]{\includegraphics[height=1.3cm,width=1.3cm]{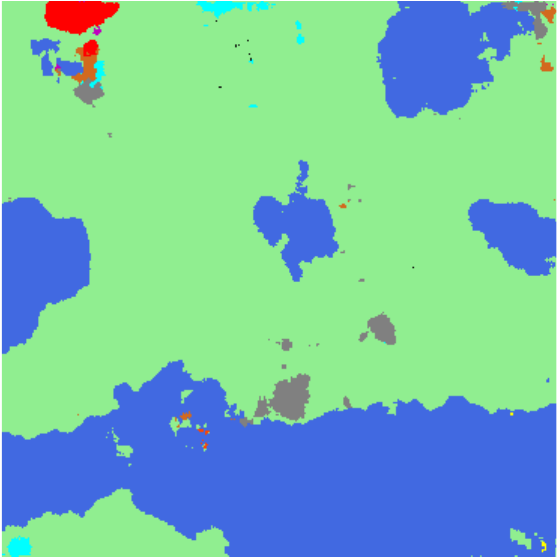}}\hspace{0.05cm}
		\subfloat[\footnotesize ]{\includegraphics[height=1.3cm,width=1.3cm]{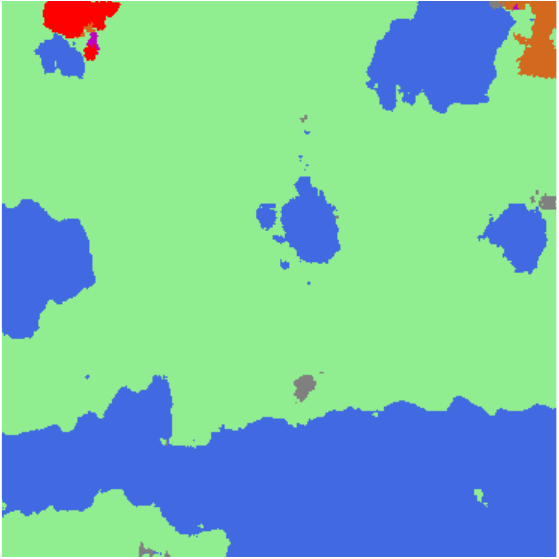}}\hspace{0.05cm}
		\subfloat[\footnotesize ]{\includegraphics[height=1.3cm,width=1.3cm]{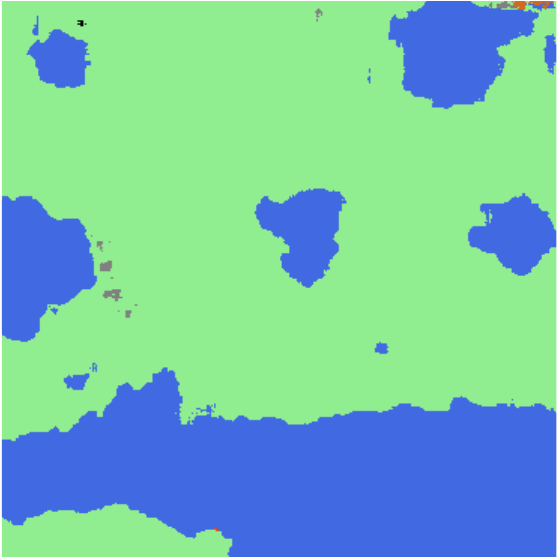}}\hspace{0.05cm}
		\subfloat[\footnotesize ]{\includegraphics[height=1.3cm,width=1.3cm]{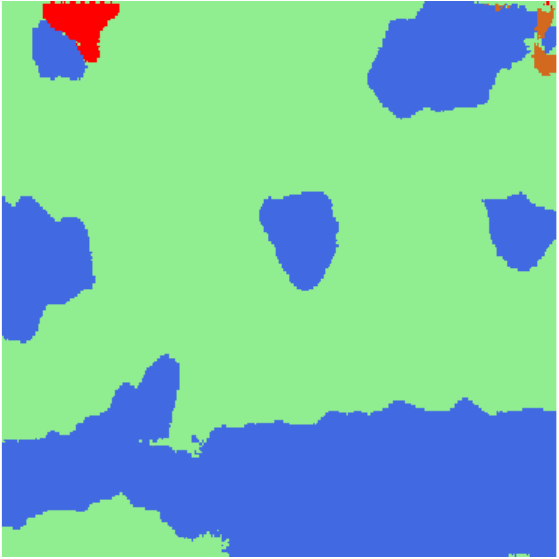}}\hspace{0.05cm}
		\subfloat[\footnotesize ]{\includegraphics[height=1.3cm,width=1.3cm]{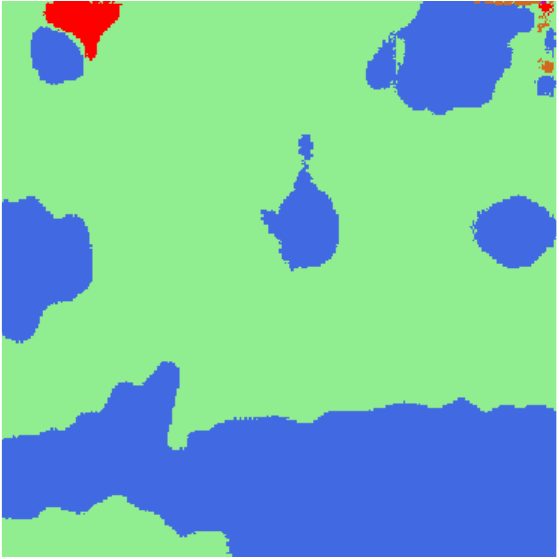}}\hspace{0.05cm}
		\subfloat[\footnotesize ]{\includegraphics[height=1.3cm,width=1.3cm]{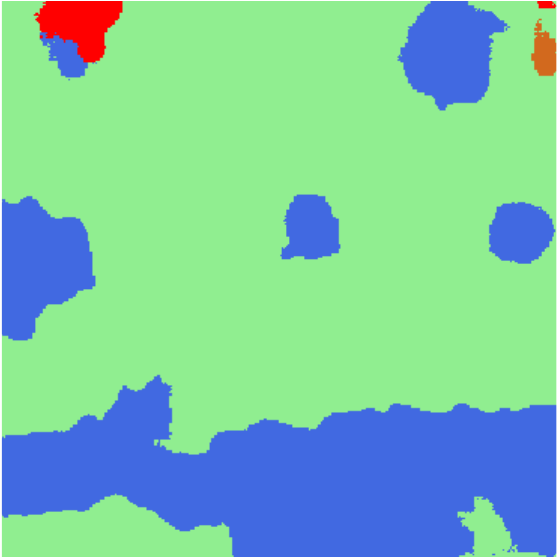}}\hspace{0.05cm}
		\subfloat[\footnotesize ]{\includegraphics[height=1.3cm,width=1.3cm]{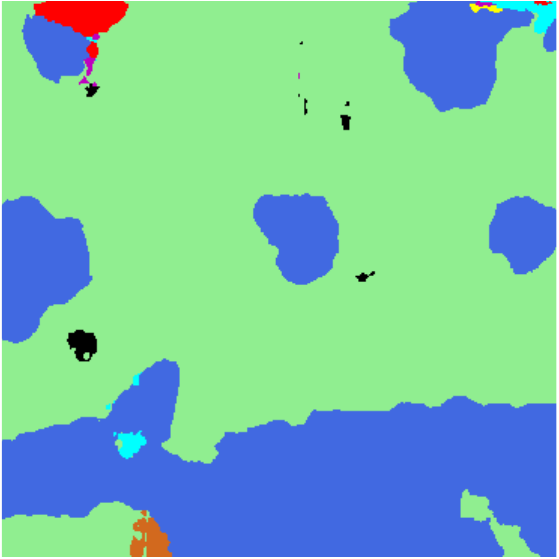}}\hspace{0.05cm}
		\subfloat[\footnotesize ]{\includegraphics[height=1.3cm,width=1.3cm]{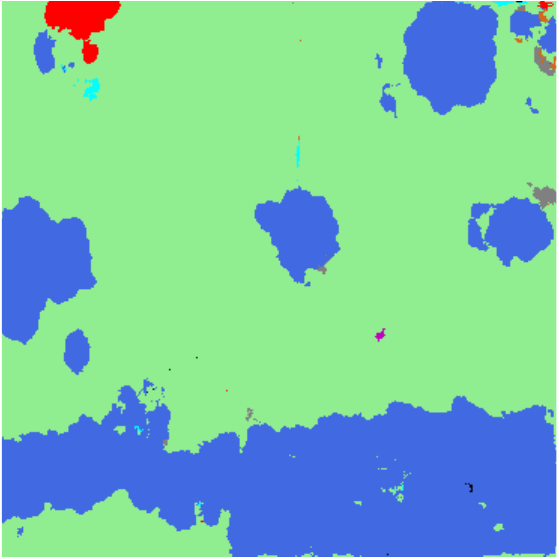}}\hspace{0.05cm}
		\subfloat[\footnotesize ]{\includegraphics[height=1.3cm,width=1.3cm]{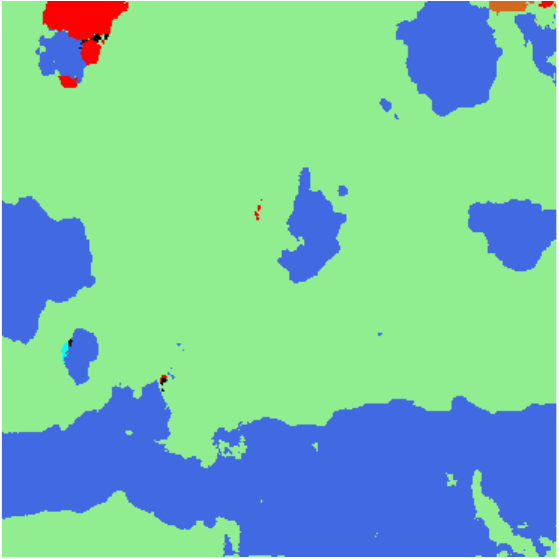}}\hspace{0.05cm}\\
		\vspace{-0.7cm}
		\centering
		\subfloat[\footnotesize ]{\includegraphics[height=1.3cm,width=1.3cm]{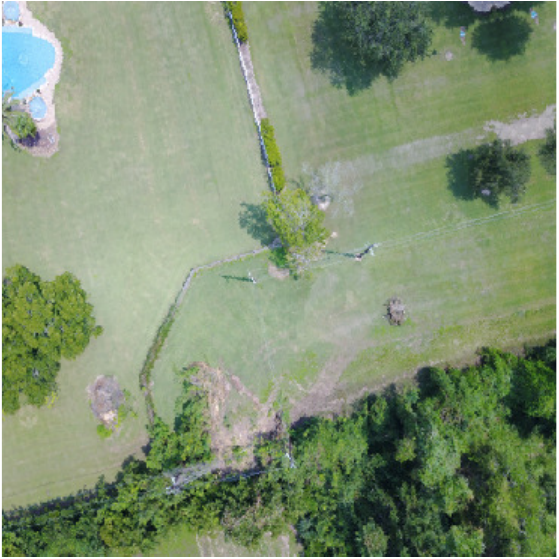}}\hspace{0.05cm}
		\subfloat[\footnotesize ]{\includegraphics[height=1.3cm,width=1.3cm]{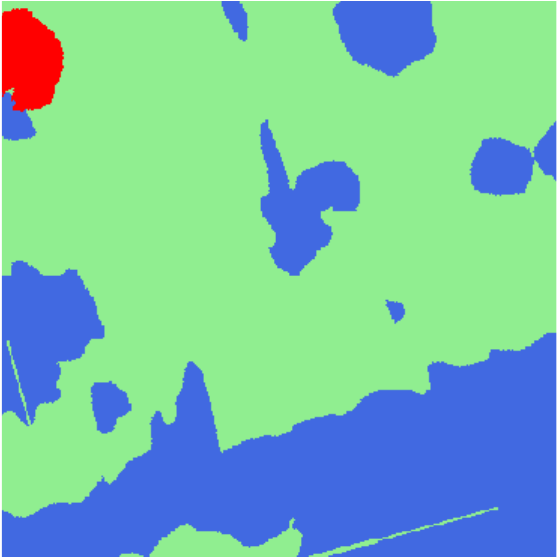}}\hspace{0.05cm}
		\subfloat[\footnotesize ]{\includegraphics[height=1.3cm,width=1.3cm]{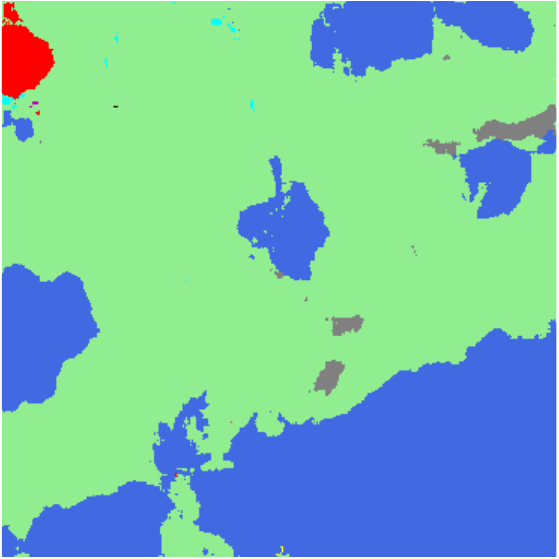}}\hspace{0.05cm}
		\subfloat[\footnotesize ]{\includegraphics[height=1.3cm,width=1.3cm]{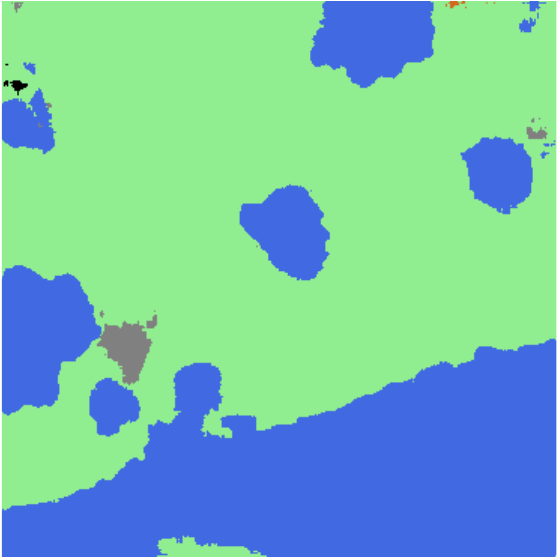}}\hspace{0.05cm}
		\subfloat[\footnotesize ]{\includegraphics[height=1.3cm,width=1.3cm]{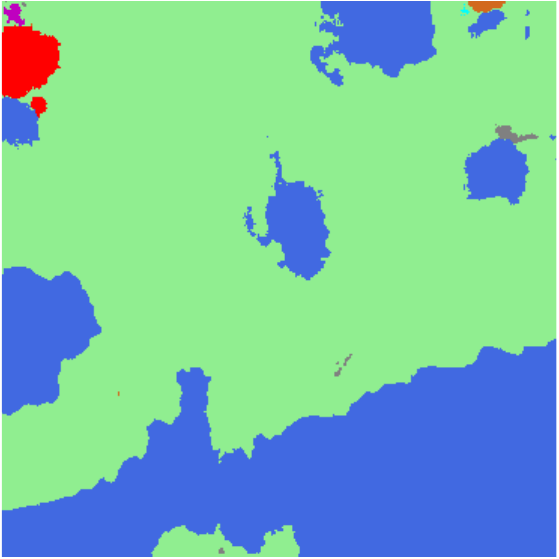}}\hspace{0.05cm}
		\subfloat[\footnotesize ]{\includegraphics[height=1.3cm,width=1.3cm]{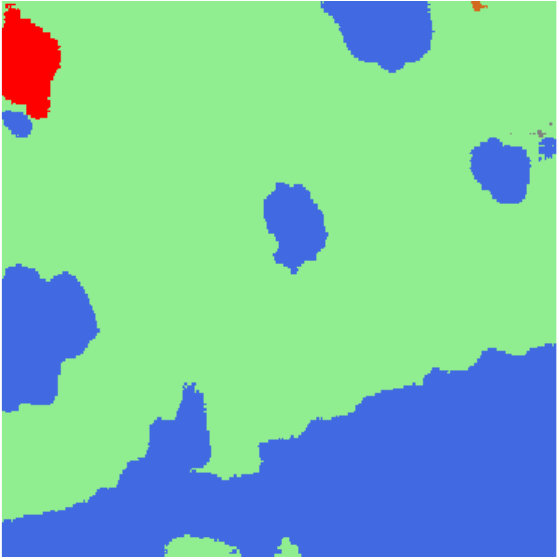}}\hspace{0.05cm}
		\subfloat[\footnotesize ]{\includegraphics[height=1.3cm,width=1.3cm]{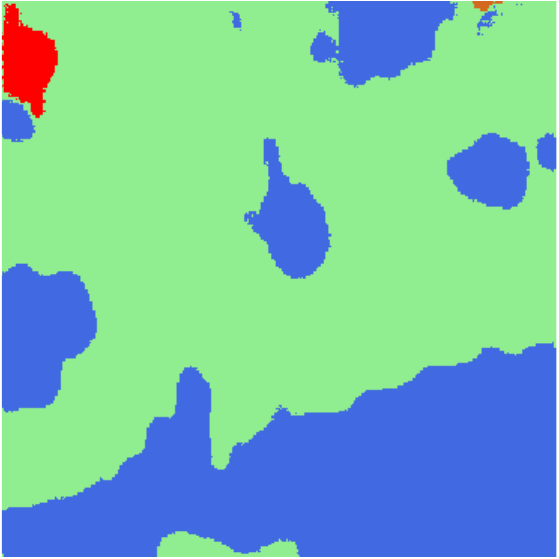}}\hspace{0.05cm}
		\subfloat[\footnotesize ]{\includegraphics[height=1.3cm,width=1.3cm]{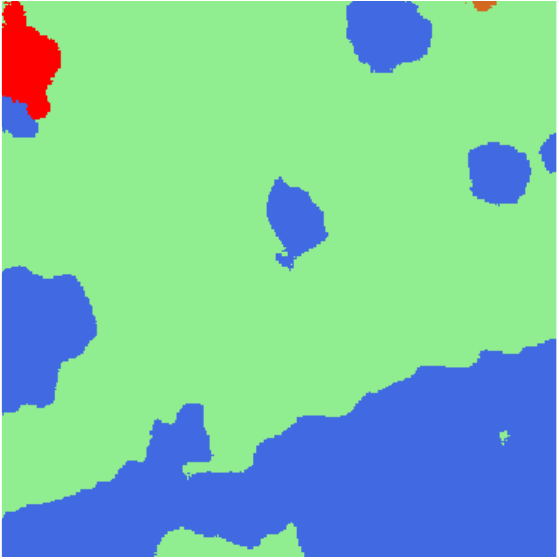}}\hspace{0.05cm}
		\subfloat[\footnotesize ]{\includegraphics[height=1.3cm,width=1.3cm]{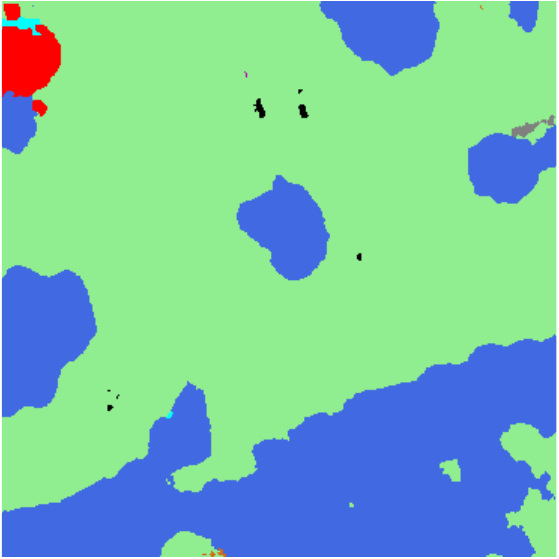}}\hspace{0.05cm}
		\subfloat[\footnotesize ]{\includegraphics[height=1.3cm,width=1.3cm]{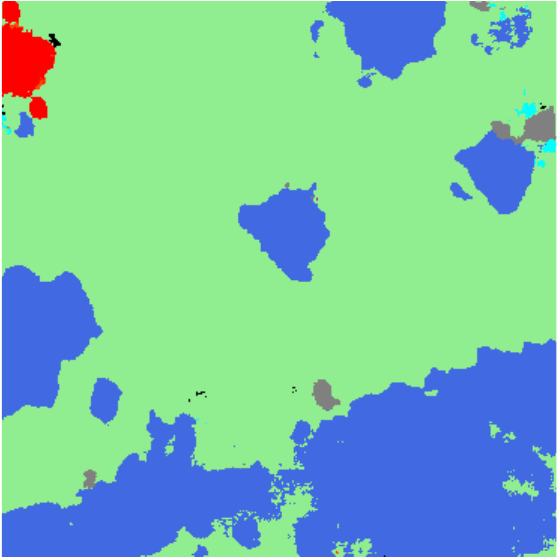}}\hspace{0.05cm}
		\subfloat[\footnotesize ]{\includegraphics[height=1.3cm,width=1.3cm]{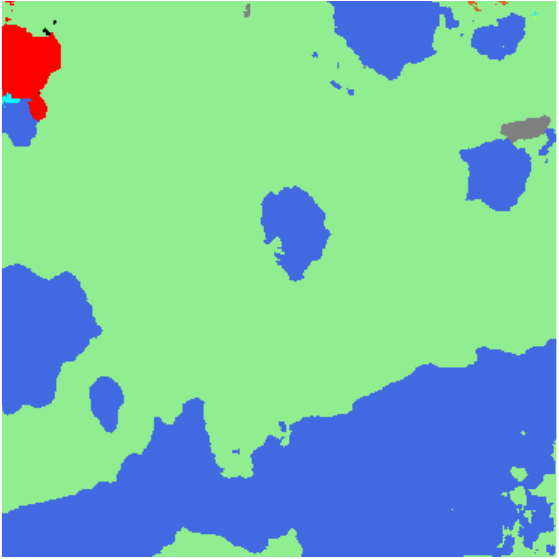}}\hspace{0.05cm}\\
		\vspace{-0.7cm}
		\centering
		\subfloat[\footnotesize ]{\includegraphics[height=1.3cm,width=1.3cm]{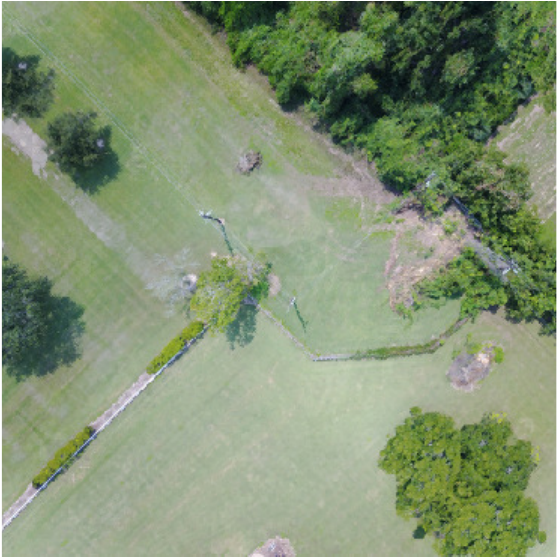}}\hspace{0.05cm}
		\subfloat[\footnotesize ]{\includegraphics[height=1.3cm,width=1.3cm]{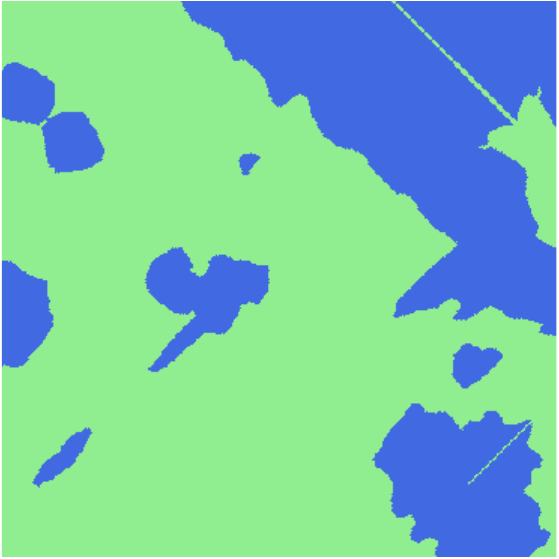}}\hspace{0.05cm}
		\subfloat[\footnotesize ]{\includegraphics[height=1.3cm,width=1.3cm]{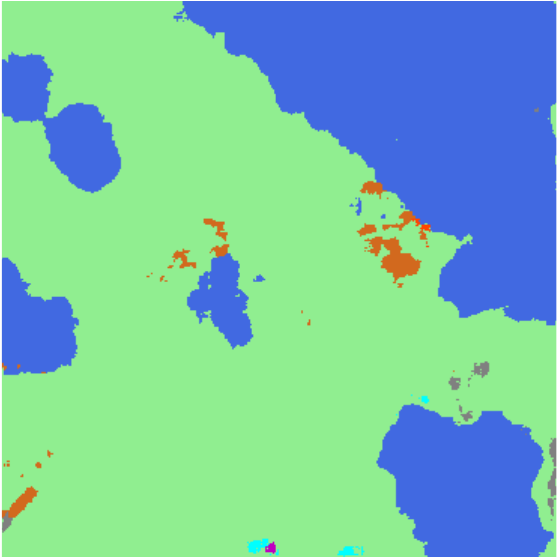}}\hspace{0.05cm}
		\subfloat[\footnotesize ]{\includegraphics[height=1.3cm,width=1.3cm]{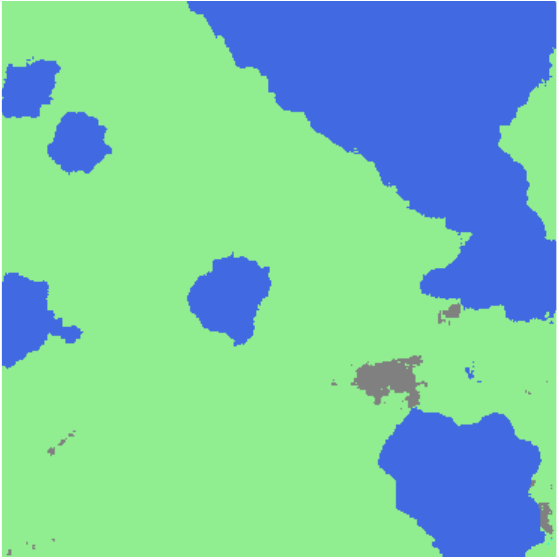}}\hspace{0.05cm}
		\subfloat[\footnotesize ]{\includegraphics[height=1.3cm,width=1.3cm]{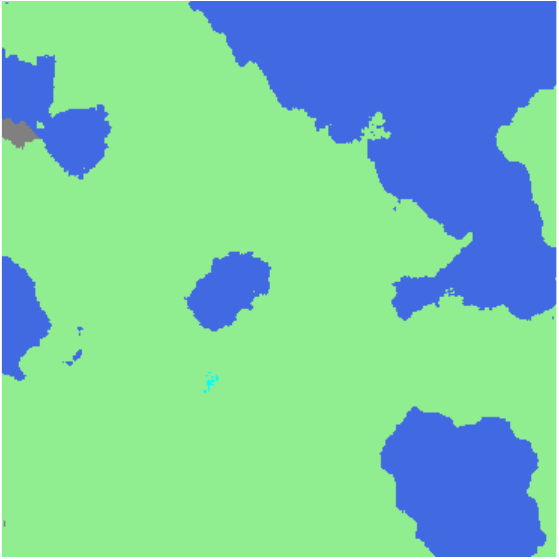}}\hspace{0.05cm}
		\subfloat[\footnotesize ]{\includegraphics[height=1.3cm,width=1.3cm]{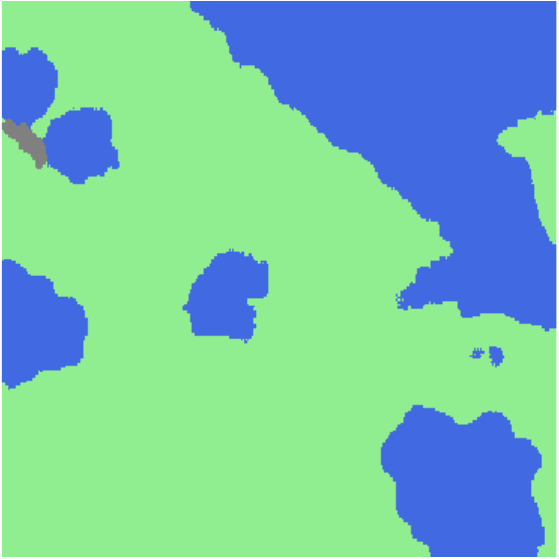}}\hspace{0.05cm}
		\subfloat[\footnotesize ]{\includegraphics[height=1.3cm,width=1.3cm]{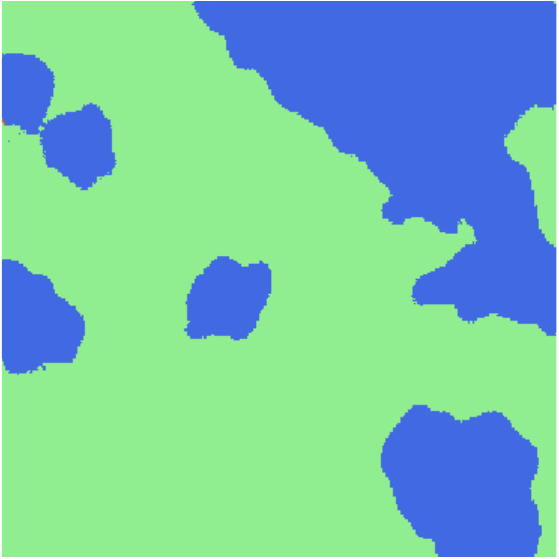}}\hspace{0.05cm}
		\subfloat[\footnotesize ]{\includegraphics[height=1.3cm,width=1.3cm]{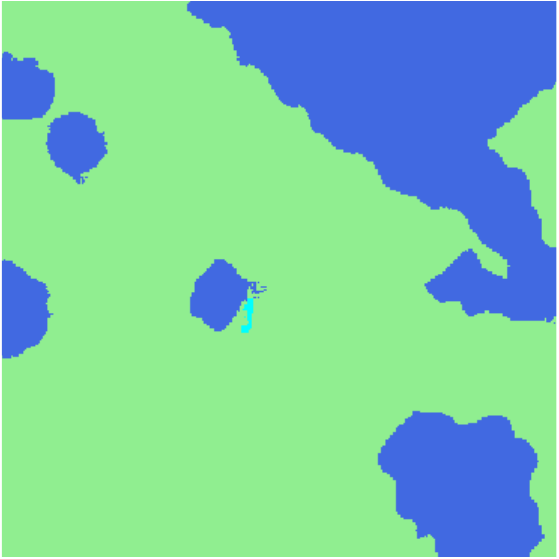}}\hspace{0.05cm}
		\subfloat[\footnotesize ]{\includegraphics[height=1.3cm,width=1.3cm]{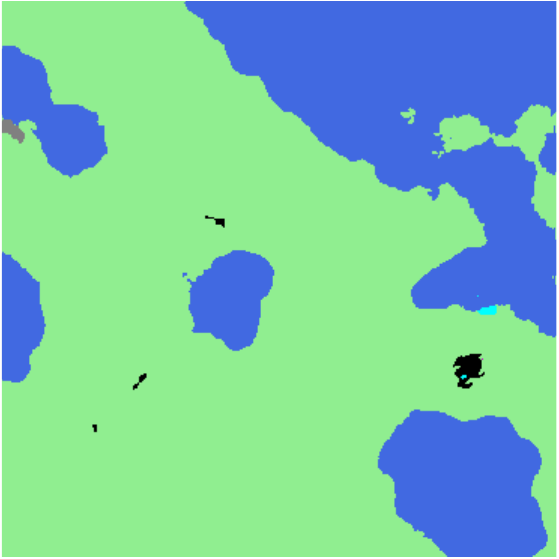}}\hspace{0.05cm}
		\subfloat[\footnotesize ]{\includegraphics[height=1.3cm,width=1.3cm]{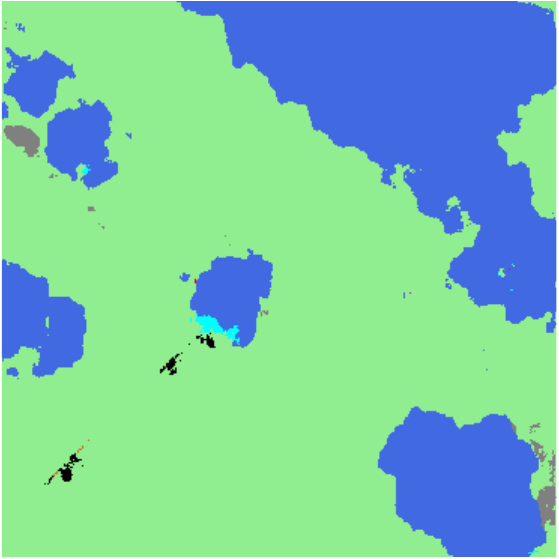}}\hspace{0.05cm}
		\subfloat[\footnotesize ]{\includegraphics[height=1.3cm,width=1.3cm]{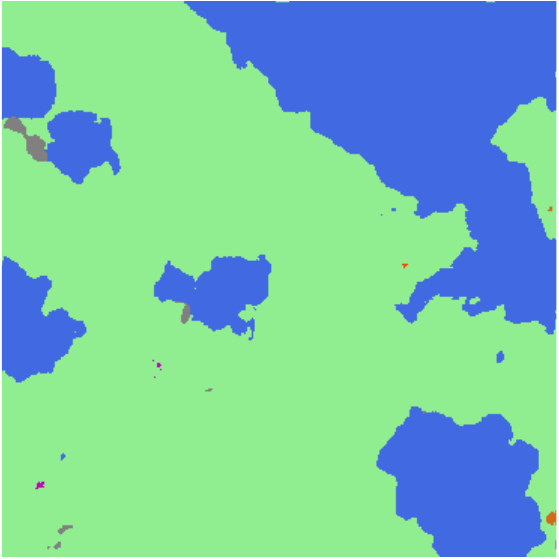}}\hspace{0.05cm}\\
		\vspace{-0.7cm}
		\centering
		\subfloat[\footnotesize ]{\includegraphics[height=1.3cm,width=1.3cm]{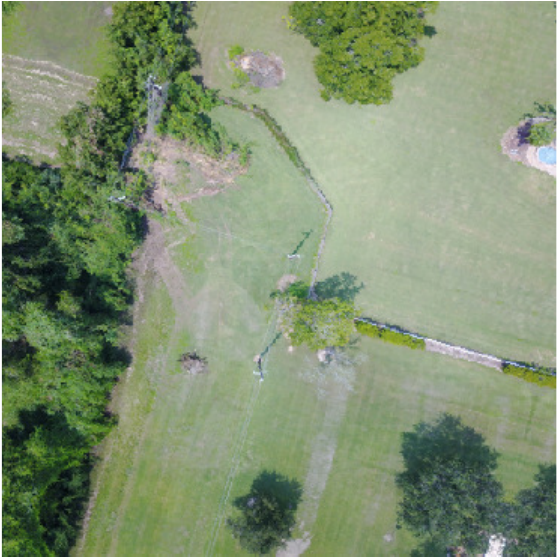}}\hspace{0.05cm}
		\subfloat[\footnotesize ]{\includegraphics[height=1.3cm,width=1.3cm]{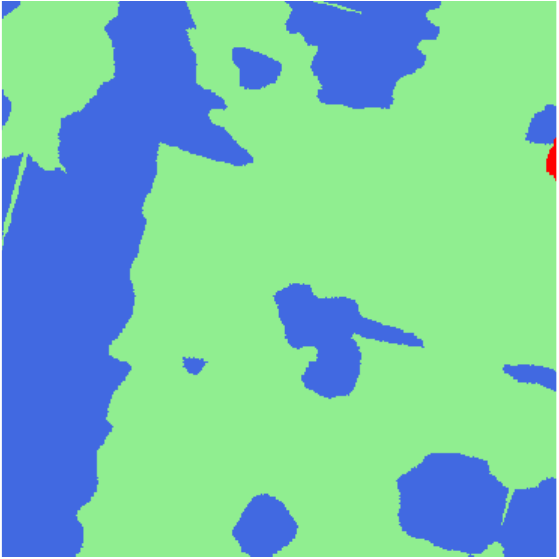}}\hspace{0.05cm}
		\subfloat[\footnotesize ]{\includegraphics[height=1.3cm,width=1.3cm]{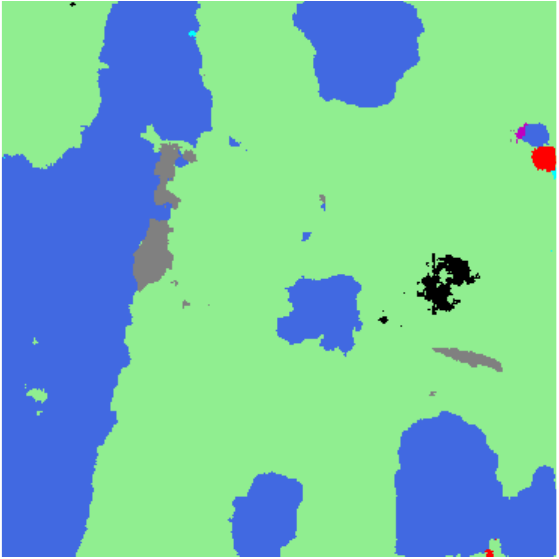}}\hspace{0.05cm}
		\subfloat[\footnotesize ]{\includegraphics[height=1.3cm,width=1.3cm]{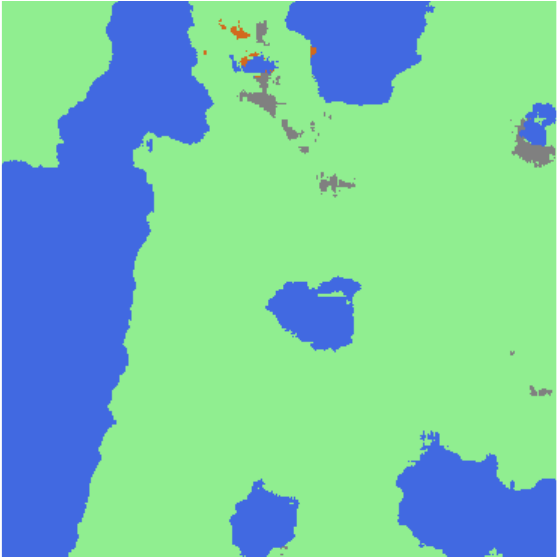}}\hspace{0.05cm}
		\subfloat[\footnotesize ]{\includegraphics[height=1.3cm,width=1.3cm]{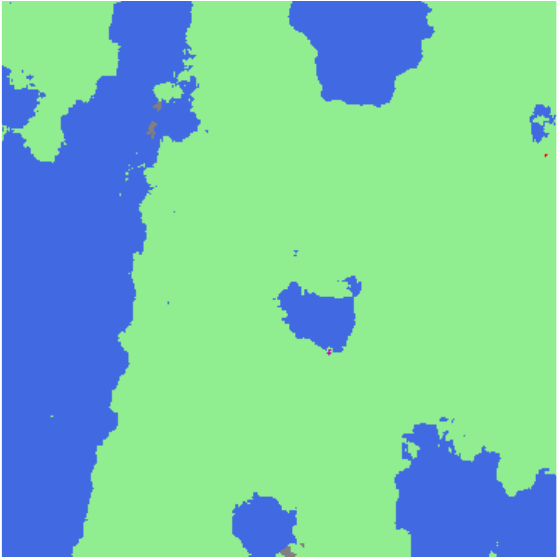}}\hspace{0.05cm}
		\subfloat[\footnotesize ]{\includegraphics[height=1.3cm,width=1.3cm]{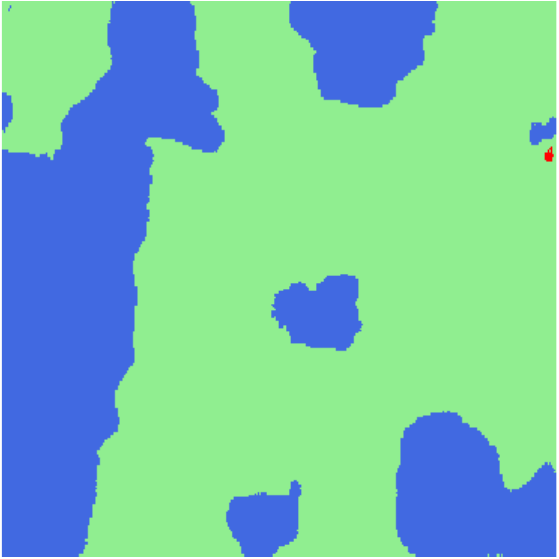}}\hspace{0.05cm}
		\subfloat[\footnotesize ]{\includegraphics[height=1.3cm,width=1.3cm]{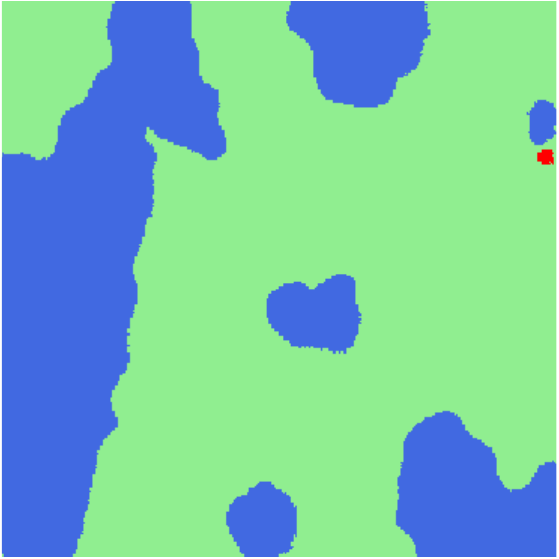}}\hspace{0.05cm}
		\subfloat[\footnotesize ]{\includegraphics[height=1.3cm,width=1.3cm]{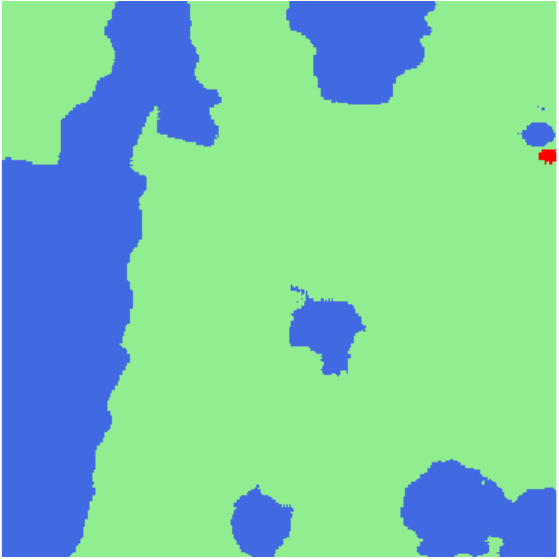}}\hspace{0.05cm}
		\subfloat[\footnotesize ]{\includegraphics[height=1.3cm,width=1.3cm]{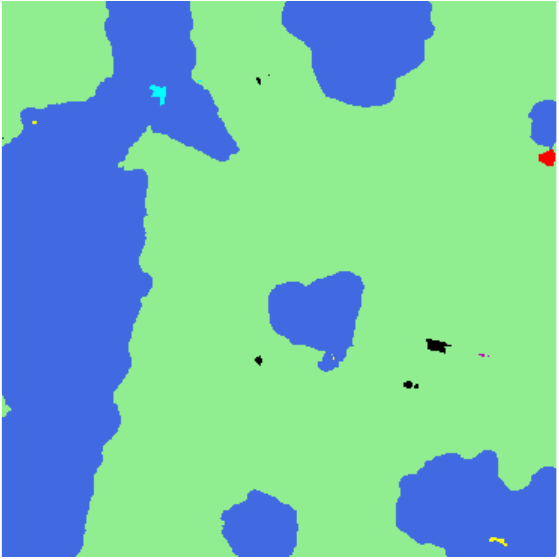}}\hspace{0.05cm}
		\subfloat[\footnotesize ]{\includegraphics[height=1.3cm,width=1.3cm]{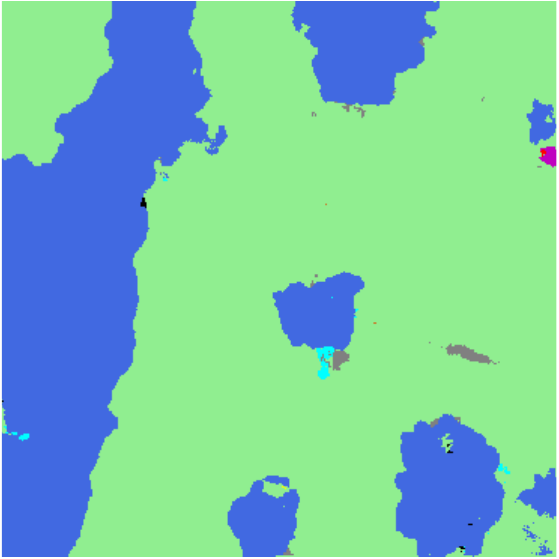}}\hspace{0.05cm}
		\subfloat[\footnotesize ]{\includegraphics[height=1.3cm,width=1.3cm]{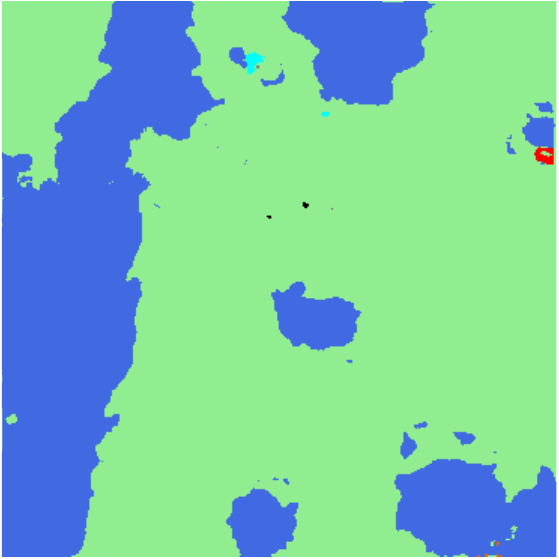}}\hspace{0.05cm}\\
		\vspace{-0.7cm}
		\centering
		\subfloat[\footnotesize (a) ]{\includegraphics[height=1.3cm,width=1.4cm]{DRA_map}}\hspace{0.05cm}
		\subfloat[\footnotesize (b) ]{\includegraphics[height=1.3cm,width=1.3cm]{gray.eps}}\hspace{0.05cm}
		\subfloat[\footnotesize (c) ]{\includegraphics[height=1.3cm,width=1.3cm]{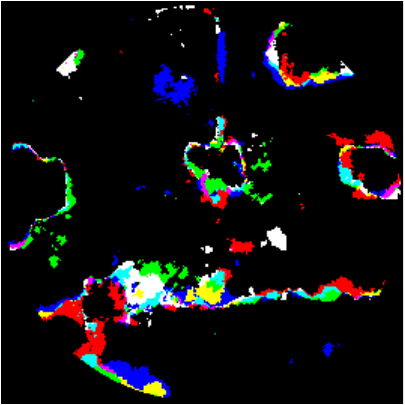}}\hspace{0.05cm}
		\subfloat[\footnotesize (d) ]{\includegraphics[height=1.3cm,width=1.3cm]{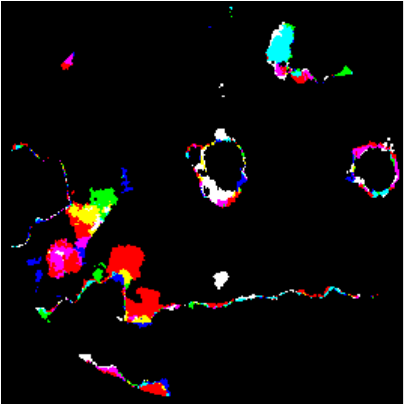}}\hspace{0.05cm}
		\subfloat[\footnotesize (e) ]{\includegraphics[height=1.3cm,width=1.3cm]{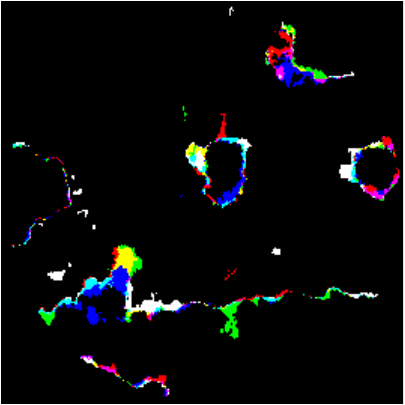}}\hspace{0.05cm}
		\subfloat[\footnotesize (f) ]{\includegraphics[height=1.3cm,width=1.3cm]{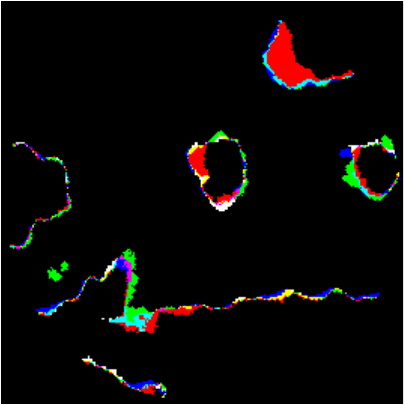}}\hspace{0.05cm}
		\subfloat[\footnotesize (g) ]{\includegraphics[height=1.3cm,width=1.3cm]{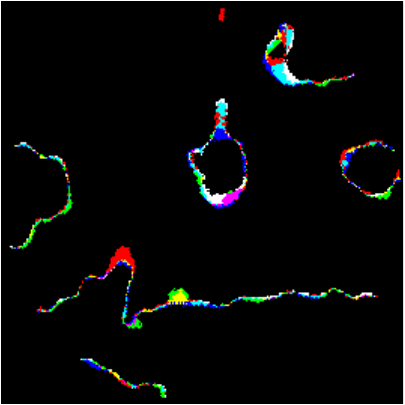}}\hspace{0.05cm}
		\subfloat[\footnotesize (h) ]{\includegraphics[height=1.3cm,width=1.3cm]{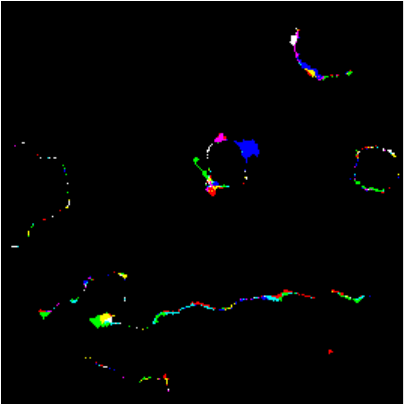}}\hspace{0.05cm}
		\subfloat[\footnotesize (i) ]{\includegraphics[height=1.3cm,width=1.3cm]{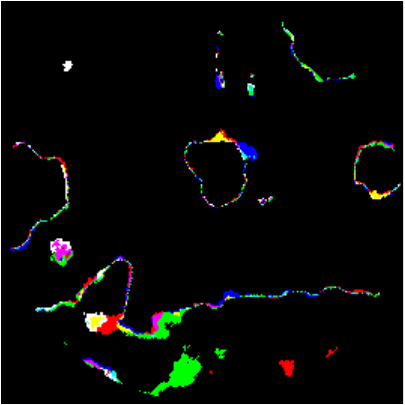}}\hspace{0.05cm}
		\subfloat[\footnotesize (j) ]{\includegraphics[height=1.3cm,width=1.3cm]{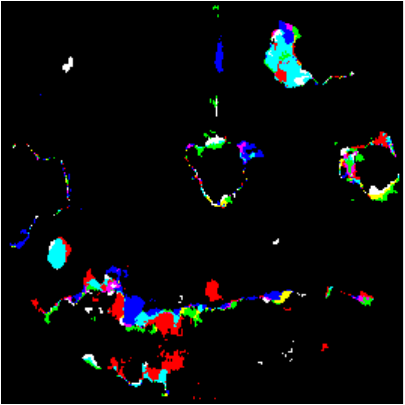}}\hspace{0.05cm}
		\subfloat[\footnotesize (k) ]{\includegraphics[height=1.3cm,width=1.3cm]{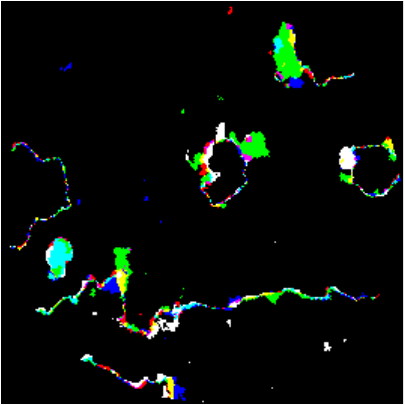}}\hspace{0.05cm}
		\label{fig_10}
		\caption{Visualization results of classical semantic segmentation networks on Floodnet.  (a) Raw optical image; (b) Ground Truth; (c) SegNet; (d) SegNet+aug; (e) SegNet(PreCM); (f) ERFNet; (g) ERFNet+aug; (h) ERFNet(PreCM); (i) RIC-CNN; (j) H-Net; (k) E2CNN. The last row are the difference maps, wherein red, green, and blue respectively denote the rotation difference between 0$^\circ$ and 15$^\circ$,  0$^\circ$ and 135$^\circ$, 0$^\circ$ and 255$^\circ$.}
	\end{figure*}
	
	3) \textit{Floodnet}: Compared to the first two datasets, this dataset is more challenging due to its inclusion of ten different categories. As can be seen from Table I, even in  more complex tasks, the PreCM-based networks demonstrate superior segmentation performance at the initial angle of 0$^\circ$. For the original networks, the segmentation results drop sharply when the test images are rotated. For instance, when the test images are rotated by random angles, the IOU value of U-net decreases abruptly from 76.04\% to 71.02\%. This fully demonstrates that the complexity of task will significantly enhances the original network's sensitivity to rotation. Specifically, as the complexity of samples increases, the model needs to learn and distinguish information from more categories, and its sensitivity to subtle transformations such as rotation may also increase, as these transformations can alter the category assignment of samples. In contrast, the PreCM-based network exhibits significant improvements in segmentation performance and resistance to rotational interference when images are rotated, both at specific and random angles.
	
	Similarly, we conduct a comparative analysis of PreCM-based networks, data augmentation techniques, and other rotation equivariant networks. Observations from Table \uppercase\expandafter{\romannumeral2} reveal that when dealing with complex multi-class segmentation tasks, networks based on data augmentation and PreCM both have their own strengths in segmentation performance. However, our method achieves a lower RD value and comparable segmentation accuracy with less training samples, which fully demonstrates that our method has stronger overall competitiveness compared to data augmentation.
	Besides, the comparison with equivariant networks in Table \uppercase\expandafter{\romannumeral3} once again highlights the advantages of our method in terms of rotation robustness.

	\subsection{Visual Analysis}
			
	Also, to visually verify the effectiveness of PreCM, the output feature maps respectively corresponding to {0$^\circ$}, {15$^\circ$}, {135$^\circ$}, and {255$^\circ$} as well as the difference maps including the rotation differences between {0$^\circ$} and {15$^\circ$} (red), {0$^\circ$} and {135$^\circ$} (green), and {0$^\circ$} and {255$^\circ$} (blue) are shown in Figs. 8-10. Besides, in order to mitigate the potential information loss and edge sensitivity issues associated with zero padding, we employ symmetric padding during the rotation operation \cite{symmetricpadding}. Note that, due to the limitation of article length, we here only present partial visual results of the networks, and the complete results can be further found in supplementary materials.
	
	Fig. 8 presents a comparison of three rotation equivariant networks: RIC-CNN, H-Net, E2CNN, traditional segmentation networks SegNet and ERFNet, along with the results of these two networks after applying data augmentation and the PreCM method on the Satellite Images of Water Bodies dataset. Observations from Fig. 8 reveal that there are significant differences between the output feature maps by the original networks when the test images are rotated at different angles. On the contrary, while networks using data augmentation techniques can achieve superior segmentation results, some noticeable pixel changes are still visible in the difference maps. Similarly, although rotation-equivariant networks improve segmentation accuracy without relying on additional data augmentation by designing equivariant representations to learn feature information from different orientations, these networks still struggle to maintain good consistency when faced with various angles. However, our proposed PreCM-based network not only demonstrates excellent segmentation performance but also shows significantly fewer changed pixels in the difference maps compared to other networks. This fully validates that in the relatively simple binary segmentation dataset, PreCM-based networks have a significant advantage in resisting rotational interference.
	
	Fig. 9 presents the test results of these networks on the DRIVE dataset. Obviously, there are significant differences between the output images at different angles for both the original networks and those incorporating rotation designs. This is primarily attributed to two factors: firstly, the relatively small number of samples results in the network lacking sufficient information to capture features; secondly, the dataset contains numerous fine branch structures, posing a considerable challenge to the network's segmentation performance. However, despite this, the network based on PreCM still outperforms the networks based on data augmentation and existing rotation equivariant networks, both in terms of the accuracy of segmentation results and rotation difference.
	
	Fig. 10 presents the visualization results on the Floodnet dataset. It can be observed that in the more complex segmentation dataset containing ten categories, the segmentation accuracy and consistency of the original semantic segmentation network are significantly lower than those of methods based on data augmentation and PreCM. Notably, although rotation equivariant networks have potential in certain datasets, their performance on this dataset is unsatisfactory, particularly evident in the rotation difference images, highlighting the limitations of current rotation equivariant networks in handling complex tasks. In contrast, networks based on PreCM perform better when dealing with such complex task, although there is still room for performance improvement, which is the focus of our future research.

	\subsection{Ablation Study}
	To deeply analyze the specific impacts of different integration strategies of PreCM on network performance, we further implement an ablation study.
		
\begin{figure}[!t]
		\centering
		\includegraphics[width=3in]{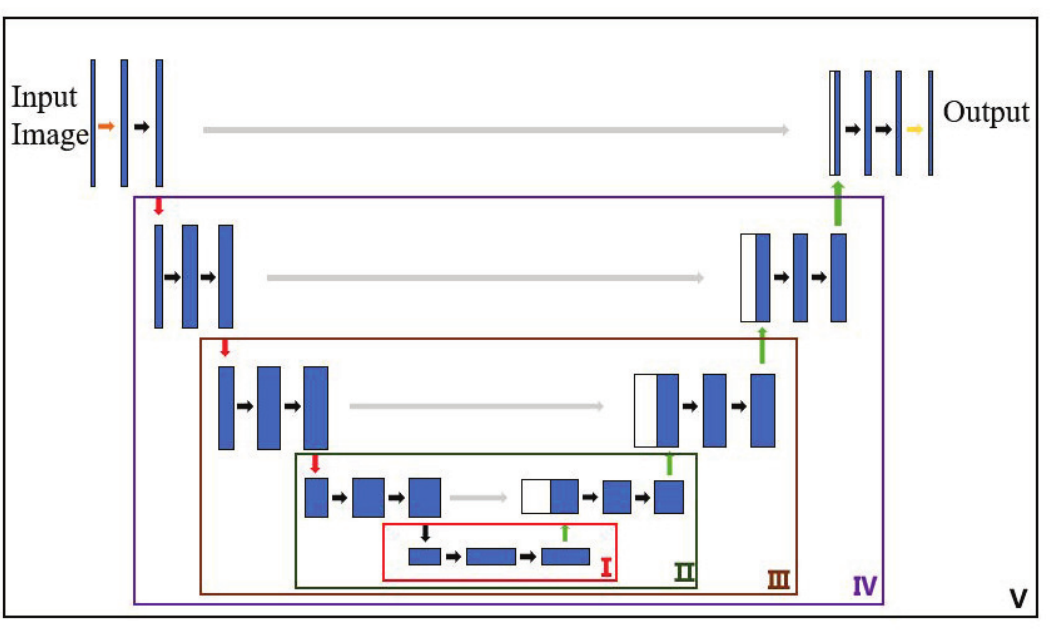}
		\caption{Different integration strategies of PreCM on U-net.}
		\label{fig_11}
	\end{figure}

	Specifically, we conduct tests on the Satellite Images of Water Bodies dataset, taking the popular network U-net as an example and designing five distinct integration schemes, as illustrated in Fig. \ref{fig_11} (labeled as \uppercase\expandafter{\romannumeral1}, \uppercase\expandafter{\romannumeral2}, \uppercase\expandafter{\romannumeral3}, \uppercase\expandafter{\romannumeral4}, and \uppercase\expandafter{\romannumeral5}). In these schemes, we evaluate the experimental performance by replacing the convolution operations within these layers, and compare the obtained results with original U-net. The detailed results under random angle testing are presented in Table \uppercase\expandafter{\romannumeral4}.
	
	 It can be observed that as the number of replaced convolutional layers gradually increases, the segmentation accuracy generally exhibits an upward trend, while the RD value correspondingly decreases step by step, which fully demonstrates the effectiveness of PreCM in capturing rotation information to enhance network performance. Notably, when PreCM replacement is only implemented in a very small number of convolutional layers of the original network (as shown in \uppercase\expandafter{\romannumeral1}), the experimental results do not undergo significant changes, indicating that small-scale replacements cannot bring substantial performance improvements. Furthermore, although the RD value decreases with the increase of replaced layers, its effect is still significantly weaker compared to the case where all convolutional layers are replaced. This result implies that in order to fully exploit the potential of PreCM, it is necessary to replace all the convolutional layers of one network.
	\begin{table}[!t]
		\renewcommand\arraystretch{1.3}
		\begin{center}
			\caption{The quantization results(\%) of different integration strategies of PreCM on U-net.}
			\begin{tabular}{c|c|c|c|c} 
				\hline
				& \textbf{IOU} & \textbf{MIOU}
				& \textbf{DICE} & \textbf{RD}
				\\
				\hline
				U-net & 76.34 & 83.39 & 84.42 &4.17
				\\
				\hline
				\uppercase\expandafter{\romannumeral1} & 76.05 &83.71  &85.09  & 4.20
				\\
				\hline
				\uppercase\expandafter{\romannumeral2} & 78.45 & 85.60 & 86.74 & 3.38
				\\
				\hline
				\uppercase\expandafter{\romannumeral3} & 81.47 & 87.22 & 89.05 & 3.11
				\\
				\hline
				\uppercase\expandafter{\romannumeral4} & 83.22 & 88.40 & 90.29 & 2.73
				\\
				\hline
				\uppercase\expandafter{\romannumeral5} & \pmb{85.19} & \pmb{89.61} & \pmb{91.32} & \pmb{1.62}
				\\
				\hline
			\end{tabular}
		\end{center}
	\end{table}

	\subsection{Other Characteristics}
	In order to comprehensively evaluate the performance of our method in terms of spatial complexity and computational cost, we hereafter select several representative networks (original U-net, the PreCM-based U-net, and three specialized rotation equivariant networks: RIC-CNN, H-Net, and E2CNN) for comparisons and adopt three commonly used evaluation metrics \cite{repvgg}: the number of parameters (\#Params), frames per second (FPS), and GPU memory usage (Memory). The specific results are presented in Table V.
	
	As can be clearly observed from Table V, under the condition of similar parameters, our network outperforms the rotation-equivariant networks in terms of FPS. Although our speed is slower compared to the original U-net, we successfully achieve significant performance improvements, reaching a more balanced overall performance. Additionally, it is worth mentioning that due to the incorporation of multiple convolutional kernels in our design, the PreCM-based U-net occupies more GPU resources than RIC-CNN and the original U-net. However, compared to H-Net and E2CNN, we still maintain a clear advantage. So, from the perspective of comprehensive performance, our network possesses certain competitive advantages.
		\begin{table}[!t]
		\renewcommand\arraystretch{1.3}
		\begin{center}
			\caption{The computational complexity of PreCM.}
			\begin{tabular}{c|c|c|c} 
				\hline
				& \textbf{\#Params(M)} & \textbf{FPS}
				& \textbf{Memory}
				\\
				\hline
				U-net & 3.35 & 149.25 & 1795		\\
				\hline
				U-net(PreCM) & 3.15 & 95.23  & 4010
				\\
				\hline
				RIC-CNN & 3.40 & 94.87 & 2919
				\\
				\hline
				H-Net & 2.96 & 5.81 & 8147
				\\
				\hline
				E2CNN & 3.51 & 17.84 & 8161
				\\
				\hline
				
			\end{tabular}
		\end{center}
	\end{table}
	\begin{figure}[!t]
		\centering
		\includegraphics[width=3in]{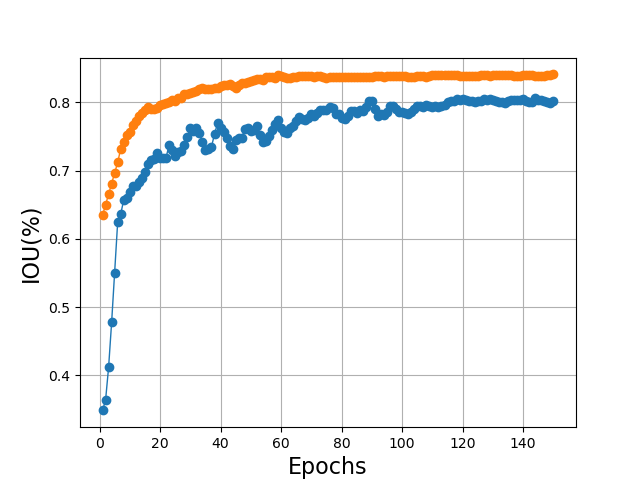}
		\caption{Comparison of convergence speed between original U-net (blue) and PreCM-based U-net (orange).}
	\end{figure}

	Besides, we also evaluated the potential convergence speed advantages of the models before and after PreCM substitution by plotting the curves of IOU versus training epochs. The results are shown in Fig. 12, from which it is evident that the IOU curve of the original U-net model gradually stabilizes after 120 training epochs. However, the U-net model with PreCM substitution, benefiting from its inherent rotation equivariant design, exhibits a significantly improved convergence speed, stabilizing after only 60 training epochs. Directly, this demonstrates the computation advantage of our method.
	
	\section{Conclusion}
	This paper introduces a novel rotation equivariant convolution mode PreCM. Through a series of rigorous mathematical derivations, we have derived rotation equivariant conditions applicable to images of arbitrary sizes and various types of convolutions (including dilated convolution, transposed convolution, asymmetric convolution, and flattened convolution), which can be realized through simple padding operations. To validate the effectiveness of PreCM, we further selected three classic semantic segmentation networks and three lightweight semantic segmentation networks as the basic segmentation architectures. By replacing the convolutional layers in these networks, we conducted an in-depth exploration of the impact of PreCM on segmentation performance. Experimental results on three datasets show that networks employing PreCM not only exhibit superior segmentation performance but also surpass the original networks, data augmentation-based networks, and three other rotation-equivariant networks in terms of segmentation consistency.
	
	However, we must explicitly point out that, although our method can enhance segmentation accuracy and resistance to rotation disturbances without increasing the parameter of networks, it cannot achieve complete rotation equivariance at arbitrary angles except for [0$^\circ$, 90$^\circ$, 180$^\circ$, 270$^\circ$]. Our subsequent research will focus on how to further reduce the errors introduced by random angle rotation in the larger-scale datasets or in the scenarios with complex object interactions.



	{\small
		\bibliographystyle{IEEEtran}
		\bibliography{ref}

\begin{thebibliography}{10}
\providecommand{\url}[1]{#1}
\csname url@samestyle\endcsname
\providecommand{\newblock}{\relax}
\providecommand{\bibinfo}[2]{#2}
\providecommand{\BIBentrySTDinterwordspacing}{\spaceskip=0pt\relax}
\providecommand{\BIBentryALTinterwordstretchfactor}{4}
\providecommand{\BIBentryALTinterwordspacing}{\spaceskip=\fontdimen2\font plus
\BIBentryALTinterwordstretchfactor\fontdimen3\font minus
  \fontdimen4\font\relax}
\providecommand{\BIBforeignlanguage}[2]{{%
\expandafter\ifx\csname l@#1\endcsname\relax
\typeout{** WARNING: IEEEtran.bst: No hyphenation pattern has been}%
\typeout{** loaded for the language `#1'. Using the pattern for}%
\typeout{** the default language instead.}%
\else
\language=\csname l@#1\endcsname
\fi
#2}}
\providecommand{\BIBdecl}{\relax}
\BIBdecl

\bibitem{1}
Z.~Liu, X.~Li, P.~Luo, C.~C. Loy, and X.~Tang, ``Deep learning markov random
  field for semantic segmentation,'' \emph{IEEE Transactions on Pattern
  Analysis and Machine Intelligence}, vol.~40, no.~8, pp. 1814--1828, 2017.

\bibitem{2}
B.~Li, S.~Liu, W.~Xu, and W.~Qiu, ``Real-time object detection and semantic
  segmentation for autonomous driving,'' in \emph{International Symposium on
  Multispectral Image Processing and Pattern Recognition (MIPPR)}, vol. 10608,
  2017, pp. 167--174.

\bibitem{xu2024information}
X.~Xu, T.~Zhang, H.~Liu, W.~Guo, and Z.~Zhang, ``An information-expanding
  network for water body extraction based on \uppercase{U}-net,'' \emph{IEEE
  Geoscience and Remote Sensing Letters}, vol.~21, p. 1502205, 2024.

\bibitem{4}
G.~L. Oliveira, W.~Burgard, and T.~Brox, ``Efficient deep models for monocular
  road segmentation,'' in \emph{IEEE/RSJ International Conference on
  Intelligent Robots and Systems (IROS)}, 2016, pp. 4885--4891.

\bibitem{5}
X.~Tao, D.~Zhang, W.~Ma, X.~Liu, and D.~Xu, ``Automatic metallic surface defect
  detection and recognition with convolutional neural networks,'' \emph{Applied
  Sciences}, vol.~8, no.~9, p. 1575, 2018.

\bibitem{7}
H.~Xu, G.~Zhu, J.~Tain, X.~Zhang, and F.~Peng, ``Image segmentation based on
  support vector machine,'' \emph{Journal of Electronic Science and
  Technology}, vol.~3, no.~3, pp. 226--230, 2005.

\bibitem{8}
B.~Kang and T.~Q. Nguyen, ``Random forest with learned representations for
  semantic segmentation,'' \emph{IEEE Transactions on Image Processing},
  vol.~28, no.~7, pp. 3542--3555, 2019.

\bibitem{9}
Z.~Kato and T.-C. Pong, ``A markov random field image segmentation model for
  color textured images,'' \emph{Image and Vision Computing}, vol.~24, no.~10,
  pp. 1103--1114, 2006.

\bibitem{10}
N.~Plath, M.~Toussaint, and S.~Nakajima, ``Multi-class image segmentation using
  conditional random fields and global classification,'' in \emph{International
  Conference on Machine Learning}, 2009, pp. 817--824.

\bibitem{wei2023rotational}
X.~Wei, S.~Su, Y.~Wei, and X.~Lu, ``{Rotational Convolution}: Rethinking
  convolution for downside fisheye images,'' \emph{IEEE Transactions on Image
  Processing}, 2023.

\bibitem{11}
J.~Long, E.~Shelhamer, and T.~Darrell, ``Fully convolutional networks for
  semantic segmentation,'' in \emph{Proceedings of the IEEE Conference on
  Computer Vision and Pattern Recognition}, 2015, pp. 3431--3440.

\bibitem{21}
L.-C. Chen, G.~Papandreou, I.~Kokkinos, K.~Murphy, and A.~L. Yuille, ``Deeplab:
  Semantic image segmentation with deep convolutional nets, atrous convolution,
  and fully connected crfs,'' \emph{IEEE Transactions on Pattern Analysis and
  Machine Intelligence}, vol.~40, no.~4, pp. 834--848, 2017.

\bibitem{6}
M.~Yang, K.~Yu, C.~Zhang, Z.~Li, and K.~Yang, ``Denseaspp for semantic
  segmentation in street scenes,'' in \emph{Proceedings of the IEEE Conference
  on Computer Vision and Pattern Recognition}, 2018, pp. 3684--3692.

\bibitem{15}
D.~K. Gupta, D.~Arya, and E.~Gavves, ``Rotation equivariant siamese networks
  for tracking,'' in \emph{Proceedings of the IEEE Conference on Computer
  Vision and Pattern Recognition}, 2021, pp. 12\,362--12\,371.

\bibitem{della2019deep}
L.~Della~Libera, V.~Golkov, Y.~Zhu, A.~Mielke, and D.~Cremers, ``Deep learning
  for 2d and 3d rotatable data: An overview of methods,'' \emph{arXiv preprint
  arXiv:1910.14594}, 2019.

\bibitem{17}
A.~Miko{\l}ajczyk and M.~Grochowski, ``Data augmentation for improving deep
  learning in image classification problem,'' in \emph{International
  Interdisciplinary PhD Workshop (IIPhDW)}.\hskip 1em plus 0.5em minus
  0.4em\relax IEEE, 2018, pp. 117--122.

\bibitem{choi}
J.~Choi, T.~Kim, and C.~Kim, ``Self-ensembling with gan-based data augmentation
  for domain adaptation in semantic segmentation,'' in \emph{Proceedings of the
  IEEE International Conference on Computer Vision}, 2019, pp. 6830--6840.

\bibitem{olsson}
V.~Olsson, W.~Tranheden, J.~Pinto, and L.~Svensson, ``Classmix:
  Segmentation-based data augmentation for semi-supervised learning,'' in
  \emph{Proceedings of the IEEE Winter Conference on Applications of Computer
  Vision}, 2021, pp. 1369--1378.

\bibitem{19}
D.~Laptev, N.~Savinov, J.~M. Buhmann, and M.~Pollefeys, ``Ti-pooling:
  transformation-invariant pooling for feature learning in convolutional neural
  networks,'' in \emph{Proceedings of the IEEE Conference on Computer Vision
  and Pattern Recognition}, 2016, pp. 289--297.

\bibitem{fei2024rotation}
J.~Fei and Z.~Deng, ``Rotation invariance and equivariance in 3d deep learning:
  a survey,'' \emph{Artificial Intelligence Review}, vol.~57, no.~7, p. 168,
  2024.

\bibitem{14}
T.~Cohen and M.~Welling, ``Group equivariant convolutional networks,'' in
  \emph{International Conference on Machine Learning}.\hskip 1em plus 0.5em
  minus 0.4em\relax ICML, 2016, pp. 2990--2999.

\bibitem{16}
J.~Li, Z.~Yang, H.~Liu, and D.~Cai, ``Deep rotation equivariant network,''
  \emph{Neurocomputing}, vol. 290, pp. 26--33, 2018.

\bibitem{romero2020attentive}
D.~Romero, E.~Bekkers, J.~Tomczak, and M.~Hoogendoorn, ``Attentive group
  equivariant convolutional networks,'' in \emph{International Conference on
  Machine Learning}.\hskip 1em plus 0.5em minus 0.4em\relax PMLR, 2020, pp.
  8188--8199.

\bibitem{he2021efficient}
L.~He, Y.~Chen, Y.~Dong, Y.~Wang, Z.~Lin \emph{et~al.}, ``Efficient equivariant
  network,'' \emph{Advances in Neural Information Processing Systems}, vol.~34,
  pp. 5290--5302, 2021.

\bibitem{marcos}
D.~Marcos, M.~Volpi, N.~Komodakis, and D.~Tuia, ``Rotation equivariant vector
  field networks,'' in \emph{Proceedings of the IEEE International Conference
  on Computer Vision}, 2017, pp. 5048--5057.

\bibitem{mo2024ric}
H.~Mo and G.~Zhao, ``{RIC-CNN}: rotation-invariant coordinate convolutional
  neural network,'' \emph{Pattern Recognition}, vol. 146, p. 109994, 2024.

\bibitem{fu2024rotation}
J.~Fu, Q.~Xie, D.~Meng, and Z.~Xu, ``Rotation equivariant proximal operator for
  deep unfolding methods in image restoration,'' \emph{IEEE Transactions on
  Pattern Analysis and Machine Intelligence}, 2024.

\bibitem{20}
W.~Sun and R.~Wang, ``Fully convolutional networks for semantic segmentation of
  very high resolution remotely sensed images combined with dsm,'' \emph{IEEE
  Geoscience and Remote Sensing Letters}, vol.~15, no.~3, pp. 474--478, 2018.

\bibitem{12}
O.~Ronneberger, P.~Fischer, and T.~Brox, ``U-net: Convolutional networks for
  biomedical image segmentation,'' in \emph{Medical Image Computing and
  Computer-Assisted Intervention--MICCAI}.\hskip 1em plus 0.5em minus
  0.4em\relax Springer, 2015, pp. 234--241.

\bibitem{23}
V.~Badrinarayanan, A.~Kendall, and R.~Cipolla, ``{SegNet}: A deep convolutional
  encoder-decoder architecture for image segmentation,'' \emph{IEEE
  Transactions on Pattern Analysis and Machine Intelligence}, vol.~39, no.~12,
  pp. 2481--2495, 2017.

\bibitem{24}
H.~Zhao, J.~Shi, X.~Qi, X.~Wang, and J.~Jia, ``Pyramid scene parsing network,''
  in \emph{Proceedings of the IEEE Conference on Computer Vision and Pattern
  Recognition}, 2017, pp. 2881--2890.

\bibitem{chen}
L.-C. Chen, G.~Papandreou, I.~Kokkinos, K.~Murphy, and A.~L. Yuille, ``Deeplab:
  Semantic image segmentation with deep convolutional nets, atrous convolution,
  and fully connected crfs,'' \emph{IEEE Transactions on Pattern Analysis and
  Machine Intelligence}, vol.~40, no.~4, pp. 834--848, 2017.

\bibitem{13}
A.~Paszke, A.~Chaurasia, S.~Kim, and E.~Culurciello, ``{ENet}: A deep neural
  network architecture for real-time semantic segmentation,'' \emph{Proceedings
  of the IEEE Conference on Computer Vision and Pattern Recognition}, 2016.

\bibitem{25}
E.~Romera, J.~M. Alvarez, L.~M. Bergasa, and R.~Arroyo, ``{ERFNet}: Efficient
  residual factorized convnet for real-time semantic segmentation,'' \emph{IEEE
  Transactions on Intelligent Transportation Systems}, vol.~19, no.~1, pp.
  263--272, 2017.

\bibitem{26}
C.~Yu, C.~Gao, J.~Wang, G.~Yu, C.~Shen, and N.~Sang, ``{BiSeNet V2}: Bilateral
  network with guided aggregation for real-time semantic segmentation,''
  \emph{International Journal of Computer Vision}, vol. 129, pp. 3051--3068,
  2021.

\bibitem{wang2019lednet}
Y.~Wang, Q.~Zhou, J.~Liu, J.~Xiong, G.~Gao, X.~Wu, and L.~J. Latecki,
  ``{LEDN}et: A lightweight encoder-decoder network for real-time semantic
  segmentation,'' in \emph{IEEE International Conference on Image Processing
  (ICIP)}, 2019, pp. 1860--1864.

\bibitem{worrall2017harmonic}
D.~E. Worrall, S.~J. Garbin, D.~Turmukhambetov, and G.~J. Brostow, ``Harmonic
  networks: Deep translation and rotation equivariance,'' in \emph{Proceedings
  of the IEEE conference on computer vision and pattern recognition}, 2017, pp.
  5028--5037.

\bibitem{29}
J.~Han, J.~Ding, N.~Xue, and G.-S. Xia, ``{ReDet}: A rotation-equivariant
  detector for aerial object detection,'' in \emph{Proceedings of the IEEE
  Conference on Computer Vision and Pattern Recognition}, 2021, pp. 2786--2795.

\bibitem{weiler2019general}
M.~Weiler and G.~Cesa, ``{General E(2)-equivariant steerable CNNs},''
  \emph{Advances in neural information processing systems}, vol.~32, 2019.

\bibitem{ma2020pcfnet}
Y.~Ma, Y.~Luo, and Z.~Yang, ``{PCFNet}: Deep neural network with predefined
  convolutional filters,'' \emph{Neurocomputing}, vol. 382, pp. 32--39, 2020.

\bibitem{30}
V.~Dumoulin and F.~Visin, ``A guide to convolution arithmetic for deep
  learning,'' \emph{International Conference on Machine Learning}, 2016.

\bibitem{33}
F.~Escobar, ``Satellite images of water bodies,''
  \url{https://www.cvmart.net/dataSets/detail/736}.

\bibitem{32}
J.~Staal, M.~D. Abr{\`a}moff, M.~Niemeijer, M.~A. Viergever, and
  B.~Van~Ginneken, ``Ridge-based vessel segmentation in color images of the
  retina,'' \emph{IEEE Transactions on Medical Imaging}, vol.~23, no.~4, pp.
  501--509, 2004.

\bibitem{31}
M.~Rahnemoonfar, T.~Chowdhury, A.~Sarkar, D.~Varshney, M.~Yari, and R.~R.
  Murphy, ``Floodnet: A high resolution aerial imagery dataset for post flood
  scene understanding,'' \emph{IEEE Access}, vol.~9, pp. 89\,644--89\,654,
  2021.

\bibitem{34}
S.~K. McFeeters, ``The use of the normalized difference water index (ndwi) in
  the delineation of open water features,'' \emph{Remote Sensing}, vol.~17,
  no.~7, pp. 1425--1432, 1996.

\bibitem{35}
P.~Shi, M.~Duan, L.~Yang, W.~Feng, L.~Ding, and L.~Jiang, ``An improved u-net
  image segmentation method and its application for metallic grain size
  statistics,'' \emph{Materials}, vol.~15, no.~13, p. 4417, 2022.

\bibitem{symmetricpadding}
S.~Wu, G.~Wang, P.~Tang, F.~Chen, and L.~Shi, ``Convolution with even-sized
  kernels and symmetric padding,'' \emph{Advances in Neural Information
  Processing Systems}, vol.~32, 2019.

\bibitem{repvgg}
X.~Ding, X.~Zhang, N.~Ma, J.~Han, G.~Ding, and J.~Sun, ``{RepVGG}: Making
  vgg-style convnets great again,'' in \emph{Proceedings of the IEEE Conference
  on Computer Vision and Pattern Recognition}, 2021, pp. 13\,728--13\,737.

\end{thebibliography}
	}

	\begin{IEEEbiography}[{\includegraphics[width=1in,height=1.25in,clip,keepaspectratio]{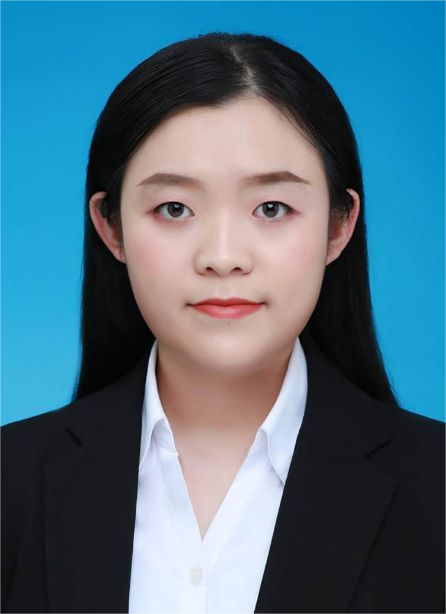}}]{{Xinyu Xu}}
		received the B.E. degree. in measurement and control technology and instrument from Tianjin University, Tianjin, China in 2022, and she is currently pursuing the M.Sc. degree in electronic information from Shanghai Jiao Tong University, Shanghai, China.
		
		Her research interests include computer vision, point cloud recognition, and deep learning in artificial intelligence.
	\end{IEEEbiography}

	\begin{IEEEbiography}[{\includegraphics[width=1in,height=1.25in,clip,keepaspectratio]{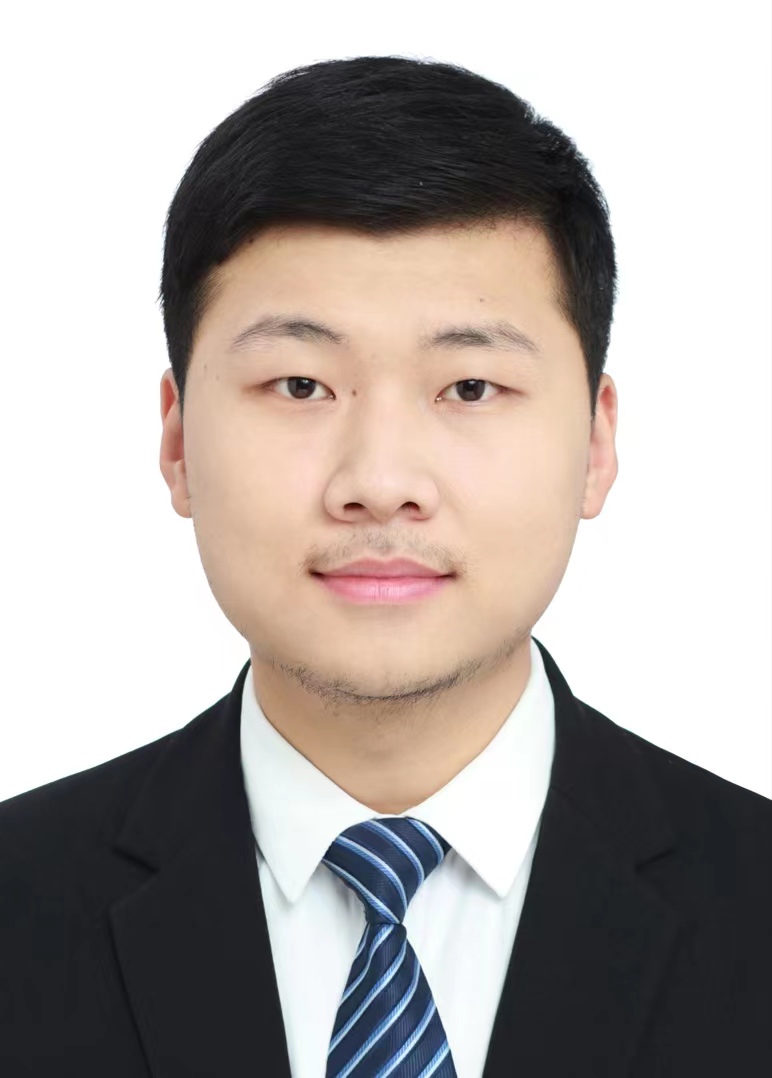}}]{{Huazhen Liu}}
		received the B.E. degree in measurement and control technology and instrument from Tian Jin University Tianjin,China, in 2022, and he is currently pursuing the M.Sc. degree in Shanghai Jiao Tong University, Shanghai, China.
		
		His primary research interests include computer vision and optical engineering in artificial intelligence.
	\end{IEEEbiography}
	
	\begin{IEEEbiography}[{\includegraphics[width=1in,height=1.25in,clip,keepaspectratio]{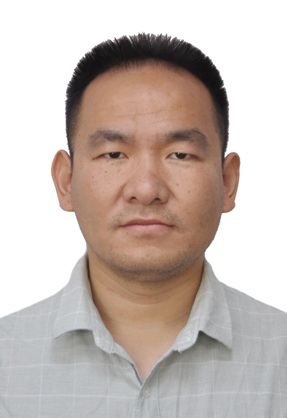}}]{Tao Zhang} (Member IEEE) received the B.S. degree in electronic information engineering from Huainan Normal University in 2011, and the M.Sc. degree in communication and information system from Sichuan University in 2014, and the Ph.D degree in control science and engineering from Shanghai Jiao Tong University in 2019. From 2017 to 2018, he was with the Department of Geoinformatics, the KTH Royal Institute of Technology, as a joint PhD student. From 2019 to 2021, he worked as a Post-Doctoral in the Department of Electronics, Tsinghua University.
		
		He is currently working as an Assistant Professor in the Shanghai Key Laboratory of Intelligent Sensing and Recognition, Shanghai Jiao Tong University. His research work focuses on PolSAR/SAR image processing, object recognition, and machine learning.
	\end{IEEEbiography}

	\begin{IEEEbiography}[{\includegraphics[width=1in,height=1.25in,clip]{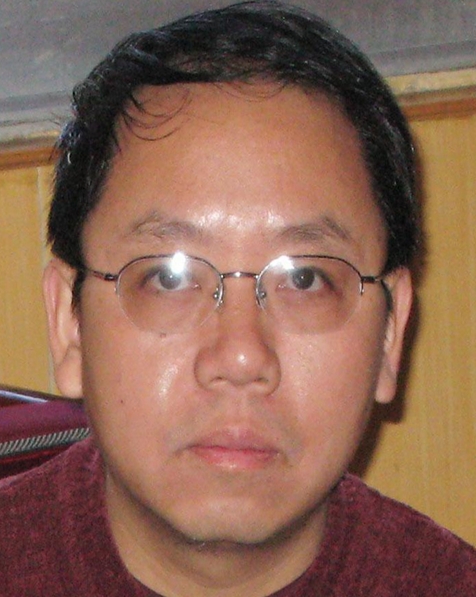}}]{Huilin Xiong} received the B.Sc. and M.Sc. degrees in mathematics from Wuhan University, Wuhan, China, in 1986 and 1989, respectively, and the Ph.D. degree in pattern recognition and intelligent control from the Institute of Pattern Recognition and Artificial Intelligence, Huazhong University of Science and Technology, Wuhan, in 1999. He joined Shanghai Jiao Tong University, Shanghai, China, in 2007, where he is a Professor with the Department of Automation.
		
		His research interests include pattern recognition, machine learning, synthetic aperture radar data processing, and bioinfomatics.
	\end{IEEEbiography}
	
	\begin{IEEEbiography}[{\includegraphics[width=1in,height=1.25in,clip,keepaspectratio]{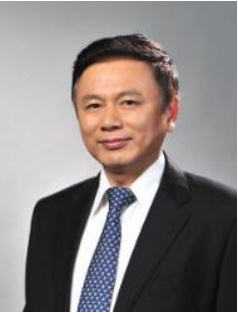}}]{Wenxian Yu} (Senior Member IEEE) was born in Shanghai, China, in 1964. He received the B.Sc., M.Sc., and Ph.D. degrees from the National University of Defense Technology, Changsha, China, in 1985, 1988, and 1993, respectively.
		
		From 1996 to 2008, he was a Professor with the College of Electronic Science and Engineering, National University of Defense Technology, where he served as the deputy dean of the school and assistant director of the National Key Laboratory of Automatic Target Recognition. In 2008, he joined the School of Electronics, Information, and Electrical Engineering, Shanghai Jiao Tong University, Shanghai, in which he served as the executive dean from 2009 to 2011. Currently, he is a Distinguished Professor under the Yangtze River Scholar Scheme, vice dean of the Advanced Industrial Technology Research Institute, and dean of the Academy of Information Technology and Electrical Engineering. His current research interests include radar target recognition, remote sensing information processing, multisensor data fusion, integrated navigation system, etc. In these areas, he has published more than 260 research papers.
	\end{IEEEbiography}

\end{document}